\documentclass[10pt,journal,compsoc]{IEEEtran}

\usepackage[dvipsnames]{xcolor} 
\usepackage{times}
\usepackage{helvet}
\usepackage{courier}
\usepackage{amsmath}
\usepackage{amssymb}
\usepackage{amsthm}
\usepackage{graphicx}
\usepackage{array}
\usepackage{booktabs}
\usepackage{amsopn}
\usepackage{amsmath,bm}
\usepackage{algorithm}
\usepackage{algorithmic}
\usepackage{enumerate}
\usepackage{multirow}
\usepackage{bbding}
\usepackage{threeparttable}
\usepackage{threeparttable} 
\usepackage{dsfont} 
\usepackage{makecell}
\usepackage{url}  
\usepackage{silence}
\usepackage{caption}
\usepackage{subcaption}
\usepackage[numbers]{natbib}
\usepackage{enumitem}
\usepackage{mathtools}
\usepackage{tcolorbox}
\usepackage{hyperref}
\usepackage{colortbl}
\usepackage{rotating}
\usepackage{cleveref}

\usepackage{adjustbox}
\usepackage{array}
\usepackage{blindtext}
\newcolumntype{R}[2]{%
    >{\adjustbox{angle=#1,lap=\width-(#2)}\bgroup}%
    l%
    <{\egroup}%
}
\newcommand*\rot{\multicolumn{1}{R{45}{1em}}}
\definecolor{section_3_1_color}{RGB}{51, 204, 0}
\definecolor{section_3_2_color}{RGB}{0, 102, 204}
\definecolor{section_3_3_color}{RGB}{204, 102, 0}
\definecolor{section_3_4_color}{RGB}{102, 0, 204}
\definecolor{section_3_5_color}{RGB}{204, 0, 0}

\newcommand{\eg}{\textit{e.g. }}
\newcommand{\etal}{\emph{et al.}}
\newcommand{\ie}{\emph{i.e.}}
\newcommand{\etc}{\emph{etc}}
\newcommand{\wrt}{w.r.t. }

\newcommand{\figref}[1]{Fig.~\ref{#1}}

\newcommand{\secref}[1]{Sec.~\ref{#1}}

\newcommand{\ma}[2]{{\textcolor{purple}{[Xianzheng: #1]}}}

\newcommand{\axc}[1]{ \noindent {\color{purple} {\bf AXC:} {#1}}}

\graphicspath{{figs/}}
\begin{document}

\title{When LLMs step into the 3D World: A Survey and Meta-Analysis of 3D Tasks via Multi-modal Large Language Models}

\author{Xianzheng Ma \textbf{*}\thanks{\textbf{*} Equal contribution. Listing order is random.},\; Brandon Smart \textbf{*},\; Yash Bhalgat \textbf{*},\;  Shuai Chen,\;  Xinghui Li,\; Jian Ding,\;\\ 
Jindong Gu,\; Dave Zhenyu Chen,\;Songyou Peng,\; Jia-Wang Bian,\;\\
Philip H Torr,\; Marc Pollefeys,\; Matthias Nießner,\;Ian D Reid,\; \\ Angel X. Chang,\;
Iro Laina,\; Victor Adrian Prisacariu

\IEEEcompsocitemizethanks{
\IEEEcompsocthanksitem Xianzheng Ma, Brandon Smart, Yash Bhalgat, Shuai Chen, Xinghui Li, Jindong Gu, Philip Torr, Iro Laina and Victor Adrian Prisacariu are with the University of Oxford.
\IEEEcompsocthanksitem Jian Ding is with the King Abdullah University of Science and Technology.
\IEEEcompsocthanksitem Dave Zhenyu Chen and Matthias Nießner are with the Technical University of Munich.
\IEEEcompsocthanksitem Jia-Wang Bian and Ian D Reid are with the Mohamed bin Zayed University of Artificial Intelligence.
\IEEEcompsocthanksitem Angel X. Chang is with Simon Fraser University.
\IEEEcompsocthanksitem Songyou Peng and Marc Pollefeys are with ETH Zurich.
\IEEEcompsocthanksitem Correspondence: xianzheng@robots.ox.ac.uk
}
}

\IEEEtitleabstractindextext{%
\begin{abstract}
As large language models (LLMs) evolve, their integration with 3D spatial data (3D-LLMs) has seen rapid progress, offering unprecedented capabilities for understanding and interacting with physical spaces. This survey provides a comprehensive overview of the methodologies that enable LLMs to process, understand, and generate 3D data. Highlighting the unique advantages of LLMs, such as in-context learning, step-by-step reasoning, open-vocabulary capabilities, and extensive world knowledge, we underscore their potential to significantly advance spatial comprehension and interaction within embodied Artificial Intelligence (AI) systems. Our investigation spans various 3D data representations, from point clouds to Neural Radiance Fields (NeRFs). It examines their integration with LLMs for tasks such as 3D scene understanding, captioning, question-answering, and dialogue, as well as LLM-based agents for spatial reasoning, planning, and navigation. The paper also includes a brief review of other methods that integrate 3D and language.
The meta-analysis presented in this paper reveals significant progress yet underscores the necessity for novel approaches to harness the full potential of 3D-LLMs. Hence, with this paper, we aim to chart a course for future research that explores and expands the capabilities of 3D-LLMs in understanding and interacting with the complex 3D world. To support this survey, we have established a project page where papers related to our topic are organized and listed: \url{https://github.com/ActiveVisionLab/Awesome-LLM-3D}.
\end{abstract}

}

\maketitle

\IEEEdisplaynontitleabstractindextext

\IEEEpeerreviewmaketitle

\IEEEraisesectionheading{\section{Introduction}\label{Sec:Introduction}}
\IEEEPARstart{T}{he} advent of Large Language Models (LLMs) has marked a transformative era in natural language processing, enabling machines to understand, generate, and interact with human language in ways previously unimaginable. However, the physical world around us is inherently three-dimensional, and understanding spatial environments is crucial for many real-world applications that involve perception, navigation, and interaction within these 3D spaces.
With recent advances, the application of LLMs has extended well beyond text.
The fusion of LLMs with 3D data presents a unique opportunity to enhance computational models' understanding of, and interaction with the physical world, leading to innovations across various domains, including autonomous systems~\cite{chen2023driving,sha2023languagempc,fu2024drive,xu2023drivegpt4, ma2022both}, augmented reality~\cite{azuma1997survey,carmigniani2011augmented,craig2013understanding,feiner1997touring}, robotic navigation~\cite{rt22023arxiv,zheng2023towards,song2023llmplanner}, and robotic manipulation~\cite{huang2023voxposer,mirjalili2023lan,li2023manipllm}. 

\begin{figure*}[ht!]
    \centering
    \includegraphics[width=\textwidth]{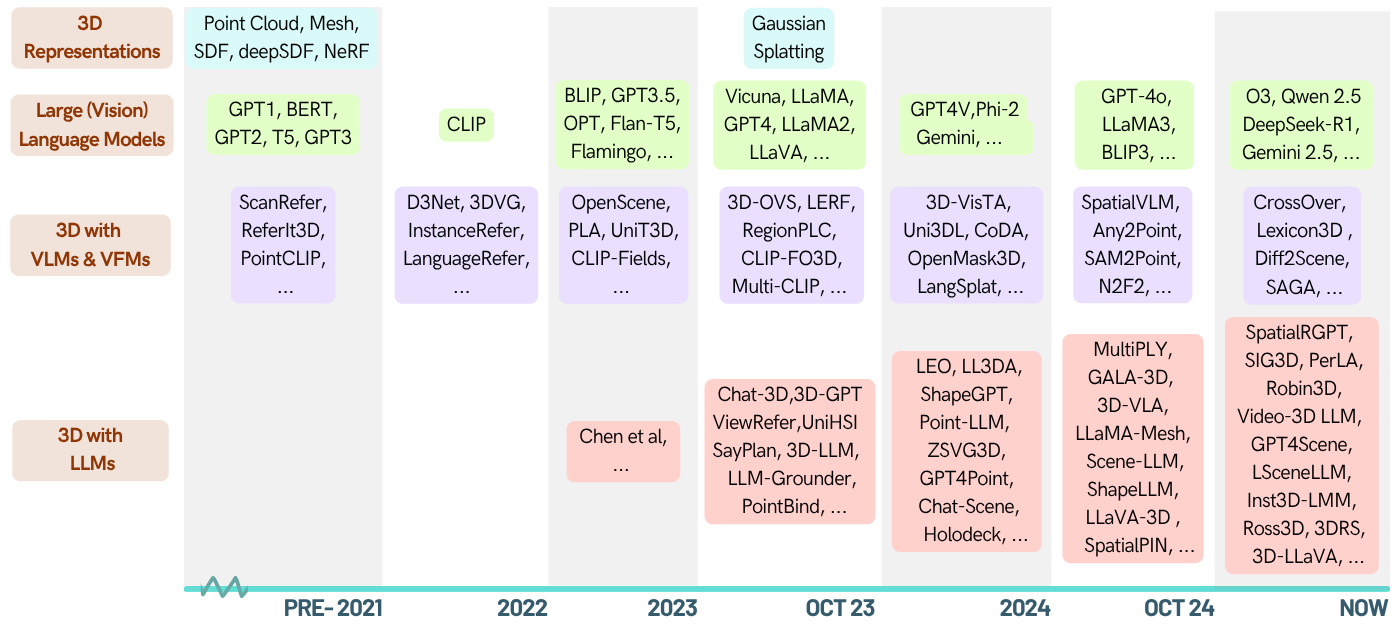}
    
    \caption{ \textbf{An overall timeline} to show the developments of 3D representations (\secref{sec:3d_representations}), L(V)LMs (\secref{sec:bkground_large_language_model} and \secref{sec:2D_VLM_VFM}), 3D w/VLMs\&VFMs(\secref{sec:3d-vlms}), 3D w/LLMs (\secref{sec:3d-llms}). This figure illustrates how the development of 3D representation and L(V)LMs inspired the 3D vision-language methods (3D w/LLMs, 3D w/VLMs VFMs). Besides, we can clearly see the rapid growth of 3D-LLM methods starting from 2023, which calls for the attention of more researchers.}
    \label{fig:timeline}
\end{figure*}

Recent works have demonstrated the potential of integrating LLMs with 3D data to interpret, reason, or plan in complex 3D environments, by leveraging the inherent strengths of LLMs, including zero-shot learning~\cite{yuan2023visual,zhang2024agent3dzero}, advanced reasoning~\cite{huang2023voxposer,jatavallabhula2023conceptfusion,chen2023ll3da}, and extensive knowledge~\cite{guo2023viewrefer,llmgrounder}.
However, despite rapid advances in this area, research remains scattered and relatively nascent, with most efforts emerging only since 2023 (see Fig.~\ref{fig:timeline}). This fragmentation creates a need for a comprehensive and systematic survey to consolidate the current state of research and provide a clear overview of the methodologies, challenges, and future directions. The aim of this survey is to fill these gaps by providing a detailed analysis of the field, enabling researchers to better understand the progress made thus far and to identify key areas for future exploration.

To the best of our knowledge, we are the first to survey the field of 3D-LLMs. We outline the methodologies that are currently being used to allow LLMs to ingest and process 3D information in Fig. \ref{fig:architectures}, and present a novel taxonomy categorizing 3D-LLM methods based on the role of the language model in solving 3D tasks in Fig. \ref{fig:taxonomy}. We describe areas where the amalgamation of LLMs and 3D data has been successfully demonstrated, such as: using LLMs' world-knowledge \cite{guo2023viewrefer,abdelreheem2023zero} and reasoning capabilities \cite{llmgrounder,fang2023transcribe3d} to enhance 3D task performance, using LLMs as multi-modal interfaces \cite{hong2024multiply,jatavallabhula2023conceptfusion} and embodied agents \cite{li2023manipllm,huang2023voxposer}, or generating complex 3D scenes with LLMs \cite{gala3d,3dgpt}. We then provide a multifaceted comparison between existing methods, with a focus on the 3D vision components, LLM components, and alignment methodologies used to train these models (see Tab.~\ref{tab:taxonomy}).

Compared to other LLM-related surveys, which either focus on single tasks like video understanding~\cite{tang2024videounderstand}—or on specific scenarios like robotics~\cite{zeng2023robot1,firoozi2023robot2,hu2023robot3} or autonomous driving~\cite{cui2023surveyAutonomousDriving}, our survey focuses on the broad potential of integrating LLMs with 3D data. Based on this, we highlight how the advantages and disadvantages of different 3D representations (\secref{sec:3d_representations}), and how these representations can be combined with LLMs to accomplish various 3D tasks (\secref{sec:3d-llms}). Additionally, we cover 3D methods utilizing Vision-Language models (VLMs) and Vision-Foundation Models (VFMs) (\secref{sec:3d-vlms}). Altogether, these solutions could provide insights for various real-world scenarios such as AR/VR, healthcare, game development, and interior design.

In addition to the above, we offer a thorough review of 3D vision-language tasks (\secref{sec:tasks}), categorizing them into five major types based on differences in the input and output modalities. We also compare datasets used for training and testing (\secref{sec:datasets}), mapping the trajectory of this field through the evolution of datasets (see Fig.~\ref{fig:dataset-timeline}). Finally, we identify key challenges and outline future directions (\secref{sec:7-discussion}).

\section{Background}~\label{sec:background}
This section offers background knowledge on 3D representations, Large Language Models (LLMs), 2D Vision-Language Models (VLMs), and Vision Foundation Models (VFMs).

\subsection{3D Representations}~\label{sec:3d_representations} 


Choosing an effective 3D representation is a critical design decision in 3D-LLMs research. It is also a fundamental research area in computer vision and graphics, driven by advances in deep learning, computational resources, and the availability of 3D data.
In practice, \textbf{image-based representations}, such as multi-view RGB images and RGB-D data, although not strictly 3D, are often employed as proxies for 3D geometry, owing to their ease of acquisition and strong compatibility with existing 2D vision backbones. Nevertheless, these representations mainly capture geometry at the pixel level.
Conversely, \textbf{structured scene representations}, such as 3D scene graphs, offer semantic abstraction over spatial layouts and are increasingly adopted for reasoning tasks in embodied AI. Below, we briefly describe the most common 3D representations.

\textbf{Point Clouds} represent 3D shapes as discrete data points in space, typically storing the XYZ coordinates and optionally additional attributes such as color or surface normals. Point cloud-based methods~\cite{qi2017pointnet, liu2019flownet3d} are lightweight and easy to obtain (e.g., from LiDAR, structured light, or photogrammetry), but lack explicit surface topology.

\textbf{Voxel Grids}~\cite{dai2017shape, dai20183dmv} divide the 3D space into regular volumetric units (voxels), analogous to pixels in 2D. Each voxel can encode occupancy (binary or probabilistic), surface distance (\eg Signed Distance Function~\cite{Peng99, Osher04}), or Truncated SDF~\cite{Curless96, newcombe2011kinectfusion}. While intuitive, voxel grids become memory-intensive at high resolution, which motivates sparse voxel structures~\cite{riegler2017octnet}.

\textbf{Meshes} use interconnected vertices, edges, and faces to compactly model 3D geometry with surface continuity. Despite their expressiveness, they are less favored in learning-based pipelines due to their irregular topology. Solutions include gradient approximation~\cite{Genova18, Kato19} and differentiable rasterization~\cite{Rhodin15}, though these often suffer from rendering artifacts or limited precision.

\textbf{Neural Fields}~\cite{xie2022neural} represent 3D scenes as continuous functions (typically MLPs) that map spatial coordinates to physical properties such as occupancy, color, or semantics. Unlike discrete voxel grids, neural fields model scenes as continuous volumes. For example, \textit{Occupancy Networks}~\cite{Mescheder19,Peng20} predict whether a point lies inside an object, while \textit{Deep SDF}~\cite{Park19, Sitzmann19SRN} learns continuous signed distance fields. These representations are compact and flexible but often slow to evaluate.

\textbf{Neural Radiance Fields (NeRF)}~\cite{Mildenhall20,barron2022mip} extend neural fields by modeling view-dependent color and density along camera rays. This enables photorealistic novel-view synthesis but requires expensive sampling and evaluation, especially in empty space, prompting research on acceleration.

\textbf{Hybrid Representations} combine neural fields with explicit structures to improve efficiency. For instance, voxel grids~\cite{yu2022plenoxels} or multi-resolution hash grids~\cite{muller2022instant} are used to condition lightweight MLPs, significantly reducing NeRF training and inference cost while maintaining high rendering quality.

\textbf{3D Gaussian Splatting}~\cite{kerbl3Dgaussians} offers a differentiable point-based representation, where each point emits anisotropic Gaussian blobs instead of sharp vertices. This enables real-time, high-fidelity rendering. Initializations typically come from SfM-based point clouds~\cite{schoenberger2016sfm}, though recent works~\cite{Fu_2024_CVPR, fan2024instantsplat} explore SfM-free pipelines.

\begin{figure*}[!th]
    \centering
    \includegraphics[width=\linewidth]{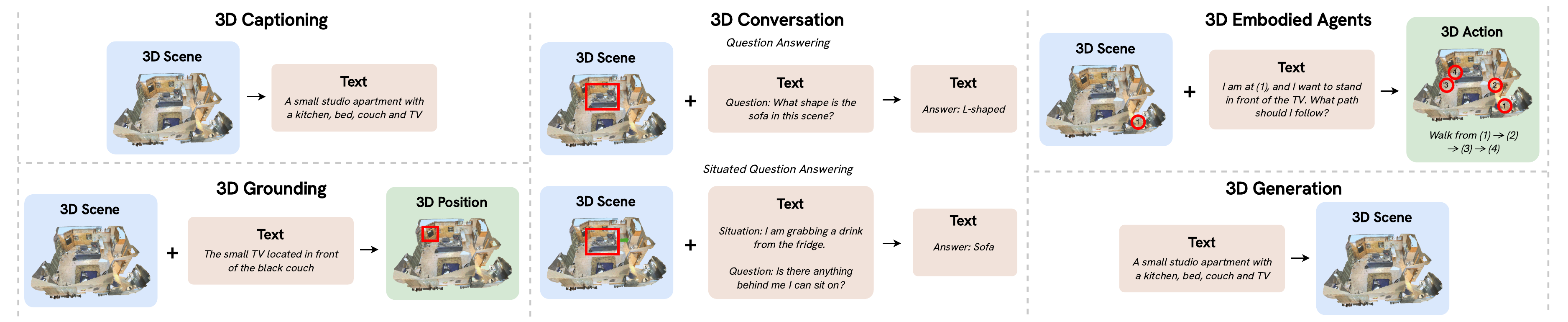}
    \vspace{-5mm}
    \caption{\textbf{Common task categories} for 3D Vision Language models. Each category contains many types of queries, but we provide a single example from each category. The scan is from the ScanNet dataset~\cite{dai2017scannet}, and the `question answering' and `situated question answering' examples are adapted from ScanQA~\cite{scanqa} and SQA3D~\cite{ma2022sqa3d} respectively.
    }
    \vspace{-3mm}
    \label{fig:tasks_overview}
\end{figure*}

\subsection{Large Language Models (LLMs) }~\label{sec:bkground_large_language_model}
Natural Language Processing (NLP) aims to enable systems to understand, generate, and manipulate text. Early approaches used rule-based systems, statistical models, or neural architectures like recurrent neural nets~\cite{elman1990finding}. The rise of LLMs, powered by transformers~\cite{Vaswani17} and vast text corpora~\cite{kaplan2020scaling}, has transformed the field. Since this paper focuses on 3D LLMs, we provide a brief background on LLMs here. For a more detailed coverage, see recent surveys on this topic~\cite{zhao2023survey, minaee2024large, wei2022emergent}.
\subsubsection{LLM Architectures} \label{sec:LLM_101}

Large language models are typically built upon either \textbf{encoder-decoder} or \textbf{decoder-only} transformer architectures~\cite{Vaswani17}. Encoder-decoder models process the input sequence via an encoder to produce contextual embeddings, which are then consumed by the decoder to autoregressively generate the output. In contrast, decoder-only architectures~\cite{liu2018generating} generate outputs directly from previous tokens without a separate encoder, making them particularly efficient for generative tasks. Another critical component is \textbf{tokenization}, which converts raw text into discrete tokens. Choices like byte-pair encoding (BPE)~\cite{bytepairencoding}, WordPiece~\cite{wordpiece}, and SentencePiece~\cite{kudo2018sentencepiece} significantly impact model performance and vocabulary expressivity. These architectural and preprocessing designs influence how language and multimodal inputs are represented and processed, and they are often adapted or extended in 3D-LLMs to support spatial encoders and cross-modal alignment.

\subsubsection{LLM Emergent Abilities} \label{sec:LLM_Emergent_Abilities}

A defining feature of LLMs is their \textit{emergent abilities}, which arise as model scale increases~\cite{wei2022emergent}, enabling zero-shot generalization across diverse tasks without explicit training. These include \textbf{in-context learning}~\cite{brown2020language}, where the model adapts to new tasks based on prompts alone; \textbf{reasoning}, often enabled by chain-of-thought prompting~\cite{wei2022chain}, where intermediate steps are generated to solve complex problems; and \textbf{instruction-following}~\cite{touvron2023llama}, where the model executes natural language commands via instruction tuning. These abilities underpin the remarkable generality of LLMs and have directly inspired recent 3D-LLMs to adopt similar paradigms—using in-context demonstrations, compositional reasoning, and open-ended instruction-following over multimodal 3D inputs to enhance task generalization without task-specific retraining.

\subsubsection{LLM Fine-tuning} \label{sec:LLM_fine-tuning}

Given the large number of parameters in LLMs, full fine-tuning is often infeasible for 3D-LLMs. Instead, \textit{parameter-efficient fine-tuning} (PEFT) methods have become the standard to adapt LLMs to 3D tasks with reduced computational cost. \textbf{LoRA}~\cite{hu2021lora} and its variants~\cite{dettmers2024qlora,zhang2023lora} insert trainable low-rank matrices into transformer layers, allowing efficient adaptation while keeping most of the original weights frozen. \textbf{Layer freezing}~\cite{devlin2018bert,howard2018universal} updates only a small subset of layers, typically at the input/output, preserving general language knowledge and reducing overfitting. \textbf{Prompt tuning} steers model behavior by modifying input prompts instead of model parameters, including handcrafted prompts~\cite{wei2022chain}, discrete token search (\textit{hard prompts})~\cite{shin2020autoprompt}, or learned embeddings (\textit{soft prompts})~\cite{lester2021power}. Finally, \textbf{adaptive tuning}~\cite{chat3d,huang2023chatV2} integrates lightweight neural modules between frozen layers to enable efficient cross-modal adaptation, such as incorporating 3D data alongside text. These PEFT strategies are widely adopted in 3D-LLMs to enable scalable and modular alignment of LLMs with 3D inputs.

\subsection{Vision-Language and Vision  Foundation Models}~\label{sec:2D_VLM_VFM}
2D Vision and Vision-Language Foundation Models aim to learn universal visual representations and multimodal alignments, enabling transfer to a wide range of downstream tasks.

\textbf{Vision-Language Models (VLMs)} such as CLIP~\cite{radford2021learning} and ALIGN~\cite{jia2021scaling} align image and text embeddings through contrastive training, achieving strong zero-shot transfer in classification and extending to tasks like detection~\cite{zhong2022regionclip}, segmentation~\cite{luddecke2022clipseg}, and video understanding~\cite{ni2022xclip}. Later VLMs like BLIP-2~\cite{li2023blip2}, Flamingo~\cite{alayrac2022flamingo}, and LLaVA~\cite{liu2024llava} support image-conditioned generation and multi-turn interaction. Text-to-image diffusion models~\cite{rombach2022sd} further expand the generative capabilities, with extensions to video~\cite{ho2022imagenvideo} and 3D generation~\cite{poole2022dreamfusion, singer2023text24d}.

\textbf{Vision Foundation Models (VFMs)} focus on learning rich visual representations via self-supervised or large-scale supervised pretraining. DINO~\cite{caron2021emerging} and its successors~\cite{zhou2021ibot, oquab2023dinov2} use contrastive and masked image modeling strategies, demonstrating generalization in classification, segmentation, and matching. Segment Anything Model (SAM)\cite{kirillov2023segment} enables zero-shot segmentation across diverse domains. Beyond this, diffusion models\cite{rombach2022sd} also serve as feature extractors~\cite{tang2024dift, tian2023diffuse}, revealing semantic priors through intermediate representations.

These 2D VLMs and VFMs lay the foundation for recent efforts extending multimodal learning into the 3D domain, which we discuss in the following sections.

\section{Tasks and Metrics}

\label{sec:tasks}

To understand the role of language in 3D understanding, it is important to first understand the tasks that 3D vision-language models attempt to solve. Research has grown to include a broad spectrum of research tasks, each with their own set of commonly used datasets and evaluation metrics. Here, we aim to summarize current 3D vision-language tasks, and their corresponding evaluation metrics. We broadly categorize the tasks by their input and output modalities, and provide an example for each category in Fig. \ref{fig:tasks_overview}.
We then describe methods for solving these tasks in \secref{sec:3d-llms} and \secref{sec:3d-vlms}. Then, in \secref{sec:datasets}, we detail the datasets that are currently used for training and evaluation for these tasks.

\subsection{3D Captioning (3D \texorpdfstring{$ \rightarrow $}{to} Text)}

Given the 3D data of a scene or object, the task of 3D captioning is to generate a corresponding short, natural language description. Here, we decompose this task into a few common variants of the problem, based on the type of data being captioned, and the type of captions that are generated.

\noindent \textbf{Object-Level Captioning} requires the model to generate a short, natural language description of a single 3D object. This caption should focus on the key characteristics of the object, including its shape and semantic characteristics.

\noindent \textbf{Scene-Level Captioning} refers to the task of generating a short, natural language caption for an entire 3D scene. These captions typically focus on global scene information (such as room types and styles), key objects in the scene, and their relationships. We consider ``grounded captioning'', where the model outputs a description of the relationships between objects in the scene, potentially alongside positional information for those objects, to be a variant of scene captioning.

\noindent \textbf{3D Dense Captioning}  refers to the joint task of localizing instances of objects in a 3D scene and describing them using natural language captions. In this instance, the output may also contain positional information about the objects being captioned. Often, the referring descriptions from 3D grounding datasets are used to produce the captioning and location data needed for 3D dense captioning. For example, captions in Scan2Cap~\cite{chen2021scan2cap} are generated using the referring expressions from ScanRefer~\cite{scanrefer}.

\noindent \textbf{Evaluation Metrics} for 3D captioning require comparing the generated captions against ground truth captions for testing samples. 
Exact Match (EM) requires that the generated caption exactly matches the ground truth. Exact Match has different accuracy thresholds, denoted as EM@$K$, which means that the correct answer is within the top ``$K$'' answers generated by the model. Commonly used thresholds are EM@1 and EM@10. However, natural language captions with the same semantic meaning can be represented in many ways, so the dominant metrics for captioning are automated text generation metrics~\cite{celikyilmaz2020evaluation} that aim to measure matching n-grams or semantic similarity rather than complete sentence matches. BLEU~\cite{papineni2002bleu} matches n-grams between the predicted and true captions, with ``BLEU@$x$'' referring to matching n-grams of length ``$x$'' (typical values are in the range 1-4). This still requires matching exact words, but is slightly more robust to rearrangements in phrasing. ROUGE~\cite{lin2004rouge} similarly aims to match n-grams, with the commonly used ROUGE-L focusing on the structural similarity of sentences. METEOR~\cite{banerjee2005meteor}, is based upon the precision and recall of unigram matches, with ``matches'' also existing between synonyms and words which are morphological variants of each other. CIDEr~\cite{vedantam2015cider} weights n-grams by their frequency, with higher frequency n-grams given lower weights. As the above metrics rely on n-gram matches, they cannot account for different but semantically similar words. Thus, various metrics that measures semantic content overlap via similarity in learned embedding spaces (\eg,  SentenceSim~\cite{reimers2019sentence} and BERTScore~\cite{zhang2019bertscore}) have been introduced.

For dense captioning, where captions are localized to parts of the scene, adjusted benchmarks are needed. Often, BLEU, ROUGE, METEOR and CIDEr scores are still used, however the score is set to zero if the intersection over union (IoU) between the predicted bounding box and the object is less than a threshold ``$k$''. Typical ``$k$'' values are 0.25 and 0.5~\cite{chen2023ll3da, chen2021scan2cap, chen2023end}. However, these metrics focus on the captioning recall while ignoring false positives. This is addressed by more recent work that additionally measures the precision and F-1 score of generated captions \wrt the BLEU, ROUGE, METEOR and CIDEr metrics~\cite{chen2023unit3d}.

\subsection{3D Grounding (3D + Text \texorpdfstring{$ \rightarrow $}{to} 3D Position)}
Given a 3D scene and a ``referring expression'' that describes an object in the scene relative to other objects, 3D grounding involves generating a position, bounding box or segmentation mask for the target object(s).

\noindent \textbf{Single-Object Grounding} involves locating a single query object in a scene given reference information, such as language descriptions~\cite{scanrefer,achlioptas2020referit3d} or additional gestures~\cite{ScanERU}.

\noindent \textbf{Multi-Object Grounding} involves locating multiple objects using a referring expression. There are two main variants for this kind of grounding. The first involves a single-sentence description that may be ambiguous, potentially referring to zero, one, or multiple target objects of the same category in the 3D scene~\cite{zhang2023multi3drefer}. The second variant uses paragraph-length referring expressions that describe multiple objects belonging to potentially different categories, and the spatial relationships between them~\cite{densegrounding}.

\noindent \textbf{Evaluation Metrics} for 3D grounding require comparing predicted locations, most often in the form of bounding boxes, with ground truth locations of objects from testing samples.
Acc@$K$IoU~\cite{scanrefer} is the widely used metric in 3D visual grounding, which measures the percentage of positive predictions which have a intersection over union (IoU) with the ground truth greater than a threshold $K$, which is usually set to 0.25 or 0.5. It is worth noting that some datasets evaluate the performance in different scenarios. For example, ScanRefer~\cite{scanrefer} divides the datasets into unique/multiple/overall splits. Some methods measure the average IoU~\cite{3dllm, TGNN} whereas other methods measure the average distance between the centers of the bounding boxes~\cite{3dllm}.
For multi-object grounding, the F1 score is used as the metric~\cite{zhang2023multi3drefer}. They first get the one-to-one matching between the predicted and ground truth bounding boxes according to IoUs. Then the pairs with IoUs higher than a threshold are regarded as true positives. 

\subsection{3D Conversation (3D + Text \texorpdfstring{$ \rightarrow $}{to} Text)}
It is also natural to consider tasks where questions are asked about a 3D scene, either in a single-turn setting or a more natural multi-turn conversational setting.

\noindent \textbf{3D Question Answering (3D-QA)} is a task where the model is required to generate answers to questions asked by users given a 3D scene. The topic of questions has a diverse range and the model has to understand both the 3D scene and the question to generate the correct response. 
The question includes both simple tasks such as determining the existence of an object and more difficult ones like spatial reasoning. As there are several well-established benchmarks, and the majority of questions in the benchmarks are factual with unique answers, 3D-QA is a popular task to evaluate a multi-task model's capability.

\noindent \textbf{3D Situated Question Answering (3D-SQA)} is a special case of 3D-QA. The key difference is that 3D-QA requires the model to answer questions from the perspective of a spectator with access to all information about the scene, while 3D-SQA needs the answer from the perspective of a player in a pre-defined situation. For example, 3D-SQA may ask ``how many chairs are in front of me?'' given the situation of ``facing the dining table''.

\noindent \textbf{3D Dialogue} requires the model to have a coherent and natural multi-turn conversation with users about the 3D scene, instead of one round QA. For example, a user may want to know about a room, so they continuously ask questions about each part of the room, while the model is expected to respond correctly and coherently. 

\noindent \textbf{Evaluation Metrics} involve comparing a model's responses with ground truth responses for testing samples. For 3D-QA and 3D-SQA, the dominant metric is the Exact Match (EM), meaning that the answer generated by the model must exactly match with the correct answer. This is because the majority of questions in existing 3D-QA benchmarks \cite{ma2022sqa3d, scanqa, qian2023nuscenes, hong20233d} are factual questions where there is only one definitively correct answer. For 3D Dialogue and Task Planning whose answers are non-unique, the semantic metrics such as BLEU \cite{papineni2002bleu}, ROUGE \cite{lin2004rouge}, METEOR \cite{banerjee2005meteor}, CIDEr \cite{vedantam2015cider}, and SPICE \cite{anderson2016spice} are applied to evaluate the similarity between generated responses and reference answers provided by benchmarks. They are also used in 3D-QA, particularly the ScanQA benchmark, to measure the semantic similarity alongside the accuracy.

\begin{figure*}[!th]
    \centering
    \includegraphics[width=\linewidth]{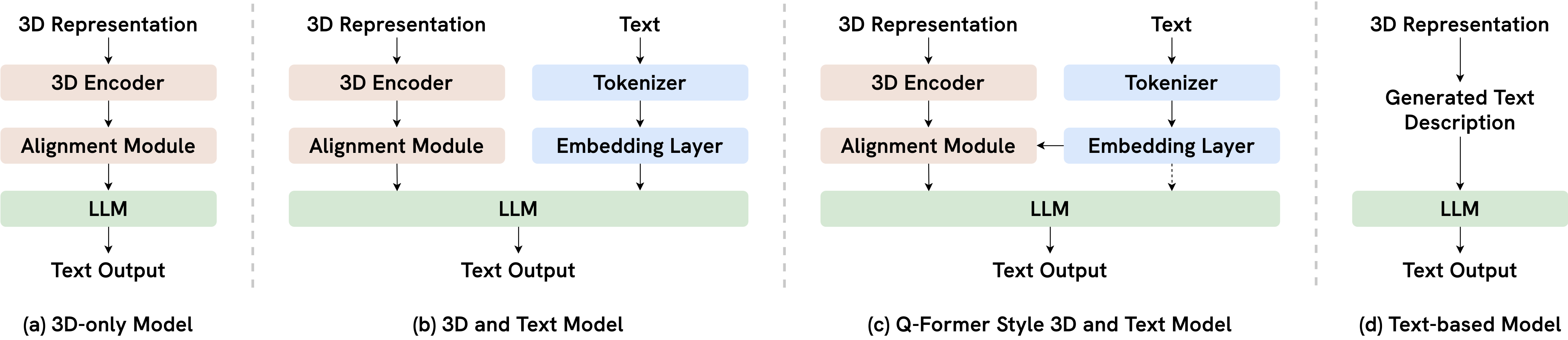}
    \caption{\textbf{Architectures for aligning 3D with text for LLMs.} Here we show four high-level architectures: (a) 3D-only model that aligns 3D features to the LLM's input space, (b) 3D+text model where 3D features and text are both aligned, (c) Q-Former style model where text is used during to condition the alignment of the 3D features, and optionally given to the LLM itself (dashed arrow), and (d) text-only approach which converts 3D representations into text strings, avoiding the need to train an alignment module.
    }
    \label{fig:architectures}
\end{figure*}
 
\subsection{3D Embodied Agents (3D + Text \texorpdfstring{$ \rightarrow $}{to} Action)}
It is also useful to consider tasks that involve interactions with the 3D scene, conditioned on a specific text prompt describing the desired action or goal.

\noindent \textbf{3D Task Planning} is the task where users provide a high-level objective and the model is required to outline low-level steps to fulfill this objective. For example, given a 3D scene of a room, users may ask how to clean the room and the model needs to offer detailed steps to clean it. 

\noindent \textbf{3D Navigation} refers to the task of enabling 3D agents, such as robots or virtual characters, to move and orient themselves within 3D spaces. This involves understanding and interpreting the 3D environment, identifying obstacles, and planning safe, efficient paths to reach designated goals.

\noindent \textbf{3D Manipulation} refers to the ability of 3D agents to physically interact with objects in their environment. This can range from picking up and moving objects to more complex sequences of actions such as assembling parts or opening doors. 

\noindent \textbf{Evaluation Metrics} for 3D Task Planning also rely on matching the textual/token output of a model with the ground truth actions for test samples. BLEU \cite{papineni2002bleu}, ROUGE \cite{lin2004rouge}, METEOR \cite{banerjee2005meteor}, CIDEr \cite{vedantam2015cider}, and SPICE \cite{anderson2016spice} are applied to evaluate the similarity between generated responses and ground-truth answers.

For 3D Navigation, there are two primary metrics to evaluate the performance. 1) Success Rate (SR) measures whether the 3D agent reaches the target locations within a predefined distance threshold. 2) Success Rate Weighted by Path Length (SPL)~\cite{anderson2018evaluation}, which is calculated as the SR weighted by the ratio of the ground truth length and actual path length, aims to reflect how efficiently the model achieved its goal. Other metrics include Oracle Success Rate (OSR), Trajectory Length (TL) and Goal Process (GP)~\cite{anderson2018vision}.
Please note that our discussion focuses exclusively on metrics used within 3D-LLMs methods. We encourage readers to refer to Gu \etal~\cite{gu2022vln} for a summary of navigation metrics.

For 3D Manipulation, the key metric is Success Rate~\cite{shridhar2022cliport}, which for manipulation is defined as the number of successful manipulations divided by the total number of task samples.
As discussed in \secref{subsec:3D Embodied Agent}, different datasets have different conventions for how their actions are represented using text, such as using structured outputs, using normalized numeric scores, or introducing new tokens.


\subsection{Text-to-3D Generation (Text \texorpdfstring{$ \rightarrow $}{to} 3D)}
Beyond using text to describe and interact with existing 3D scenes, it is also possible to generate 3D objects and scene via language specification.  Here we give a brief summary of this area, see Lee \etal~\cite{lee2024text} for a more in-depth survey. 

\noindent\textbf{3D Object Generation} involves generating 3D models of individual objects from text descriptions. The text input can provide details about the object's category, attributes, part structure, and other properties that should be reflected in the generated 3D shape.

\noindent\textbf{3D Scene Generation} is the task of creating full 3D environments such as rooms or outdoor spaces based on text scene descriptions. This involves generating 3D models for objects specified in the text as well as intelligently arranging and composing multiple 3D object models given the constraints specified in the text, like object categories, counts, spatial relationships, and scene attributes.

\noindent\textbf{3D Editing} refers to modifying existing 3D assets like shapes or scenes based on text instructions. This could involve adding, removing, or transforming objects, changing materials or colors, or altering high-level scene properties according to the given text.

\noindent\textbf{Evaluation Metrics} for 3D generation tasks assess the quality of the generated shape/scenes and how well the generated content matches the input text.  Common metrics to measure generated geometry include Chamfer Distance (CD) and Mesh-Volume/Surface Distance (MVD). CD is calculated by summing squared point-to-point distances w.r.t.\ ground-truth 3D data, while MVD calculates the volume/surface between two meshes to measure the geometric error. To assess overall quality, classification accuracy checks if semantic properties are preserved, while the Fréchet Inception Distance (FID) captures realism and diversity. To check if the generated shape matches the input text, it is common to measure the similarity of the text with either aligned embeddings of the 3D shape (\eg ULIP~\cite{xue2023ulip}) or rendered images (\eg CLIP~\cite{radford2021learning}).  
It is also common to use human studies for evaluation.  However, recent work~\cite{wu2024gpt} has shown it is possible to use LVLMs like GPT-v4 as alternative to using human judges. 
For text-based 3D editing, CD and IoU evaluate how well instructed edits were applied to input geometry without excessive distortion.

\section{3D Tasks with LLMs}
\label{sec:3d-llms}

3D scene understanding tasks have been widely studied. 
At its core, scene understanding entails recognizing and categorizing all objects present within a designated 3D environment, a process known as semantic~\cite{armeni20163d,sengupta2013urban,mccormac2017semanticfusion,dai20183dmv,HuangZhan2020PartAssembly,cheng2023score} or instance-level~\cite{jiang2020pointgroup,hou20193d,wang2018sgpn,han2020occuseg,song2019apollocar3d,zhan2023amodal,Zhan2023physd} understanding. This stage is critical, as it forms the basis upon which more nuanced interpretations are built. 
Subsequently, a higher level of scene understanding focuses on spatial understanding, which refers to the construction of spatial scene graphs ~\cite{feng2023exploring,zhang2021holistic} and the semantics of object relationships~\cite{zhang2021deeppanocontext,zhan2022triocc}. 
Going further, potential interactions can be predicted, such as affordances~\cite{huang2023voxposer,mirjalili2023lan,li2023manipllm,SceneFun3D,cheng2023occlusion}, scene changes~\cite{qiu20203d,looper20233d}, and understanding the broader context of the scene, \eg the functionality and aesthetic styles~\cite{fu2024scenellm}.
3D data also presents unique challenges that are not present in 2D, such as the relatively high cost of obtaining and labeling 3D data, working with sparse 3D data structures that are not uniformly dense or aligned to a grid, and the need to reconcile multiple (possibly occluded) viewpoints of the same objects~\cite{zhan2022triocc,zhan2023amodal}. 
To this end, researchers have leveraged the power of language, where the semantics and relationships within a 3D world can be embedded.
Recent efforts in integrating LLMs with 3D data have shown promise in achieving multi-level understanding and interaction, leveraging LLMs' inherent strengths, namely zero-shot learning, in-context learning, step-by-step reasoning, and extensive world knowledge.

\begin{figure*}[!th]
    \centering
    \includegraphics[width=\linewidth]{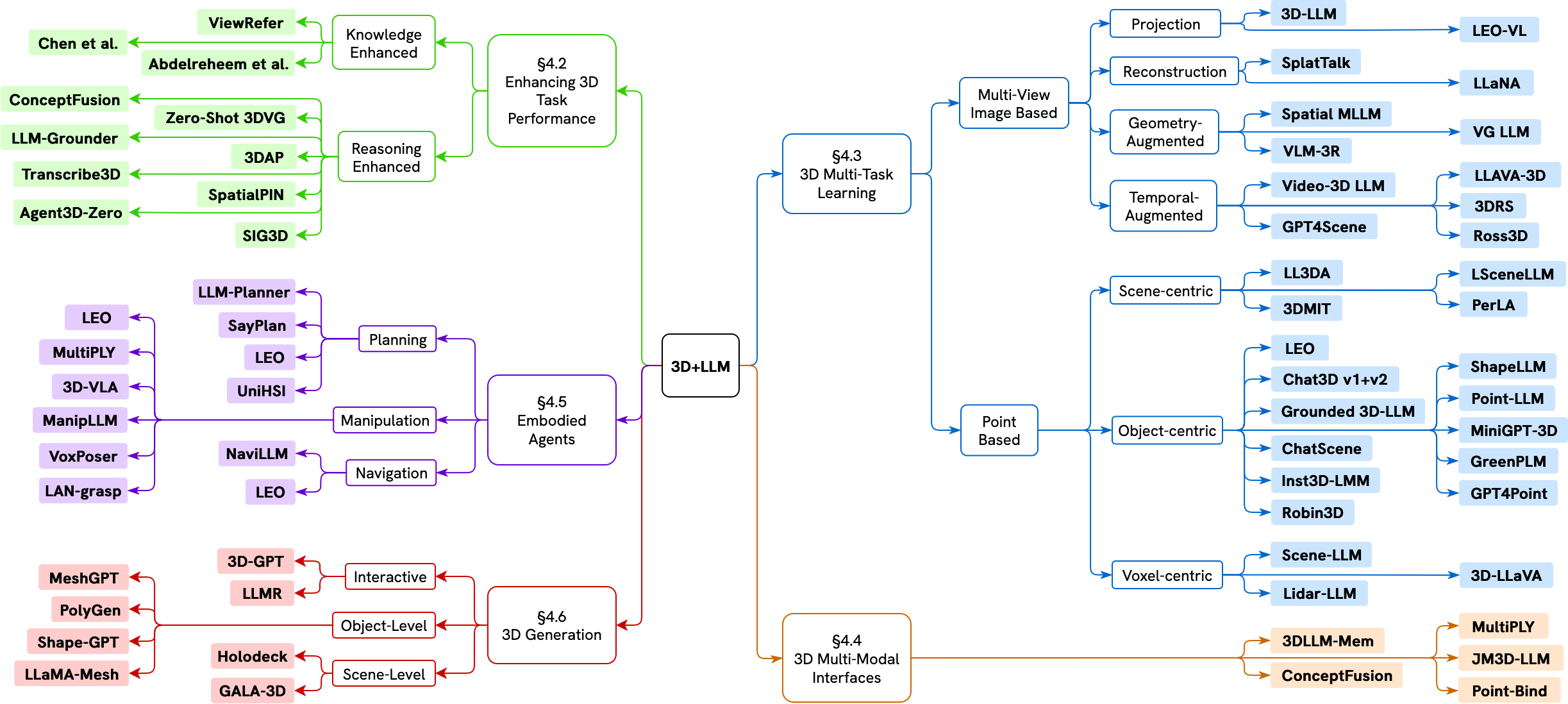}
    \caption{\textbf{Taxonomy of 3D with LLM methods.} In \secref{sec:3d-llms}, we analyze the role LLMs have played in solving 3D tasks from five perspectives: \textcolor{section_3_1_color}{Enhancing 3D Tasks}, \textcolor{section_3_2_color}{Multi-Task Learning}, \textcolor{section_3_3_color}{3D Multi-modal Interfaces}, \textcolor{section_3_4_color}{Embodied Agents}, and \textcolor{section_3_5_color}{3D Generation}.
    }
    \label{fig:taxonomy}
    \vspace{-3mm}
\end{figure*}

In \secref{sec:3.1.1}, along with Fig.~\ref{fig:architectures}, we provide a brief description of how LLMs process 3D scene information, highlighting how 3D features are aligned with language so that they can be interpreted and reasoned with via LLMs, which is foundational for subsequent sections. The rest of this section is structured to align the taxonomy presented in Fig.~\ref{fig:taxonomy}, which describes the role LLMs have played in solving 3D tasks. 
We start by showing how the world-knowledge (sometimes referred to as `common-sense knowledge') and reasoning abilities of LLMs can enhance performance on 3D tasks in \secref{sub:3D Task Analyzer}.
In \secref{subsec:LLM_as_3D_Multi_task_learner}, we elaborate on how to integrate multiple 3D tasks into one LLM to achieve multi-task learning. We explore how LLMs can be used as a unified interface for combining other modalities in \secref{sub:3D Multi-modal Interface}.
We then describe how LLMs serve as embodied agents to interact with the 3D world in \secref{subsec:3D Embodied Agent}.
Finally, we present how LLMs serve as assistants for generating semantically diverse 3D objects and scenes in \secref{sec:llm-generation}.

In addition, we provide Tab.~\ref{tab:taxonomy} to contrast 3D-LLMs methods across three axes: 3D components, LLMs components, and the alignment of 3D vision and language, aiming to offer a high-level insight into the various approaches within this evolving field.

\begin{table*}

\begin{adjustbox}{width=\textwidth}
\begin{tabular}{lc|ccc|ccccc|c|c}
\toprule
\textbf{}                           & \multicolumn{1}{l}{\textbf{}} & \multicolumn{3}{|c|}{\textbf{3D Component}}                                                                                               & \multicolumn{5}{c|}{\textbf{LLM Component}}                                                                                                                                                        & \textbf{3D+LLM}                                                      & \multicolumn{1}{l}{\textbf{}} \\
\midrule
\multicolumn{1}{l}{\textbf{Method}} & \textbf{Subsection} & \textbf{\begin{tabular}[c]{@{}c@{}}3D \\ Geometry\end{tabular}} & \textbf{\begin{tabular}[c]{@{}c@{}}Vision\\ Model\end{tabular}} & 
\textbf{Tuning} &\textbf{\begin{tabular}[c]{@{}c@{}}LLM \\ Abilities\end{tabular}} & \textbf{Fine-tuning} & \textbf{\begin{tabular}[c]{@{}c@{}}LLM\\ Base\end{tabular}} & \textbf{\# Parameters} & \textbf{Hardware} & \textbf{\begin{tabular}[c]{@{}c@{}}Alignment \\ Module\end{tabular}} & \textbf{Date} \\
\midrule
Chen \etal ~\cite{chen2022leveraging} & 4.2             & SG                                                                 & CLIP  & F                                                       & WK                                                              & None                 & GPT2                                                        & 1.5B                  & 1 3080            & -                                                          & 09/22         \\
ConceptFusion~\cite{jatavallabhula2023conceptfusion} & 4.2/4.4             & RGB-D                                                                 & OpenSeg   & F                                                      & IF/R                                                              & None                 & GPT3                                                        & 175B                  & 1 3090            & -                                                          & 02/23         \\
ViewRefer~\cite{guo2023viewrefer}   & 4.2                 & MVI                                                                   & Multi-View Transformer     & T                                     & WK                                                              & None                 & GPT3                                                        & 175B                  & 4 A100            & Transformer                                                          & 03/23         \\
LLM-Grounder~\cite{llmgrounder}     & 4.2                 & PC/NeRF                                                               & OpenScene/LERF      & F                                            & IF/ICL/R                                                          & None                 & GPT3.5/4                                                    & -                      & -                 & -                                                                    & 09/23         \\
Abdelreheem \etal~\cite{abdelreheem2023zero} & 4.2                 & Mesh                                                                 & Vision Transformer   & F                                           & IF/WK                                                           & None                 & GPT3.5                                                      & -                      & 1 3090            & -                                                                    & 09/23         \\
Transcribe3D~\cite{fang2023transcribe3d} & 4.2                 & PC                                                                    & Groupfree Transformer  & F                                         & IF/R                                                              & None                 & GPT3.5/4                                                    & -                      & -                 & -                                                                    & 10/23         \\
zero-shot 3DVG~\cite{yuan2023visual}        & 4.2                 & RGB-D                                                                 & Mask3d              & F                                                 & IF/ICL/R                                                          & None                 & GPT3.5/4                                                 & -                      & -                 & Transformer                                                          & 11/23         \\
3DAP~\cite{liu20233daxiesprompts}  & 4.2                 & MVI                                                                   & -         & -                                                      & IF/ICL/R                                                            & PT                  & GPT4V                                                      & -                      & -                 & -                                                                    & 12/23         \\
SpatialPIN~\cite{ma2024spatialpin}  & 4.2                 & RGB-D                                                                   & LanSAM/Depth Anything         & F                                                      & IF/R                                                            & PT                  & GPT4V                                                      & -                      & -                 & -                                                                    & 03/24         \\
SIG3D~\cite{man2024SIG3D}     & 4.2                 & PC                                                                 & OpenScene       & F                                            & IF/ICL/R                                                          & LF                  & Flan-T5                                                     & 2.7B                     & 1 A100           & MLP                                                         & 10/24         \\
\midrule
3D-LLM~\cite{3dllm}                 & 4.3                 & PC/VG                                                                    & Mask2Former/SAM      & F                                           & IF/ICL/R                                                          & LF                  & OPT/Flan-T5                                                 & 9B/2.7B/3B             & 64 V100           & QFormer                                                              & 07/23         \\
Chat-3D~\cite{chat3d}               & 4.3                 & PC                                                                    & Point-BERT          & F                                            & IF/ICL/R                                                          & AF                  & Vicuna                                                      & 7B                     & -                 & Linear layer                                                         & 08/23         \\
LEO~\cite{LEO}                      & 4.3                 & PC/MVI                                                                 & OpenClip/PointNet++     & F                                        & IF/ICL/R                                                          & LoRA                & Vicuna                                                      & 7B                     & 8 A100            & Transformer                                                          & 11/23         \\
LL3DA~\cite{chen2023ll3da}          & 4.3                 & PC/3D-BB                                                              & Vote2Cap-DETR    & F                                      & IF/ICL/R                                                            & AF                  & OPT                                                         & 1.3B                   & 8 3090            & QFormer                                                              & 11/23         \\
Point-LLM~\cite{xu2023pointllm}     & 4.3                 & PC                                                                    & Point-BERT            & F                                          & IF/ICL/R                                                          & Full                & LLaMA                                                       & 7B/13B                 & 8 A100            & Linear layer                                                         & 12/23         \\
GPT4Point~\cite{qi2023gpt4point}    & 4.3                 & PC                                                                    & Point-BERT            & F                                          & IF/ICL/R                                                          & LF                  & OPT/Flan-T5                                                 & 6.7B/3B                & 8 A100            & QFormer                                                              & 12/23         \\
Chat-3D v2~\cite{huang2023chatV2}   & 4.3                 & PC                                                                    & Uni3D             & F                                             & IF/ICL/R                                                          & AF                  & Vicuna                                                      & 7B                     & 4 A40             & MLP                                                                  & 12/23         \\
LiDAR-LLM~\cite{yang2023lidar}      & 4.3                 & PC/VG                                                              & VoxelNet             & F                                  & IF/ICL/R                                                          & AF                  & LLaMA                                                       & 7B                     & 4 A100            & Transformer                                                          & 12/23         \\
Chat-Scene~\cite{huang2024chatscene}     & 4.3                 & PC/MVI                                                                 & Uni-3D       & F                                            & IF/ICL/R                                                          & LoRA                  & Vicuna-v1.5                                                     & 7B                     & 4 A100           & MLP                                                         & 12/23         \\
3DMIT~\cite{li20243dmit}            & 4.3                 & PC                                                                    & EPCL/Uni3D          & F                                                  & IF/ICL/R                                                          & LoRA                & Vicuna                                                      & 7B                     & 8 A100            & Linear layer                                                         & 01/24         \\
ShapeLLM~\cite{qi2024shapellm}     & 4.3                 & PC                                                                 & ReCon++       & F                                            & IF/ICL/R                                                          & Full                  & Vicuna                                                     & 7B                     & 8 A800           & Linear layer                                                         & 02/24         \\
Scene-LLM~\cite{fu2024scenellm}     & 4.3                 & PC/VG                                                                 & ConceptFusion       & F                                            & IF/ICL/R                                                          & LF                  & LLaMA-2                                                     & 7B                     & 32 A100           & Linear layer                                                         & 03/24         \\
MiniGPT-3D~\cite{tang2024minigpt}     & 4.3                 & PC                                                                 & Point-BERT       & F                                            & IF/ICL/R                                                          & LoRA                  & Phi-2                                                     & 2.7B                     & 1 3090           & QFormer                                                         & 05/24         \\
LLaNA~\cite{amaduzzi2024llana}     & 4.3                 & NeRF                                                                 & nf2vec       & F                                            & IF/ICL/R                                                          & AF                  & LLaMA-2                                                     & 7B                     & 4 A100           & MLP                                                         & 06/24         \\
GreenPLM~\cite{tang2024greenPLM}     & 4.3                 & PC                                                                 & Uni3D       & F                                            & IF/ICL/R                                                          & LoRA                  & Phi-2                                                     & 2.7B                     & 1 3090           & MLP                                                         & 09/24         \\
LLaVA-3D~\cite{zhu2024llava3d}     & 4.3                 & MVI/RGB-D                                                                 & SigLIP       & F                                            & IF/ICL/R                                                          & Full                  & Qwen2                                                     & 7B                     & 16 A100           & MLP                                                         & 09/24         \\
Robin-3D~\cite{kang2025robin3d}     & 4.3                 & PC/MVI                                                                 & Uni-3D       & F                                            & IF/ICL/R                                                          & LoRA                  & Vicuna-v1.5                                                     & 7B                     & 8 A6000           & MLP                                                         & 09/24         \\
PerLA~\cite{mei2025PerLA}     & 4.3                 & PC/SG                                                                 & Vote2Cap-DETR      & F                                            & IF/ICL/R                                                          & AF                  & OPT                                                     & 1.3B                     & 2 H100           & QFormer                                                         & 11/24         \\
Video-3D LLM~\cite{zheng2025video3dllm}     & 4.3                 & MVI/RGB-D                                                                 & SigLIP       & F                                            & IF/ICL/R                                                          & Full                  & Qwen2                                                    & 7B                     & 8 A100           & Transformer                                                         & 12/24         \\
GPT4Scene~\cite{qi2025gpt4scene}     & 4.3                 & MVI                                                                 & SigLIP       & F                                            & IF/ICL/R                                                          & Full                 & Qwen2                                                     & 7B                     & 8 A100           & Transformer                                                         & 01/25         \\
LSceneLLM~\cite{zhi2025lscenellm}     & 4.3                 & PC                                                                 & OpenScene      & F                                            & IF/ICL/R                                                          & Full                  & LLaMA-2                                                     & 7B                     & 64 V100           & MLP                                                         & 02/25         \\
Inst3D-LMM~\cite{yu2025inst3d}     & 4.3                 & PC/RGB-D                                                                & Uni-3D       & F                                            & IF/ICL/R                                                          & LoRA                  & Vicuna-v1.5                                                     & 7B                     & 8 A100           & MLP                                                         & 03/25         \\
SplatTalk~\cite{thai2025splattalk}     & 4.3                 & MVI/3DGS                                                                 & SigLIP       & F                                            & IF/ICL/R                                                          & LoRA                  & Qwen2                                                     & 7B                     & 1 H100           & MLP                                                         & 03/25         \\
Ross3D~\cite{wang2025ross3d}     & 4.3                 & MVI                                                                 & SigLIP       & F                                            & IF/ICL/R                                                          & Full                  & Qwen2                                                     & 7B                     & 8 A100           & MLP                                                         & 04/25         \\
3D-LLaVA~\cite{deng20253dllave}     & 4.3                 & PC                                                                 & 3D U-Net       & F                                            & IF/ICL/R                                                          & LoRA                  & Vicuna-v1.5                                                     & 7B                     & 8 3090           & MLP                                                         & 04/25         \\
Spatial-MLLM~\cite{ma2025spatialllm}     & 4.3                 & MVI                                                                 & VGGT       & F                                            & IF/ICL/R                                                          & Full                  & Qwen2.5                                                     & 3B                     & 4 A800           & MLP                                                         & 05/25         \\
VG LLM~\cite{zheng2025VGLLM}     & 4.3                 & MVI                                                                 & VGGT       & F                                            & IF/ICL/R                                                          & Full                  & Qwen2.5                                                    & 3B                     & 8 H800           &  Linear layer                                                        & 05/25         \\
VLM-3R~\cite{fan2025vlm3r}     & 4.3                 & MVI                                                                 & CUT3R       & F                                            & IF/ICL/R                                                          & LoRA                  & Qwen2                                                    & 7B                     & 16 H200           &  MLP                                                        & 05/25         \\
LEO-VL~\cite{huang2025leovl}     & 4.3                 & MVI/RGB-D                                                                 & Qwen2.5-VL ViT       & F                                            & IF/ICL/R                                                          & LoRA                  & Qwen2.5                                                    & 7B                     & 8 A100           & MLP                                                         & 06/25         \\
3DRS~\cite{huang20253DRS}     & 4.3                 & MVI/RGB-D                                                                 & VGGT       & F                                            & IF/ICL/R                                                          & Full                  & Qwen2                                                     & 7B                     & 8 H100           & Transformer                                                         & 06/25         \\
\midrule
Point-Bind~\cite{guo2023point}      & 4.4                 & PC                                                                    & I2P-MAE           & F                                              & IF/ICL/R                                                            & LF                  & LLaMA                                                       & 7B                     & 8 A100            & Linear layer                                                         & 09/23         \\
JM3D-LLM~\cite{wang2023beyond}      & 4.4                 & PC                                                                    & PointNet++/Point-BERT    & F                              & IF/ICL/R                                                          & LF                  & Vicuna                                                      & 7B                     & 3 A100            & MLP                                                                  & 10/23         \\
MultiPLY~\cite{hong2024multiply}    & 4.3/4.4/4.5         & PC                                                                    & ConceptGraph       & T                                             & IF/ICL/R                                                          & AF                  & Vicuna                                                      & 13B                    & 128 V100          & Linear layer                                                         & 01/24         \\
3DLLM-Mem~\cite{hu20253dllm-mem} & 4.4                & MVI/RGB-D                                                                 & SigLIP       & F                                            & IF/ICL/R                                                          & Full                  & Qwen2                                                     & 7B                     & 8 TPU v5p           & MLP                                                                  & 05/25         \\
\midrule
LLM-Planner~\cite{song2023llmplanner} & 4.5                 & VG                                                                    & Seg\&Depth           & F                                           & IF/ICL/R                                                          & None                 & GPT3                                                        & 175B                  & -                 & -                                                                    & 03/23         \\
SayPlan~\cite{rana2023sayplan}      & 4.5                 & SG                                                                    & SG generator        & F                                            & IF/ICL/R                                                          & None                 & GPT4                                                        & -                      & -                 & -                                                                    & 07/23         \\
VoxPoser~\cite{huang2023voxposer}   & 4.5                 & RGB-D                                                                 & OWL-ViT/SAM         & F                                            & IF/ICL/R                                                          & None                 & GPT4                                                        & -                      & -                 & -                                                                    & 07/23         \\
UniHSI~\cite{xiao2023unified}       & 4.5                 & RGB-D                                                                 & -               & -                                                & IF/ICL/R                                                          & None                 & GPT3.5/4                                                    & -                      & 1 A100            & -                                                                    & 09/23         \\
LAN-grasp~\cite{mirjalili2023lan}   & 4.5                 & MVI                                                                   & OWL-ViT         & F                                                & IF/ICL/R                                                          & None                 & GPT4                                                        & -                      & -                 & -                                                                    & 10/23         \\
Agent3D-Zero~\cite{zhang2024agent3dzero} & 4.3/4.5                 & MVI                                                                   & -      & -                                                         & IF/ICL/R                                                          & PT                 & GPT4V                                                      & -                      & -                 & -                                                                    & 03/24         \\
NaviLLM~\cite{zheng2023towards}     & 4.5                 & MVI                                                                   & EVA-CLIP-Large     & F                                             & IF/ICL/R                                                          & LF                  & Vicuna                                                      & 7B                     & 8 A100            & Transformer                                                          & 12/23         \\
ManipLLM~\cite{li2023manipllm}      & 4.5                 & RGB-D                                                                 & CLIP               & F                                             & IF/ICL/R                                                          & AF                  & LLaMA                                                       & 7B                     & 1 A100            & Linear layer                                                         & 03/24         \\
3D-VLA~\cite{zhen20243d}            & 4.3/4.5                 & PC/MVI                                                                & Mask2Former/SAM      & F                                           & IF/ICL/R                                                          & LoRA                & Flan-T5                                                     & 3B                     & 384 V100          & QFormer                                                              & 03/24         \\
\midrule
PolyGen~\cite{nash2020polygen}       & 4.6                 & \textit{n-gon} Mesh                                                   & Custom Transformer     & F                                         & -                                                                 & Full                & Custom                                                      & -                      & 4 V100            & -                                                                    & 02/20         \\
LLMR~\cite{de2023llmr}              & 4.6                 & Unity, Mesh                                                           & Dall-E-2, CLIP        & F                                          & IF/ICL/R                                                          & None                 & GPT4                                                        & -                      & 1 3080            & -                                                                    & 09/23         \\
3D-GPT~\cite{3dgpt}                 & 4.6                 & Blender, Mesh                                                         & -                & -                                               & IF/R                                                              & None                 & GPT3.5/4                                                    & -                      & -                 & -                                                                    & 10/23         \\
MeshGPT~\cite{siddiqui2023meshgpt}   & 4.6                 & Mesh                                                                  & GPT-2-style Transformer    & F                                     & -                                                                 & Full                & GPT-2-style                                                 & $\sim$345M             & 4 A100            & -                                                                    & 11/23         \\
ShapeGPT~\cite{shapegpt}             & 4.6                 & SDF                                                                   & 3D VQ-VAE, CLIP       & F                                          & IF/R                                                              & Full                & T5                                                          & -                      & 4 A100            & T5                                                                   & 11/23         \\
Holodeck~\cite{yang2024holodeck}    & 4.6                 & Mesh                                                                  & -         & -                                                      & IF/R                                                              & None                 & GPT4                                                       & -                      & 8 RTX 8k          & CLIP/SBERT                                                           & 12/23         \\
GALA-3D~\cite{gala3d}               & 4.6                 & 3DGS                                                                  & MVDream, Control-Net     & F                                       & IF/R                                                              & None                 & Any (GPT3.5)                                               & -                      & -                 & -                                                                    & 02/24         \\
LLaMA-Mesh~\cite{wang2024llamamesh}               & 4.6                 & Mesh                                                                  & -     & F                                       & IF/R                                                              & Full                 & LLaMA-3.1                                              & 8B                      & 32 A100                 & MLP                                                                    & 11/24         \\
\bottomrule
\end{tabular}
\end{adjustbox}
    \vspace{-1mm}
\caption{\textbf{Summarization of 3D-LLMs methods.} The \textit{3D Geometry} column collects the 3D geometric information being used for each method, such as Point Cloud (PC), Multi-view Images (MVI), RGB+Depth (RGB-D), 3D Bounding Box (3D-BB), Scene Graph (SG), Voxel Grid (VG), Gaussian Splatting (3DGS) and NeRF. The tuning column outlines whether the vision model is finetuned during training (True/False).
IF, ICL, R, and WK denote the \textit{LLM Abilities} of Instruction-following, In-Context Learning, Reasoning, and World Knowledge. The \textit{Fine-tuning} column summarizes how LLM components are fine-tuned, such as Prompt Tuning (PT), Low-Rank Adaptation (LoRA), Adaptive Fine-tuning (AF), Layer Freezing (LF), or Full Fine-tuning (Full). The \textit{Hardware} column shows methods that are trained either with numbers of Nvidia GPU and the specific GPU type or with no training involved.
}~
\vspace{-4mm}
\label{tab:taxonomy}
\end{table*}

\subsection{How do LLMs process 3D scene information?} \label{sec:3.1.1}

Traditional LLMs are limited to text as both input and output, making the ability to ingest 3D information a primary concern for all 3D-LLM methods. The general idea is to map 3D objects or scene information into the language space, enabling LLMs to understand and process these 3D inputs. Specifically, this typically involves two steps: (i) using a pre-trained 3D encoder to process the corresponding 3D representations, yielding original 3D features; (ii) employing an alignment module to translate these 3D features into 3D tokens that LLMs can process, akin to the tokenization process mentioned in \secref{sec:LLM_101}. Pre-trained LLMs can then use these aligned 3D tokens when generating outputs.

Given the diversity of 3D representations, as described in~\secref{sec:3d_representations}, there are multiple ways to obtain 3D features. As shown in 3D Geometry column in Tab.~\ref{tab:taxonomy}, \textit{Point Clouds}~\cite{llmgrounder,hong2024multiply,chen2023ll3da,chat3d,huang2023chatV2,xu2023pointllm,qi2023gpt4point,li20243dmit,3dllm,LEO,yang2023lidar,fu2024scenellm,guo2023point} are most common due to their simplicity and compatibility with various pre-trained 3D encoders, which makes them a popular choice for multi-task and multi-modal learning methods. \textit{Multi-view Images}~\cite{zheng2023towards,mirjalili2023lan,zhang2024agent3dzero,guo2023viewrefer,LEO,liu20233daxiesprompts} are also frequently used as research on 2D feature extraction is well-established, which means that 3D feature extraction requires only an additional 2D to 3D lifting scheme. 
\textit{RGB-D} data, readily obtained using depth cameras, is commonly used in 3D embodied agent systems to extract viewpoint-relevant information for navigation and understanding~\cite{zhu2024llava3d,zheng2025video3dllm,yu2025inst3d,huang2025leovl,huang20253DRS}. 
\textit{3D Scene Graphs} provide more abstract 3D representation, excelling in modeling the presence of objects and their relationship, and capturing high-level information of scenes. 
They are often utilized in 3D scene classification~\cite{chen2022leveraging},  understanding~\cite{mei2025PerLA} and planning tasks~\cite{rana2023sayplan}. \textit{NeRFs} are currently less used in 3D-LLM methods~\cite{llmgrounder,amaduzzi2024llana}. We believe this is due to their implicit nature which makes them more difficult to tokenize and integrate with feed-forward neural networks.

Current methods use different architectures (see Fig.~\ref{fig:architectures}) and modules that align 3D features with LLM input spaces (see 3D+LLM column in Tab.~\ref{tab:taxonomy}).
For models which only accept 3D inputs (Fig.~\ref{fig:architectures}a), a linear layer~\cite{li20243dmit,hong2024multiply,fu2024scenellm} or an MLP~\cite{chat3d,huang2023chatV2} is used as the alignment module to transform 3D features into the LLM input space.
Models which accept 3D and text as input often use two separate branches to align 3D features and text (Fig.~\ref{fig:architectures}b).
Some works~\cite{chat3d,huang2023chatV2} employ a one-layer vanilla transformer to allow 3D object features to attend to each other during alignment.
Others, such as~\cite{LEO,yang2023lidar}, create transformer-based alignment modules, where the standard transformer architecture is tweaked to better suit the different types of 3D data, such as dense point clouds and sparse LiDAR scans. Meanwhile, text is encoded using the pre-existing LLM text embedding table.
Other works~\cite{3dllm,qi2023gpt4point,zhen20243d} follow the Q-Former style approach of~\cite{li2023blip2} to align 3D features and text (Fig.~\ref{fig:architectures}c), introducing fixed length learnable query tokens as additional input, and following a BERT-based structure to facilitate interaction between 3D and text features during alignment.
Mostly, these three types of architectures described above achieve alignment by utilizing 3D captioning datasets~\cite{scanrefer}, where the captioning loss, \ie, the cross-entropy loss between captions generated by LLMs and the brief, ground truth description of the scene, is used to fine-tune the alignment module, while freezing the pre-trained 3D feature extractor and LLM.

Finally, some models~\cite{llmgrounder,guo2023viewrefer,abdelreheem2023zero,fang2023transcribe3d,liu20233daxiesprompts,song2023llmplanner,rana2023sayplan,xiao2023unified,mirjalili2023lan,zhang2024agent3dzero} employ closed-source models like ChatGPT and do not train an alignment module at all (Fig.~\ref{fig:architectures}d). Instead of aligning 3D features with the LLM input space, text descriptions are directly generated from the 3D data, such as by describing 3D bounding boxes, positions and relationships, or by using pre-existing captions. These text descriptions are input into ChatGPT. No additional alignment module is proposed in these works, and thus no training is required.

\subsection{LLMs for Enhancing 3D Task Performance}
\label{sub:3D Task Analyzer}

LLMs trained on large amounts of data have been shown to acquire common-sense knowledge about the world~\cite{li-etal-2022-systematic}.
The potential of LLMs' world knowledge and reasoning capabilities has been explored to enhance 3D scene understanding and reformulate the pipelines of several 3D tasks. In this section, we focus on methods which aim to use LLMs to improve the performance of existing methods on 3D vision-language tasks. 
When applying LLMs to 3D tasks, we can categorize their use into two distinct groups: \textit{knowledge-enhanced} and \textit{reasoning-enhanced} approaches. Knowledge-enhanced approaches tap into the vast world knowledge embedded within LLMs to enhance 3D task performance. This may provide contextual insights, filling in knowledge gaps, or might enhance the semantic understanding of 3D environments. Alternatively, rather than relying on their world knowledge, reasoning-enhanced approaches leverage the ability of LLMs to infer step-by-step, thereby offering better generalization to tackle more complex 3D challenges.
The following two sections describe these approaches respectively.

\subsubsection{Knowledge-enhanced approaches} 
\label{subsec:knowledge-enhanced}
There are several methods that utilize LLM world knowledge.
Chen \etal~\cite{chen2022leveraging} use LLMs for 3D room classification from RGB-D images.
Here, the knowledge embedded in LLMs is used to determine the room category based on the object category information contained in the room.
Firstly, this approach creates scene graphs from Matterport3D~\cite{Matterport3D} data with nodes for regions and objects, with object nodes linked to room nodes.
Next, key objects are selected to form queries per room type. LLMs score descriptions extracted from the selected objects, with the highest score predicting the room label. 

ViewRefer~\cite{guo2023viewrefer} uses LLMs to expand the grounding texts with view-related descriptions. For instance, given the original text ``Facing the front of the couch, the table that is to the right of the couch'', the LLM is used to create a similar sentence but from another speaker's point of view, \eg ``With back to the front of the couch, pick the table that is on the left side of the couch''. By rephrasing input texts and their opposite-view synonyms several times, the model improves cross-view grounding. 
Abdelreheem \etal~\cite{abdelreheem2023zero} address the problem of semantic correspondence in 3D shapes by using a BLIP2 model to generate lists of text class proposals from different rendered views of a 3D shape. ChatGPT~\cite{chatgpt} then unifies these proposals into a single class proposal for the entire object.

The knowledge-enhanced strategies described above enable strong performance, especially in zero-shot scenarios where no labeled 3D data is available for a particular object or scene type. This allows open-ended reasoning about object parts, relationships, and semantics beyond fixed ontologies, as demonstrated by (i) Chen \etal~\cite{chen2022leveraging} generating spatial and semantic object descriptions, (ii) ViewRefer ~\cite{guo2023viewrefer} describing multi-view object relationships, and (iii) Abdelreheem \etal~\cite{abdelreheem2023zero} generating and matching object part semantics across shapes.

\subsubsection{Reasoning-enhanced approaches}
\label{subsec:Reasoning-enhanced}
Along with world knowledge, the reasoning abilities of LLMs also help tackle other 3D tasks, particularly visual grounding in complicated 3D scenes with detailed geometry and multiple objects. In such cases, text descriptions of the objects should include their appearance and spatial relationship with surrounding items.
Ordinary grounding methods~\cite{he2021transrefer3d} often struggle in this setting due to their inability to understand detailed text descriptions.
LLM-Grounder~\cite{llmgrounder}, Transcribe3D~\cite{fang2023transcribe3d}, and zero-shot 3DVG~\cite{yuan2023visual} approach this problem by leveraging the reasoning capability of LLMs to analyze the text descriptions and generate a sequence of instructions for locating the objects using existing grounding toolboxes. Specifically, an LLM first identifies anchor and target objects from the text descriptions. It then analyzes the spatial relationship (or described attributes) between multiple candidate objects based on their coordinates returned by the grounding tool to select the candidate that best matches the text description.

A common drawback of these approaches is the ``blindness'' of the LLM, since it is provided only with abstract textual descriptions of the 3D scene instead of the original point cloud. This may lead to the loss of crucial scene details. Consequently, when a 3D scene contains multiple objects of the same class, the absence of necessary scene details means that ambiguities of the text-based references cannot be resolved, which limits overall performance.

Besides visual grounding, the reasoning capability of LLMs can facilitate other tasks. 3DAP~\cite{liu20233daxiesprompts} leverages GPT-4V to infer 3D information of an object from its 2D images using a visual prompting technique, where it annotates the input image with a 3D axis to enhance the LLM's awareness of the 3D scale. ConceptFusion~\cite{jatavallabhula2023conceptfusion} uses GPT3 to generate instructions using pre-defined elementary spatial comparison modules to enable more complex spatial reasoning using their proposed 3D feature map. 
SpatialPIN~\cite{ma2024spatialpin} utilizes GPT-4V to infer interactive objects, potential tasks and plan of actions from a single RGB image. 

\subsection{LLMs for 3D Multi-Task Learning} 
\label{subsec:LLM_as_3D_Multi_task_learner}

\begin{table*}[ht!]
\renewcommand{\arraystretch}{0.75} 
\centering
\resizebox{\textwidth}{!}{
\begin{tabular}{lcccccccccccc}
\toprule
\multirow{2}{*}{Method} & \multicolumn{3}{c}{3D Captioning} & \multicolumn{2}{c}{3D Grounding} & \multicolumn{3}{c}{3D Conversation} & \multicolumn{3}{c}{3D Embodied Agents} & \multicolumn{1}{c}{3D Generation} \\
\cmidrule(lr){2-4} \cmidrule(lr){5-6} \cmidrule(lr){7-9} \cmidrule(lr){10-12} \cmidrule(lr){13-13}
 & \begin{tabular}[c]{@{}c@{}}Object\end{tabular} & \begin{tabular}[c]{@{}c@{}}Scene \end{tabular} & \begin{tabular}[c]{@{}c@{}}Dense\end{tabular} & \begin{tabular}[c]{@{}c@{}}Single\end{tabular} & \begin{tabular}[c]{@{}c@{}}Multi \end{tabular} & QA & Situated-QA & Dialogue & Plan. & Nav. & Manip. & object\\
\midrule
\rowcolor[HTML]{EFEFEF}
3D-LLM~\cite{3dllm}                     & \checkmark     & \checkmark    &              & \checkmark      & \checkmark & \checkmark  &            & \checkmark & \checkmark  & \checkmark   &            &            \\
Chat-3D~\cite{chat3d}                   & \checkmark     &               &              &                 &            & \checkmark  &            &            &             &              &            &            \\
\rowcolor[HTML]{EFEFEF}
LEO~\cite{LEO}                          & \checkmark     & \checkmark    &              &                 &            & \checkmark  & \checkmark & \checkmark & \checkmark  & \checkmark   & \checkmark &            \\
LL3DA~\cite{chen2023ll3da}              & \checkmark     & \checkmark    & \checkmark   &                 &            & \checkmark  &            & \checkmark & \checkmark  &              &            &            \\
\rowcolor[HTML]{EFEFEF}
PointLLM~\cite{xu2023pointllm}          & \checkmark     &               &              &                 &            & \checkmark  &            & \checkmark &             &              &            &            \\
GPT4Point~\cite{qi2023gpt4point}        & \checkmark     &               &              &                 &            & \checkmark  &            & \checkmark &             &              &            & \checkmark \\
\rowcolor[HTML]{EFEFEF}
Chat-3D V2~\cite{huang2023chatV2}       & \checkmark     & \checkmark    & \checkmark   & \checkmark      & \checkmark & \checkmark  &            &            &             &              &            &            \\
LiDAR-LLM~\cite{yang2023lidar}          & \checkmark     & \checkmark    &              & \checkmark      & \checkmark & \checkmark  & \checkmark &            & \checkmark  & \checkmark   &            &            \\
\rowcolor[HTML]{EFEFEF}
3DMIT~\cite{li20243dmit}                & \checkmark     & \checkmark    &              & \checkmark      & \checkmark & \checkmark  &            & \checkmark &             &              &            &            \\
MultiPLY~\cite{hong2024multiply}        & \checkmark     &               &              &                 &            & \checkmark  & \checkmark & \checkmark &             & \checkmark   &            &            \\
\rowcolor[HTML]{EFEFEF}
3D-VLA~\cite{zhen20243d}                & \checkmark     &               & \checkmark   & \checkmark      & \checkmark & \checkmark  & \checkmark &            & \checkmark  &              & \checkmark &            \\
Agent3D-Zero~\cite{zhang2024agent3dzero}&                & \checkmark    &              & \checkmark      & \checkmark & \checkmark  &            & \checkmark & \checkmark  & \checkmark   & \checkmark &            \\
\rowcolor[HTML]{EFEFEF}
ShapeLLM~\cite{qi2024shapellm}         & \checkmark     &   \checkmark            &    &    \checkmark  & \checkmark   & \checkmark  &            & \checkmark & \checkmark  &              &            &            \\
Scene-LLM~\cite{fu2024scenellm}         & \checkmark     &               & \checkmark   &                 &            & \checkmark  &            & \checkmark & \checkmark  &              &            &            \\
\rowcolor[HTML]{EFEFEF}
\rowcolor[HTML]{EFEFEF}
Spartun3D-LLM~\cite{zhang2024spartun3d} &      &               &              &                 &            & \checkmark  & \checkmark   &  &   \checkmark    &    \checkmark    &            &            \\
Chat-Scene~\cite{wang2023beyond}         & \checkmark     & \checkmark    & \checkmark   & \checkmark      & \checkmark & \checkmark  & \checkmark & \checkmark & \checkmark &             &            &            \\
\rowcolor[HTML]{EFEFEF}
LLaVA-3D~\cite{zhu2024llava3d}         & \checkmark     & \checkmark    & \checkmark   & \checkmark      & \checkmark & \checkmark  & \checkmark & \checkmark & \checkmark  &             &            &            \\
Robin-3D~\cite{kang2025robin3d}         & \checkmark     & \checkmark    & \checkmark   & \checkmark      & \checkmark & \checkmark  & \checkmark & \checkmark &              &             &            &            \\
\rowcolor[HTML]{EFEFEF}
Video-3D LLM~\cite{zheng2025video3dllm}         & \checkmark     & \checkmark    & \checkmark   & \checkmark      & \checkmark & \checkmark  & \checkmark & \checkmark &              &             &            &            \\
GPT4Scene~\cite{qi2025gpt4scene}         & \checkmark     & \checkmark    & \checkmark   & \checkmark      & \checkmark & \checkmark  & \checkmark & \checkmark &              & \checkmark    &            &            \\
\rowcolor[HTML]{EFEFEF}
LSceneLLM~\cite{zhi2025lscenellm}         & \checkmark     & \checkmark    & \checkmark   &      &  & \checkmark  & \checkmark & \checkmark & \checkmark  &     &            &            \\
Ross3D~\cite{wang2025ross3d}         & \checkmark     & \checkmark    & \checkmark   & \checkmark      & \checkmark & \checkmark  & \checkmark & \checkmark &              &             &            &            \\
\rowcolor[HTML]{EFEFEF}
3D-LLaVA~\cite{deng20253dllave}         & \checkmark     & \checkmark    & \checkmark   & \checkmark      &  & \checkmark  & \checkmark & \checkmark &              &             &            &            \\
VG-LLM~\cite{zheng2025VGLLM}         & \checkmark     & \checkmark    & \checkmark   & \checkmark      &  & \checkmark  & \checkmark & \checkmark &              &             &            &            \\
\rowcolor[HTML]{EFEFEF}
3DRS~\cite{huang20253DRS}         & \checkmark     & \checkmark    & \checkmark   & \checkmark      & \checkmark & \checkmark  & \checkmark & \checkmark &              & \checkmark    &            &            \\
\bottomrule
\end{tabular}
}
    \vspace{-1mm}
\caption{\textbf{Comparison of tasks} included in various 3D multi-task learning methods. QA stands for Question Answering. The methods are sorted chronologically from top (oldest) to bottom (most recent).
}
    \vspace{-3mm}
\label{tab:multitask}
\end{table*}

Many works focus on using the instruction-following and in-context learning capabilities of LLMs to unify multiple 3D tasks into a single language space. By using different text prompts to denote different tasks, these works allow for LLMs to serve as a unifying dialogue interface. Achieving multi-task learning using an LLM typically involves several key steps, beginning with the construction of 3D-text data pairs~\cite{chen2023ll3da,3dllm,LEO}. 
These pairs require crafting task instructions in text form and defining the output for each different task. 
Next, 3D data (typically in the form of point clouds) are fed into 3D encoders~\cite{yu2022point,qi2017pointnet++} to extract 3D features. 
Alignment modules~\cite{chat3d,3dllm,huang2023chatV2,fu2024scenellm} are subsequently used to (i) align 3D features with text embeddings from the LLMs across multiple levels (object-level, relationship-level, and scene-level) and (ii) translate the 3D features into tokens interpretable by the LLMs.
Finally, an appropriate training strategy~\cite{chen2023ll3da,3dllm,guo2023point,LEO,li20243dmit,fei2024kestrel} needs to be selected, such as single or multi-stage 3D-language alignment training and multi-task instruction fine-tuning.
In the remainder of this section, we explore these aspects in detail. We additionally summarize the scope and capabilities of each method reviewed in 
this section in Table~\ref{tab:multitask}.

\subsubsection{Data for Multi-Task Learning}

As shown in Fig.~\ref{fig:tasks_overview}, we divide tasks into five categories: captioning, grounding, question answering (QA), embodied agent tasks (\textit{i.e.}, planning, navigation, and manipulation), and generation. Accordingly, each task's textual output follows a predefined format. For captioning and QA tasks, the output is plain text and is not constrained by a specific format. The output of the grounding task is a 3D bounding box, typically the coordinates of the referred object's center along with its 3D size. Normally, the values for point and size are normalized to fall within the range of 0-255~\cite{chen2023ll3da} which limits the range of tokens the LLM is required to predict.
For planning, the model outputs a series of steps to execute a task in text form, whereas for navigation, the output is a series of spatial coordinates. For manipulation, the output is action sequences in text form. 
Existing methods follow these guidelines to construct their multi-task instruction fine-tuning datasets.

Once the text format has been decided upon, different works use different strategies to obtain text annotations for their datasets, constructing 3D-text pairs for training. Several approaches ask human labelers to generate `ground truth' annotations for each sample~\cite{scanrefer, achlioptas2020referit3d, scanqa, ma2022sqa3d}. However this can be an expensive and time-consuming process.
Another approach~\cite{li20243dmit, yang2023lidar, chat3d, qi2024shapellm, huang2023chatV2} is to use an LLM (\eg ChatGPT~\cite{chatgpt}) to generate text annotations for each sample. 
Here, 3D scene data is converted into text (usually by describing object bounding boxes and spatial relations textually), and a task description is created to describe the desired output. To guide ChatGPT towards the expected output format for the task, demonstration examples are provided, which allows ChatGPT to perform in-context learning to generate plausible text annotations for other 3D scenes. 
Alternatively, other multi-task datasets~\cite{chen2023ll3da,fu2024scenellm} are constructed by merging existing 3D Vision-Language (VL) datasets~\cite{scanrefer,achlioptas2020referit3d,scanqa,3dllm}.
Several recent multi-task datasets adopt a hybrid strategy that combines these three approaches~\cite{LEO, yang2023lidar, 3dllm}, aiming to leverage both the accuracy of human annotations and the scalability of LLM-generated ones.

\subsubsection{3D Feature Tokenization for LLM training}


The first step in training an LLM for multi-task 3D understanding lies in extracting semantically and spatially meaningful 3D features. We group existing methods into two main categories depending on their input modality and spatial encoding strategies.

\textbf{Point-based methods} operate directly on point clouds, offering explicit spatial grounding and high geometric precision. They can be further divided into: (1) \textit{Scene-centric} models~\cite{3dllm, zhu2024llava3d, chen2023ll3da, qi2023gpt4point, li20243dmit, qi2024shapellm, xu2023pointllm}, which process raw scene point clouds using pretrained or custom point encoders such as Point-BERT~\cite{yu2022point} to obtain the 3D scene features, then alignment module (such as MLP) is used to project 3D features to 3D tokens; 
(2) \textit{Object-centric} models~\cite{LEO, chat3d, huang2023chatV2, huang2024chatscene, yu2025inst3d, kang2025robin3d, qi2024shapellm, xu2023pointllm,tang2024minigpt,tang2024greenPLM}, instead of processing the whole 3D scenes, each 3D object is segmented via 3D segmentation, such as Mask3D~\cite{takmaz2023openmask3d} and then encoded to 3D object feature. Additional 3D positional embeddings are incorporated into each object feature to generate the final 3D tokens.
While effective at modeling discrete entities, they often lack comprehensive scene-level spatial understanding; and 
(3) \textit{Voxel-centric} methods~\cite{fu2024scenellm, yang2023lidar}, which discretize the 3D space into voxel grids and apply an encoder such as VoxelNet~\cite{zhou2018voxelnet} to get 3d voxel feature. Despite the advantage of maintaining spatial structures and metric accuracy, these methods generally suffer from limited semantic richness due to annotation scarcity and the sparsity of 3D data. Moreover, due to the relatively small scale of existing point cloud datasets, 3D-LLMs trained solely on point-based representations typically underperform their 2D counterparts in terms of general-purpose semantic understanding~\cite{jin2025revisiting}.

\textbf{Multi-view image-based methods} directly encode multi-view images of a 3D scene using a 2D image encoder, and then lift the 2D image features into 3D space to obtain 3D tokens. The advantage of this paradigm is that it combines the semantic richness and well-aligned text-image representation of large-scale 2D vision-language models (VLMs) with 3D spatial reasoning.
These can be categorized as: 
(1) \textit{Projection-based methods} such as LEO-VL~\cite{huang2025leovl}, which project 2D pixel features from 2D images into 3D coordinates using calibrated camera intrinsics, thus obtaining a feature for each 3D point. By further embedding the XYZ coordinates through 3D positional encoding, the final 3D tokens are generated; 
(2) \textit{Reconstruction-based methods}~\cite{thai2025splattalk,amaduzzi2024llana}, which first reconstruct a 3D scene (e.g., via 3D gaussian splatting or NeRF) and then align it with language; and 
(3) \textit{Geometry-augmented methods}~\cite{zheng2025VGLLM, ma2025spatialllm, fan2025vlm3r}, which use VGGT~\cite{wang2025vggt}-like encoder to generate both geometry-aware and semantic representations without explicit projection or reconstruction. 
These approaches benefit from pretrained VLMs and rich 2D image-text datasets, yet the additional lifting step can introduce projection errors and lead to imprecise spatial relationships in the constructed 3D space.
(4) \textit{Temporal-augmented methods}~\cite{zhu2024llava3d,zheng2025video3dllm,qi2025gpt4scene,wang2025ross3d,huang20253DRS} repurpose video streams as implicit 3D observations, leveraging temporal continuity as a proxy for spatial structure. these methods feed video frames to existing 2D-VLMs, without architectural modifications, directly inherit their strong language grounding and reasoning capabilities. Additionally, videos are widely available, and adjacent frames are naturally correlated in both time and space, implicitly encoding 3D geometry through motion. This category currently yields the best empirical performance among 3D-LLMs due to the scalability and semantic richness of video-text data, as shown in Table~\ref{tab:benchmarks}.


\subsubsection{Training Strategy for Multi-Task Learning}

LLMs can be fine-tuned to incorporate multiple 3D tasks using different strategies discussed in Sec.~\ref{sec:LLM_fine-tuning}.
LEO~\cite{LEO}, MiniGPT-3D~\cite{tang2024minigpt}, and 3DMIT~\cite{li20243dmit} use low-rank adaptation (LoRA) for fine-tuning. As a result, the total trainable parameters, including those of alignment modules and 3D encoders, amount to less than 10\% of the original LLMs' parameters, significantly enhancing training efficiency.
Chat-3D~\cite{chat3d}, LL3DA~\cite{chen2023ll3da}, Chat-3D v2~\cite{huang2023chatV2}, LiDAR-LLM~\cite{yang2023lidar}, and MultiPLY~\cite{hong2024multiply} adopt adaptive fine-tuning. Specifically, these models include modules that align spatial information in 3D scenes with language, \eg a transformer layer, to capture object relationships. These modules, along with the pre-trained 3D encoders and LLMs, are fine-tuned for alignment.
3D-LLM~\cite{3dllm}, Scene-LLM~\cite{fu2024scenellm}, Point-LLM~\cite{xu2023pointllm}, and GPT4Point~\cite{qi2023gpt4point} employ layer freezing. By freezing most LLM layers and fine-tuning certain layers like the embedding layer, this strategy preserves linguistic capabilities while improving 3D understanding.
Lastly, Agent3D-Zero~\cite{zhang2024agent3dzero} uses prompt tuning, a training-free method for guiding LLMs toward 3D task comprehension. This approach utilizes tailored prompts, 
adding grid lines and tick marks onto a BEV image of 3D scene, which help the 2D VLMs understand the 3D geometry.

Training these models for 3D multi-task learning involves fine-tuning to achieve \textit{3D-language feature alignment}.
A line of work adopts single-stage alignment methods~\cite{xu2023pointllm,3dllm,fu2024scenellm,LEO,qi2023gpt4point}.
Specifically, Point-LLM~\cite{xu2023pointllm} trains an MLP using only captioning data and additionally updates the input embedding layer to accommodate newly added tokens marking the start and end of point cloud tokens with ($\langle$p\_start$\rangle$, $\langle$p\_end$\rangle$).
3D-LLM~\cite{3dllm} uses a custom dataset to train the alignment module along with the input and output embedding layers updating weights for newly added location tokens.
Scene-LLM~\cite{fu2024scenellm} trains only a linear layer to enable LLMs to understand both ego-centric and scene-centric perspectives using 3D frame-language pair caption tasks in camera and world coordinate systems. It also updates the input embedding layer to accommodate newly added tokens marking the start and end of 3D tokens ($\langle$3D$\rangle$, $\langle$/3D$\rangle$).
LEO~\cite{LEO} also trains the alignment modules using captioning tasks, but uniquely collects three types of captioning data: object-level~\cite{objaverse}, object-in-the-scene~\cite{achlioptas2020referit3d,zhu20233d}, and scene-level~\cite{Wald2019RIO}, training its alignment module with all three datasets.
GPT4Point~\cite{qi2023gpt4point} follows the structure and training strategy of BLIP2~\cite{li2023blip2}, achieving alignment through three tasks: Point-Text Contrast (PTC), Point-Text Matching (PTM), and Point Caption Generation (PTG).

By contrast, other works~\cite{chat3d,huang2023chatV2,yang2023lidar,tang2024minigpt} employ a multi-stage alignment process, with each stage designed to enhance different aspects of the LLM's 3D capabilities.
LiDAR-LLM~\cite{yang2023lidar} focuses on enhancing both local and global scene perception through 3D captioning tasks in two phases: first, concentrating on single-view captions and then expanding to panoramic scene descriptions. They develop instance-level perception capabilities through a blend of captioning and grounding tasks. 
Chat-3D~\cite{chat3d} first aligns 3D objects with text using datasets for 3D object classification~\cite{chang2015shapenet,objaverse,uy-scanobjectnn-iccv19}, aiming to maximize the cosine similarity between mapped 3D object features and object category word embeddings. In the second stage, it utilizes ScanRefer~\cite{scanrefer} to enable captioning capabilities, updating an additional transformer layer specifically to model spatial relationships of objects. 
Similarly, Chat-3D v2~\cite{huang2023chatV2} incorporates object-level and scene-level alignments, with the second stage training an additional position embedding layer.
MiniGPT-3D~\cite{tang2024minigpt} additionally fine-tunes a `Mixture of Query' module to enhance their model's 3D perception ability from multiple perspectives. GreenPLM~\cite{tang2024greenPLM} propose to only align text caption of 3D data to LLM at the first two stages and then achieve object-level alignment by tuning the MLP layer.
For training efficiency, both LL3DA~\cite{chen2023ll3da} and 3DMIT~\cite{li20243dmit} skip the alignment stages and focus solely on the stage of instruction tuning described below.

Nearly all multi-task learning methods ultimately require the ability to complete various 3D tasks based on instructions. Thus, as the final stage of training, each method typically uses their own constructed multitask instruction-following dataset for instruction fine-tuning\footnote{Instruction fine-tuning refers to the process of further fine-tuning LLMs on a dataset consisting of (INSTRUCTION, OUTPUT) pairs~\cite{zhang2023instruction}}.
Since all task outputs are unified into textual form, the training loss used is the standard auto-regressive loss used in LLMs, except that Ross3D~\cite{wang2025ross3d} uses an additional reconstructive loss. This stage usually involves jointly training the alignment module and the LLM.
An exception is Agent3D-Zero~\cite{zhang2024agent3dzero}, which completes various 3D tasks by feeding GPT-4V with 2D images from different viewpoints, thus it does not require any training.

\subsection{LLMs as 3D Multi-Modal Interfaces} 
\label{sub:3D Multi-modal Interface}

In addition to exploring 3D multi-task learners, some recent studies incorporate information across different modalities to further improve the capability of the models and enable novel interactions. Apart from text and 3D scenes, multi-modal 3D-LLMs may also include 2D images, audio, or touch information in scenes as inputs. 

Most works aim to construct a common representation space across different modalities. Since several existing works~\cite{beaumont2022clip, guzhov2022audioclip} have already provided pre-trained encoders that map text, images, or audio to common spaces, some works choose to learn a 3D encoder that aligns 3D embeddings to the embedding space of pre-trained encoders for other modalities. JM3D-LLM~\cite{wang2023beyond} learns a 3D point cloud encoder that aligns the embedding space of the point cloud to the text-image embedding space of SLIP~\cite{mu2022slip}. It renders a sequence of images of the point cloud and constructs a hierarchical text tree during training to enable a detailed alignment. 
Point-Bind~\cite{guo2023point} also learns a similar 3D encoder and aligns it to ImageBind~\cite{girdhar2023imagebind} to unify the embedding space of images, text, audio, and point clouds.  
This enables tackling different tasks such as retrieval, classification, and generation among the various modalities using different task heads. However, one notable limitation is that this approach only works on small-scale object-level scenes as it is computationally expensive for a 3D encoder to process a large scene with millions of points. Moreover, most pre-trained multi-modal encoders like CLIP are designed for single-object scenes and are not suitable for large scenes with multiple objects and local details.

Large scenes instead require more meticulous design to incorporate multiple modalities. 
ConceptFusion~\cite{jatavallabhula2023conceptfusion} constructs an enhanced feature map that fuses both global information and local details for each constituent image of a large scene. This is achieved by using pre-trained feature extractors~\cite{li2022lseg, ghiasi2021openseg} that have already been aligned to different modalities including text and audio. It then maps the feature maps to the point cloud of the scene using traditional SLAM approaches. 
MultiPLY~\cite{hong2024multiply} adopts a similar representation to ConceptGraph~\cite{gu2023conceptgraphs}. It identifies all salient objects in the scene, obtains the global embedding for each object, and finally constructs a scene graph. The resultant representation is a scene embedding aligned with the embedding space of Llama~\cite{touvron2023llama}. The embeddings of other modalities including audio, temperature, and sense of touch can also be mapped to the same space using linear projection. All embeddings are tokenized and sent to the LLM at once.
Compared with methods on object-level scenes, approaches that can process large scenes reduce costs by relying on pre-trained encoders to bridge the modality gap, instead of learning new ones from scratch.

\subsection{LLMs for 3D Embodied Agents}
\label{subsec:3D Embodied Agent}

3D embodied agents can be created using the planning, tool-using, and decision-making abilities of LLMs.
Such abilities enable LLMs to generate intelligent decisions that encompass navigation within a 3D environment~\cite{LEO, rana2023sayplan, zheng2023towards}, interaction with objects~\cite{mirjalili2023lan}, and selecting the appropriate tools to execute specific tasks~\cite{hong2024multiply}. 
This section describes how 3D embodied agents perform the tasks of planning, navigation, and manipulation.

\subsubsection{3D Task Planning}
For embodied agents, `task planning' refers to the ability to generate steps to execute a specific task, given the task description and the 3D environment.
Task planning usually serves as a prerequisite for navigation and manipulation tasks~\cite{song2023llmplanner,rana2023sayplan} as the accuracy of planning directly influences the performance of subsequent tasks.

LEO~\cite{LEO} and LLM-Planner~\cite{song2023llmplanner} utilize LLMs to generate step-by-step plans which they adjust dynamically based on environmental perceptions. LEO~\cite{LEO} emphasizes scene-aware planning grounded in the current scene's configurations, whereas LLM-Planner~\cite{song2023llmplanner} adopts GPT-3~\cite{brown2020language} to divide planning into high-level subgoals and low-level actions, and replan when the agent gets stuck during task execution. 
3D-VLA~\cite{zhen20243d} integrates 3D perception, reasoning, and action through a generative world model. It focuses on enhancing planning capabilities by utilizing its generative model to predict future state representations (\eg goal images and point clouds). 
Agent3D-Zero~\cite{zhang2024agent3dzero} introduces Set-of-Line Prompting (SoLP) that enhances the VLM's comprehension of the scene's geometric aspects by generating diverse observational viewpoints. Specifically, SoLP superimposes grid lines and tick marks onto BEV images, and prompts the VLM to provide more accurate camera positions and orientations, which enables the task of the VLM to understand 3D spatial concepts.
UniHSI~\cite{xiao2023unified} addresses the task of Human-Scene Interaction (HSI), which involves generating interactions between humans and objects within 3D environments based on input language commands. It uses an LLM as a planner to translate language commands into task plans represented as Chains of Contacts (CoC), a sequence that represents the chronological relationships between human joint points and object positions. While the aforementioned approaches focus on planning within a single scene, SayPlan~\cite{rana2023sayplan} can handle multiple rooms and floors by (i) leveraging 3D scene graphs for semantic search and (ii) integrating classical path planning with an iterative replanning pipeline for plan refinement.
NLMap~\cite{chen2022nlmapsaycan} proposes an open-vocabulary, queryable scene map for LLM planners, enabling them to see and query available objects before generating plans. 

\subsubsection{3D Navigation}

3D Navigation refers to the ability of an embodied agent to move and orient itself within 3D environments, often based on visual inputs and language instructions. Each of the described methods -- LEO~\cite{LEO}, Agent3D-Zero~\cite{zhang2024agent3dzero}, LLM-Planner~\cite{song2023llmplanner}, and NaviLLM~\cite{zheng2023towards} -- implements 3D navigation in different ways. 
LEO~\cite{LEO} processes egocentric 2D images and object-centric 3D point clouds alongside textual instructions. It generates a sequence of action tokens that correspond to executable navigation commands like `move forward' or `turn right'. LEO employs `shortest path navigation trials', which provide a less noisy and more straightforward learning environment compared to human demonstrations.
Agent3D-Zero~\cite{zhang2024agent3dzero} navigates by continuously selecting new viewpoints based on assessments of the environment. It incorporates historical data from previous viewpoints to refine its navigation path towards specific goals, such as finding a printer in an office setting.  
LLM-Planner~\cite{song2023llmplanner} employs a hierarchical approach that first generates high-level plans as sequences of subgoals, which are then translated into a series of primitive actions by a low-level planner. This makes the overall process adaptable to the immediate environment. NaviLLM~\cite{zheng2023towards} transforms various embodied navigation tasks into generation problems using schema-based instructions. These instructions include $4$ elements: a task defined by a word sequence, observations of all reachable viewpoints, the history of past visual observations, and output hints that guide action generation (\eg selecting a direction or object).

\subsubsection{3D Object Manipulation}

Manipulation in the context of 3D embodied agents refers to their ability to physically interact with objects, ranging from moving objects to complex sequences such as assembling parts or opening doors. The core idea used to enable LLMs to perform manipulation tasks lies in tokenizing action sequences. To let LLMs output specific actions, it is first necessary to define action tokens that allow LLMs to generate said actions, based on the task and 3D scene context. Subsequently, platforms like CLIPort~\cite{shridhar2022cliport} or motion planning modules in robot arms translate these tokenized actions into physical movements to be executed by the agent.

LEO~\cite{LEO}, MultiPLY~\cite{hong2024multiply}, and 3D-VLA~\cite{zhen20243d} each use different action tokens to convert spoken or written instructions into actions for robots in 3D spaces. 
LEO~\cite{LEO} uses more than 500 specific tokens to make robot movements precise. Specifically, for the CLIPort~\cite{shridhar2022cliport} task, action poses are encoded using 516 tokens: 320 tokens for the x-axis pose bins, 160 tokens for the y-axis, and 36 tokens for z-rotation bins.
MultiPLY~\cite{hong2024multiply} extends this by introducing tokens like $\langle$SELECT$\rangle$ for object interaction, $\langle$NAVIGATE$\rangle$ for movement, $\langle$OBSERVE$\rangle$ for scrutiny, $\langle$TOUCH$\rangle$ for tactile feedback, $\langle$HIT$\rangle$ for auditory feedback, $\langle$PICK-UP$\rangle$ and $\langle$PUT-DOWN$\rangle$ for manipulation, and $\langle$LOOK-AROUND$\rangle$ for awareness. This approach also integrates sensory feedbacks (tactile, temperature, and auditory) which enhances the interaction of the robot with its surroundings.
3D-VLA~\cite{zhen20243d} incorporates (i) object tokens ($\langle$obj$\rangle$$\langle$/obj$\rangle$) to identify manipulated objects, (ii) location tokens ($\langle$loc0-255$\rangle$) for spatial grounding, and (iii) specialized tokens for robotic actions like arm location/rotation/gripper state. These tokens are separated by $\langle$ACT SEP$\rangle$. This token structure enables understanding and executing complex 3D manipulation.

While these systems enable robots to perform complex tasks by mapping instructions to actions, they overlook semantic understanding of manipulable objects and are often unable to distinguish between suitable and unsuitable parts for manipulation. To address this issue, VoxPoser~\cite{huang2023voxposer}, LAN-grasp~\cite{mirjalili2023lan}, and ManipLLM~\cite{li2023manipllm} focus on ``affordance'', and create affordance maps to represent objects and features in their surroundings that can be utilized to perform specific tasks, such as graspable handles~\cite{mirjalili2023lan,li2023manipllm}, pressable buttons~\cite{li2023manipllm}, or movable objects~\cite{huang2023voxposer}.
Specifically, VoxPoser~\cite{huang2023voxposer} uses an LLM to decompose free-form language instructions, infer affordances and constraints, and compose 3D voxel maps by interacting with VLMs using code interfaces. These maps enable generating closed-loop robot trajectories that are robust to dynamic changes, with capability to learn from online experiences in contact-rich environments.
LAN-grasp~\cite{mirjalili2023lan} employs foundation models to deepen robots' understanding of objects for semantically appropriate grasping by combining multiple models to identify graspable parts, without the need for retraining. 
ManipLLM~\cite{li2023manipllm} predicts manipulation outcomes by identifying 3D coordinates for contact points and gripper orientations from text prompts, RGB images, and depth maps.

\subsection{LLMs for 3D Generation} \label{sec:llm-generation}

Traditionally, 3D modeling has been a complex time-intensive process with a high barrier to entry, requiring detailed attention to geometry, texture, and lighting to achieve realistic results. In this section, we scrutinize the integration of LLMs with 3D generative techniques, where we show how language provides a way to generate contextualized objects in scenes and provides innovative solutions to 3D content creation and manipulation.

\subsubsection{Object-level Generation}

Shape-GPT~\cite{shapegpt} quantizes 3D shapes into discrete ``shape word" tokens using a shape-specific 3D VQ-VAE. This enables the integration of shape data into a multi-modal input for the T5 language model~\cite{2020t5} along with text and images. This multi-modal representation enables T5 to learn cross-modal interactions for, \eg text-to-shape generation and shape editing/completion. GPT4Point~\cite{qi2023gpt4point} uses a two-stream approach - aligning point cloud geometry with text via Point-QFormer, which then feeds into coupled LLM and diffusion paths for text comprehension and high-fidelity 3D object generation conforming to text input. 

In contrast, MeshGPT~\cite{siddiqui2023meshgpt}, LLaMA-Mesh~\cite{wang2024llamamesh}, and PolyGen~\cite{nash2020polygen} do not condition the generation on text, but they still adopt an autoregressive approach akin to sequence modelling in LLMs.
MeshGPT uses graph convolutions to encode mesh geometry/topology into rich embeddings compressed via Residual Vector Quantization and fed into a GPT-style transformer to autoregressively predict tokens/embeddings for generating meshes with desired properties. PolyGen~\cite{nash2020polygen} is an autoregressive transformer-based model of 3D meshes, which utilizes pointer networks. It consists of a vertex model that unconditionally models mesh vertices, and a face model for mesh faces conditioned on input vertices using autoregressive networks to output face indices and vertex coordinates for generating diverse high-quality meshes.
LLaMA-Mesh~\cite{wang2024llamamesh} represents 3D meshes in a purely textual form by expressing vertex coordinates and face indices as plain text sequences, with coordinate values discretized into integers for compactness, allowing the direct use of pretrained LLM tokenizers and architectures without introducing a mesh-specific vocabulary.

\subsubsection{Scene-scale Generation}
Holodeck~\cite{yang2024holodeck} and GALA-3D~\cite{gala3d} employ multi-stage pipelines to progressively refine an initial coarse 3D scene layout from text into a detailed realistic 3D environment. 
Holodeck employs specialized modules to craft a basic layout, select materials, and incorporate elements like doors and windows based on GPT-4’s spatial reasoning and placement/style suggestions.
It then populates the layout with Objaverse assets matched to GPT-4's textual descriptions. An optimizer arranges these objects adhering to spatial relational constraints obtained from GPT-4 encouraging realistic object layouts and interactions.

GALA-3D~\cite{gala3d} first generates coarse layouts from text using an LLM, then translates them into a 3D Gaussian representation. This representation serves as the foundation for creating detailed 3D content using instance-level text-to-image diffusion priors. It employs compositional optimization to fine-tune the parameters of the layout-guided Gaussians, ensuring the final scene aligns with the text for object placement, scale, and interaction.
Both leverage complementary strengths of LLMs to extract high-level semantic layouts and generative models/optimization to convert these into geometrically and physically plausible 3D scenes.

\subsubsection{Procedural Generation and Manipulation}

LLMR~\cite{de2023llmr}, 3D-GPT~\cite{3dgpt} and SceneCraft~\cite{hu2024scenecraft} adopt modular architectures with specialized components/agents for interactive 3D world creation and code generation from natural language.
LLMR consists of distinct components for generating code to construct scenes in Unity, understanding existing scene objects and properties to enable modifications, identifying required functions to execute instructions, and evaluating final code quality. Similarly, 3D-GPT has components for interpreting instructions and determining required generation functions, enriching descriptions with detailed modeling attributes, and translating enriched descriptions to Python code for the Blender API. Overall, these methods showcase task decomposition and specialization of LLM components to handle instruction interpretation, function mapping, and robust code generation.

\section{3D Tasks with VLMs}
\label{sec:3d-vlms}

While \Cref{sec:3d-llms} discussed methods integrating LLMs in 3D tasks, a vast body of research has explored various aspects of 3D understanding through the lens of 2D Vision-Language Models (VLMs). VLMs include richer visual priors that can be linked to 3D representations. This section reviews contributions from recent papers spanning open-world understanding, instance-level recognition, unified end-to-end architectures, spatial reasoning, and generation.

\subsection{Open-Vocabulary 3D Scene Understanding}
Open-vocabulary 3D scene understanding recognizes and describes scene elements using natural language instead of predefined category labels. OpenScene~\cite{Peng2023OpenScene} adopts a zero-shot strategy by predicting dense 3D features co-embedded with CLIP's text and image embeddings in a shared space, enabling task-agnostic querying of objects, materials, affordances, and room types. CLIP-FO3D~\cite{zhang2023clipfo3d} similarly distills dense pixel features from CLIP to 3D point clouds via feature projection and training a 3D model to absorb CLIP knowledge. Semantic Abstraction~\cite{semanticabstraction} extracts CLIP relevance maps as abstract object representations to generalize across vocabularies and domains. Open-Fusion~\cite{kashu2023openfusion} integrates SEEM~\cite{zou2024segment} with TSDF 3D mapping for real-time open-vocabulary scene creation using region-based embeddings and confidence maps.

PLA~\cite{ding2022language} and RegionPLC~\cite{yang2023regionplc} use contrastive learning to associate captions with 2D and 3D modalities. PLA aligns multi-view images with captions via 3D-caption contrastive pairs, while RegionPLC aligns region-level 2D captions to 3D points. OVIR-3D~\cite{lu2023ovir3d} fuses 2D region proposals with CLIP-aligned features to support efficient open-vocabulary 3D instance retrieval. CoDA~\cite{cao2023coda} incorporates annotated 3D geometry priors with 2D semantics from CLIP to discover novel objects using Discovery-driven Cross-Modal Alignment (DCMA).

For instance-level understanding, OpenMask3D~\cite{takmaz2023openmask3d} and Open3DIS~\cite{nguyen2023open3dis} use class-agnostic 3D instance masks and 2D segment-level CLIP embeddings for open-vocabulary 3D instance segmentation. OpenIns3D~\cite{huang2023openins3d} bypasses image alignment by predicting 3D mask proposals, rendering synthetic images, and assigning mask labels via a language model. Rozenberszki~\etal~\cite{rozenberszki2022language} align 3D features with CLIP for semantic and instance-level segmentation. Diff2Scene~\cite{zhu2024open} distills semantic knowledge from frozen text-to-image diffusion models, using geometry- and saliency-aware masks to enable label-free open-vocabulary 3D segmentation and grounding.

Language grounding with NeRFs has shown promise in open-vocabulary 3D understanding. Methods like DFF~\cite{kobayashi2022distilledfeaturefields}, LERF~\cite{lerf2023}, VL-Fields~\cite{tsagkas2023vlfields}, and 3D-OVS~\cite{liu20243dovs} distill 2D VLM knowledge into implicit 3D feature fields by minimizing rendering error w.r.t. 2D features. LERF further learns a scale-conditioned language field over 3D space. LangSplat~\cite{qin2023langsplat} and N2F2~\cite{bhalgat2024n2f2} extend this to 3D Gaussian Splatting, enabling efficient querying via hierarchical supervision and multiscale feature fields.

\subsection{Text-Driven 3D Generation}
\Cref{sec:llm-generation} reviewed LLM-guided generation; here, we survey methods leveraging 2D VLMs~\cite{radford2021learning} and text-to-image diffusion models~\cite{rombach2022high,saharia2022imagen} with differentiable rendering. Early methods like DreamFields~\cite{jain2022dreamfields}, CLIP-Mesh~\cite{khalid2022clipmesh}, CLIP-Forge~\cite{sanghi2022clipforge}, and Text2Mesh~\cite{michel2022text2mesh} explore CLIP-guided zero-shot 3D optimization.

DreamFusion~\cite{poole2022dreamfusion} introduced Score Distillation Sampling (SDS), optimizing NeRF parameters so rendered views are realistic under a text-to-image diffusion model (Imagen~\cite{saharia2022imagen}). Magic3D~\cite{lin2022magic3d} adds a two-stage pipeline - coarse NeRF generation followed by mesh optimization using high-res latent diffusion. Fantasia3D~\cite{chen2023fantasia3d} disentangles geometry and appearance via a DMTet~\cite{shen2021dmtet} hybrid with spatially varying BRDFs. ProlificDreamer~\cite{wang2024prolificdreamer} introduces Variational SDS (VSD), treating 3D parameters as particles for more diverse generation. Dream3D~\cite{xu2023dream3d} enhances synthesis using explicit 3D shape priors. MVDream~\cite{shi2023mvdream} trains multi-view consistent diffusion on few-shot data. Text2NeRF~\cite{zhang2024text2nerf} uses pre-trained diffusion models to guide NeRF generation from language. CrossOver~\cite{sarkar2025crossover} learns a modality-agnostic space across images, point clouds, CAD models, floorplans, and text, enabling cross-modal scene retrieval and generation.

Some works target text-guided texture synthesis only~\cite{richardson2023texture, chen2023text2tex, chen2023scenetex}. For avatars, AvatarCraft~\cite{jiang2023avatarcraft} uses diffusion to learn geometry/texture from text and animate avatars via warping fields. AvatarCLIP~\cite{hong2022avatarclip} provides a zero-shot CLIP-supervised framework for geometry, texture, and motion synthesis. GenZI~\cite{li2024genzi} and CG-HOI~\cite{diller2024cghoi} model human-object interactions from text.

Compositional approaches like CG3D~\cite{vilesov2023cg3d} assemble Gaussian radiance fields from object parts. Po~\etal~\cite{po2023compositional} enable local scene control via text and bounding boxes. GraphDreamer~\cite{gao2024graphdreamer} composes scenes from graph-derived global-local descriptions using object SDF optimization.

\subsection{End-to-End Architectures for 3D Vision \& Language}
Transformer-based 3D-text models learn joint visual-language representations. 3D-VisTA~\cite{3dvista} models both modalities with self-attention and is pre-trained on masked modeling and scene-text matching. UniT3D~\cite{chen2023unit3d} combines PointGroup, BERT, and multi-modal fusion, trained on synthetic 3D-text data. SpatialVLM~\cite{chen2024spatialvlm} trains VLMs on synthetic 3D spatial reasoning tasks for improved 3D spatial VQA and chain-of-thought reasoning. Multi-CLIP~\cite{delitzas2023multiclip} aligns 3D scene encodings with CLIP's embedding space for better cross-modal transfer. Lexicon3D~\cite{man2024lexicon3d} benchmarks a wide range of foundation models across 3D vision-language tasks, identifying key strengths (e.g., DINOv2 for vision, diffusion for grounding) and challenges in multimodal alignment.

End-to-end 3D-language models unify perception, grounding, and captioning. D3Net~\cite{d3net} includes a detector, a speaker module for captioning, and a listener for grounding. Uni3DL~\cite{li2023uni3dl} operates directly on point clouds with shared encoders for text and points, predicting outputs for segmentation, detection, grounding, and captioning. InstanceRefer~\cite{yuan2021instancerefer} and LanguageRefer~\cite{roh2022languagerefer} combine linguistic and spatial cues for grounding in 3D point clouds. 3DVG-Transformer~\cite{zhao20213dvg} uses coordinate-guided attention and contextual feature fusion for grounding descriptions to 3D objects. SAGA~\cite{cen2025segment} enables prompt-based segmentation using 3D Gaussian splats with SAM-derived affinity fields, supporting real-time open-vocabulary interaction across granularities.

\section{Datasets and Benchmarking}
\label{sec:datasets}

\begin{table*}[ht!]
    \renewcommand{\arraystretch}{0.9} 
    \begin{center}
    \scalebox{0.95}{
        \begin{tabular}{l|ccccccccc|cc|ccc|ccc|cc} 
            \multicolumn{1}{c}{Dataset} & \rot{Object Captioning} & \rot{Scene Captioning} & \rot{Dense Captioning} & \rot{Single Object Grounding} & \rot{Multiple Object Grounding} & \rot{Question Answering} & \rot{Situated Question Answering} & \rot{Dialogue} & \rot{Task Planning} & \rot{Real} & \rot{Synthetic} & \rot{Object} & \rot{Indoor} & \rot{Outdoor} & \rot{Human-Annotated} & \rot{Model-Annotated} & \rot{Template-Based} & \rot{\# Language Pairs} & \rot{\# Scenes} \\
            \addlinespace[0.5ex] 
            \toprule
            \rowcolor[HTML]{EFEFEF}
            Cap3D~\cite{luo2024scalable} & \checkmark & & & & & & & & & \checkmark & \checkmark & \checkmark & & & & \checkmark & & 660k & 660k \\
            Text2Shape~\cite{chen2019text2shape} & \checkmark & & & & & & & & & & \checkmark & \checkmark & & & \checkmark & & \checkmark & 75k & 15k \\
            \rowcolor[HTML]{EFEFEF}
            SceneVerse~\cite{jia2024sceneverse} & \checkmark & \checkmark & & \checkmark & & & & & & \checkmark & \checkmark & \checkmark & & & \checkmark & \checkmark & \checkmark & 2.5M & 68k \\
            nu-Caption~\cite{yang2023lidar} & & \checkmark & & & & & & & & \checkmark & & & & \checkmark & \checkmark & \checkmark & & 420k & - \\
            \rowcolor[HTML]{EFEFEF}
            nu-Grounding~\cite{yang2023lidar} & & \checkmark & & \checkmark & & & & & & \checkmark & & & & \checkmark & \checkmark & \checkmark & & 280k & - \\
            ScanRefer~\cite{scanrefer} & & & \checkmark & \checkmark & & & & & & \checkmark & & & \checkmark & & \checkmark & & & 51k & 800 \\
            \rowcolor[HTML]{EFEFEF}
            ReferIt3D~\cite{achlioptas2020referit3d} & & & \checkmark & \checkmark & & & & & & \checkmark & & & \checkmark & & \checkmark & & \checkmark & 125k & 707 \\
            Multi3DRefer~\cite{zhang2023multi3drefer} & & & \checkmark & & \checkmark & & & & & \checkmark & & & \checkmark & & \checkmark & \checkmark & & 60k & 800 \\
            \rowcolor[HTML]{EFEFEF}
            Chat-3D v2~\cite{huang2023chatV2} & & \checkmark & & \checkmark & & & & & & \checkmark & & & \checkmark & & & \checkmark & & 705 & 705 \\
            EmbodiedScan~\cite{wang2023embodiedscan} & & & & \checkmark & & & & & & \checkmark & & & \checkmark & & & \checkmark & \checkmark & 970k & 3.4k \\
            \rowcolor[HTML]{EFEFEF}
            ScanEnts3D~\cite{abdelreheem2024scanents3d} & & & \checkmark & \checkmark & & & & & & \checkmark & & & \checkmark & & \checkmark & & & 84k & 700 \\
            WildRefer~\cite{lin2023wildrefer} & & & & \checkmark & & & & & & \checkmark & & & \checkmark & \checkmark & \checkmark & & & 30.8k & 3.8k \\
            \rowcolor[HTML]{EFEFEF}
            RIORefer~\cite{RIORefer} & & & & \checkmark & & & & & & \checkmark & & & \checkmark & & \checkmark & & & 1.4k & 63k \\
            ARKitSceneRefer~\cite{ARKitSceneRefer} & & & & \checkmark & & & & & & \checkmark & & & \checkmark & & \checkmark & \checkmark & & 15k & 1.6k \\
            \rowcolor[HTML]{EFEFEF}
            ScanERU~\cite{ScanERU} & & & & \checkmark & & & & & & \checkmark & \checkmark & & \checkmark & & \checkmark & & & 46k & 706 \\
            PhraseRefer~\cite{yuan2022toward} & & & & \checkmark & & & & & & \checkmark & & & \checkmark & & \checkmark & & \checkmark & 170k & 700 \\
            \rowcolor[HTML]{EFEFEF}
            DenseGrounding~\cite{densegrounding} & & & & & \checkmark & & & & & \checkmark & & & \checkmark & & \checkmark & & & - & - \\
            3DVQA~\cite{etesam20223dvqa} & & & & & & \checkmark & & & & \checkmark & & & \checkmark & & & & \checkmark & 484k & 707 \\
            \rowcolor[HTML]{EFEFEF}
            ScanQA (Azuma \etal)~\cite{scanqa} & & & & & & \checkmark & & & & \checkmark & & & \checkmark & & \checkmark & & & 41k & 800 \\
            ScanQA (Ye \etal)~\cite{ye2021tvcg3dqa} & & & & & & \checkmark & & & & \checkmark & & & \checkmark & & \checkmark & & & 10k & 806 \\
            \rowcolor[HTML]{EFEFEF}
            3DMV-VQA~\cite{hong20233d} & & & & & & \checkmark & & & & \checkmark & & & \checkmark & & & & \checkmark & 50k & 5k \\
            NuScenes-QA~\cite{qian2023nuscenes} & & & & & & \checkmark & & & & \checkmark & & & & \checkmark & & & \checkmark & 460k & 34k \\
            \rowcolor[HTML]{EFEFEF}
            CLEVR3D~\cite{yan2023comprehensive} & & & & & & \checkmark & & & & \checkmark & \checkmark & & \checkmark & & & & \checkmark & 171k & 8.7k \\
            SQA-3D~\cite{ma2022sqa3d} & & & & & & & \checkmark & & & \checkmark & & & \checkmark & & \checkmark & & & 33.4k & 650 \\
            \rowcolor[HTML]{EFEFEF}
            MSQA/MSNN~\cite{linghu2024multi} & & & & & & & \checkmark & & \checkmark & \checkmark & & & \checkmark & & & \checkmark & & 251k & 1.7k \\
            3D-LLM~\cite{3dllm} & \checkmark & \checkmark & \checkmark & \checkmark & & \checkmark & \checkmark & \checkmark & \checkmark & \checkmark & & & \checkmark & & & \checkmark & & 200k & 900 \\
            \rowcolor[HTML]{EFEFEF}
            ScanScribe~\cite{zhu20233d} & & & \checkmark & \checkmark & & \checkmark & \checkmark & & & \checkmark & \checkmark & & \checkmark & & \checkmark & \checkmark & \checkmark & 278k & 1.2k \\
            M3DBench~\cite{li2023m3dbench} & & \checkmark & \checkmark & \checkmark & \checkmark & \checkmark & \checkmark & \checkmark & \checkmark & \checkmark & & & \checkmark & & \checkmark & \checkmark & \checkmark & 327k & - \\
            \rowcolor[HTML]{EFEFEF}
            GPT4Point~\cite{qi2023gpt4point} & \checkmark & & & & & \checkmark & & \checkmark & & \checkmark & \checkmark & \checkmark & & & & \checkmark & & 1M & 1M \\
            LAMM~\cite{yin2023lamm} & \checkmark & \checkmark & & & & \checkmark & & \checkmark & & \checkmark & \checkmark & \checkmark & \checkmark & & & \checkmark & \checkmark & - & - \\
            \rowcolor[HTML]{EFEFEF}
            3D-GRAND/3D-POPE~\cite{yang2024_3D_GRAND} & \checkmark & \checkmark & \checkmark & \checkmark & \checkmark & \checkmark & &  & &  & \checkmark & & \checkmark & & & \checkmark & \checkmark & 6.2M & 40k \\
            LEO~\cite{LEO} & \checkmark & \checkmark & & & & \checkmark & \checkmark & \checkmark & \checkmark & \checkmark & \checkmark & \checkmark & \checkmark & & & \checkmark & & - & - \\
            \rowcolor[HTML]{EFEFEF}
            3DMIT~\cite{li20243dmit} & & \checkmark & & \checkmark & & \checkmark & & \checkmark & & \checkmark & & & \checkmark & & \checkmark & \checkmark & & 75k & 1.5k \\
            Spartun3D~\cite{zhang2024spartun3d} & & \checkmark & & \checkmark & & \checkmark & & \checkmark & & \checkmark & & & \checkmark & & \checkmark & \checkmark & & 133k & - \\
            \rowcolor[HTML]{EFEFEF} 
            3D-CoT~\cite{chen2025integrating} & & & & & & \checkmark & & & & \checkmark & \checkmark & \checkmark & & & \checkmark & \checkmark & & 1.51M & - \\
            Beacon3D~\cite{huang2025unveiling} & & & & \checkmark & & \checkmark & & & & \checkmark & & & \checkmark & & \checkmark & & & 5k & 30 \\
            \rowcolor[HTML]{EFEFEF} 
            SPAR-Bench~\cite{zhang2025flatland} & & & & & & \checkmark & \checkmark & & \checkmark & \checkmark & & & \checkmark & & & & \checkmark & 7M & 4k \\
            SpatialVQA~\cite{ma2025spatialllm} & & & & & & \checkmark & & & & \checkmark & & & \checkmark & \checkmark & & & \checkmark & 1.3k & - \\
            \rowcolor[HTML]{EFEFEF} 
            Space3D-Bench~\cite{szymanska2025space3d} & & & & & & \checkmark & & & \checkmark & \checkmark & & & \checkmark & & \checkmark & & & 1k & - \\
            City-3DQA~\cite{sun20243d} & & & & & & \checkmark & & & & \checkmark & & & & \checkmark & \checkmark & \checkmark & \checkmark & 450k & - \\
            \rowcolor[HTML]{EFEFEF} 
            MMScan~\cite{lyu2024mmscan} & & & & & & \checkmark & & & & \checkmark & & & \checkmark & & \checkmark & \checkmark & \checkmark & 6.9M & 5.2k \\
            Reason3D & & & & \checkmark & \checkmark & & & & \checkmark & & & & \checkmark & & & \checkmark & & 2.5k & - \\\rowcolor[HTML]{EFEFEF} 
            LEO-VL & \checkmark & \checkmark & & & & \checkmark & \checkmark & \checkmark & \checkmark & \checkmark & & & \checkmark & & \checkmark & \checkmark & \checkmark & 700k & - \\
            SpaCE-10 & & & & & & \checkmark & \checkmark & & \checkmark & \checkmark & & & \checkmark & & \checkmark & & \checkmark & 6k & 811 \\
            \bottomrule
        \end{tabular}
    }
    \end{center}
    \vspace{-3mm}
    \caption{
    \textbf{An overview of datasets} used for 3D-Related tasks using Large Language Models. For each dataset, we show which tasks that dataset has been used to demonstrate, whether the data is captured from the real world or synthetically generated, whether the 3D data is of an object, an indoor scene or an outdoor scene, and how the annotations are obtained. We focus primarily on newer datasets which are used to evaluate methods in recent research papers. 
    }
    \vspace{-3mm}
    \label{tab:datasets}
\end{table*}

\begin{figure*}
    \centering
    \includegraphics[width=0.99\textwidth]{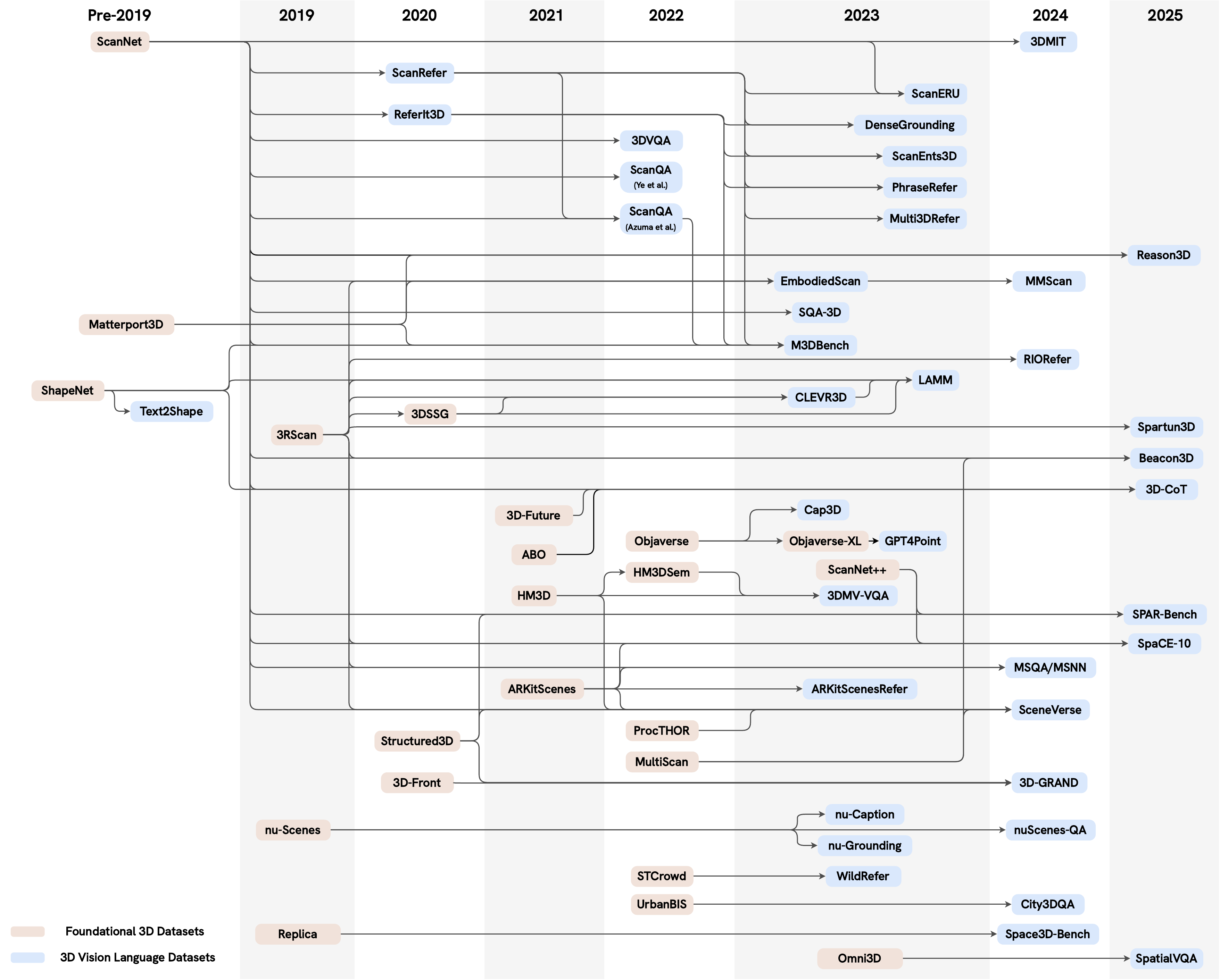}
    \caption{\textbf{Timeline of datasets.} A timeline showing how existing datasets are combined and annotated to form new datasets for 3D vision language tasks. Datasets in orange are foundational 3D datasets without language annotations and datasets in blue are the annotated datasets used in 3D vision language tasks. Note that WildRefer also introduces new 3D data and annotations for vision-language tasks. 
    }
    \label{fig:dataset-timeline}
\end{figure*}

We now provide a high-level overview of the datasets that are used to train and evaluate 3D vision-language models. In Table \ref{tab:datasets}, we provide a list of datasets alongside the tasks they are used for, as well as information about their 3D data and annotations.
In \figref{fig:dataset-timeline}, we present these datasets on a timeline, showing where each dataset sourced 3D information from. Current 3D vision-language datasets are almost exclusively generated by taking existing 3D vision datasets, and applying human, model or templated annotations to samples. As seen in Table \ref{tab:datasets}, a majority of existing datasets focus on real, indoor scenes due to using 3D scans from either ScanNet~\cite{dai2017scannet} or 3RScan~\cite{Wald2019RIO}. Many of the datasets presented here share the same 3D data, and instead differ primarily through their choice of annotation strategy, and the 3D vision-language task they are designed to be used for. Among outdoor datasets, the majority use nuScenes~\cite{nuscenes2019}, which focuses on LiDAR data for autonomous driving.

There are three current strategies for collecting language data for these datasets: human-based, template-based and model-based annotation. Prior to ChatGPT~\cite{chatgpt}, language data was collected from human annotators labelling data with natural language, or by having humans construct natural language `templates', where appropriate classes and spatial references are taken from the dataset (such as from scene graphs) and placed into the template phrases. However, human-based annotation is expensive and time-consuming, and template-based annotation can have limited linguistic diversity. Recently, datasets have been using LLMs to analyze and annotate 3D data, either by asking the LLM to analyze a generated text description of each 3D scene, or by using 2D vision or vision-language models to analyze multiple 2D images of the scene and collating the responses. This has allowed 3D vision language datasets to contain orders of magnitude more language annotations, but can fail when the LLM does not generate accurate and naturalistic annotations for the samples.
See Table \ref{tab:datasets} for a breakdown of how each dataset is annotated.

Due to LLMs' ability to act as multi-task learners, several recent datasets~\cite{qi2023gpt4point, li20243dmit, LEO, zhu20233d, li2023m3dbench, yin2023lamm, yang2024_3D_GRAND} have focused on collecting data for many different tasks, attempting to unify the input and output formats for each task (see \cref{subsec:LLM_as_3D_Multi_task_learner}). These datasets typically draw 3D data from a greater range of datasets to expand data size and diversity (see \figref{fig:dataset-timeline}). There are a few strategies to improve the data diversity by leveraging multiple datasets, such as randomly replacing 10\% of the objects in scenes with a same-category object from Objaverse~\cite{zhu20233d}, or randomly replacing objects in the scene graphs with objects from a generated object pool~\cite{yan2023comprehensive}. Many datasets can also be used beyond their intended task. Dense captioning methods such as  Scan2Cap~\cite{chen2021scan2cap} can use the referring expressions from grounding datasets like ScanRefer~\cite{scanrefer} to create the captions needed to train a model. Datasets that were designed for text-to-3D generation tasks, such as ~\cite{chen2019text2shape}, can also be used for 3D-to-text captioning tasks. Other datasets are introduced as a way of testing the cross-dataset generalization abilities of models~\cite{RIORefer}.


Datasets for 3D navigation and manipulation using language are often designed around specific requirements and have a large overlap with existing bodies of research. We refer readers to existing survey papers \cite{zeng2023large,zhou2023language} for an overview of these datasets. Similarly, for text-to-3D generation datasets, we direct readers to the recent survey by Lee \etal~\cite{lee2024text}. We omit further discussion here due to prior extensive coverage, and because many methods use 2D vision-language data rather than 3D-specific datasets. In Fig. ~\ref{tab:benchmarks}, we provide a quantitative comparison between a number of existing 3D-VLM methods on the common testing datasets ScanRefer~\cite{scanrefer}, Multi3DRefer~\cite{zhang2023multi3drefer}, Scan2Cap~\cite{chen2021scan2cap}, ScanQA~\cite{scanqa} and SQA3D~\cite{ma2022sqa3d}. We include these results to highlight the variety of metrics and datasets used even among similar methods.

\begin{table*}[ht!]
\centering
\resizebox{\textwidth}{!}{%
    \begin{tabular}{lcccccccccccccccc}
        \toprule
        \multirow{2}{*}{\textbf{Model}} & \multicolumn{2}{c}{\textbf{ScanRefer}} & \multicolumn{2}{c}{\textbf{Multi3DRefer}} & \multicolumn{4}{c}{\textbf{Scan2Cap}} & \multicolumn{6}{c}{\textbf{ScanQA (val)}} & \multicolumn{2}{c}{\textbf{SQA3D (test)}} \\
        \cmidrule(lr){2-3} \cmidrule(lr){4-5} \cmidrule(lr){6-9} \cmidrule(lr){10-15} \cmidrule(lr){16-17}
        & Acc@0.25 & Acc@0.5 & F1@0.25 & F1@0.5 & B-4@0.5 & C@0.5 & M@0.5 & R@0.5 & M & R & C & EM & B-1 & B-4 & EM & EM-R \\
        \midrule
        \multicolumn{17}{l}{\textit{Specialists}} \\
        \midrule
        ScanRefer~\cite{scanrefer} & 37.3 & 24.3 & - & - & - & - & - & - & - & - & - & - & - & - & - & - \\
        Scan2Cap~\cite{chen2021scan2cap} & - & - & - & - & 22.4 & 35.2 & 22.0 & 44.8 & - & - & - & - & - & - & - & - \\
        Vote2Cap-DETR++~\cite{chen2023end} & - & - & - & - & 34.5 & 61.8 & \textbf{26.2} & \textbf{54.4} & - & - & - & - & - & - & - & - \\
        ScanQA~\cite{scanqa} & - & - & - & - & - & - & - & - & 13.1 & 33.3 & 64.9 & 21.1 & 30.2 & 10.1 & 47.2 & - \\
        SQA3D~\cite{ma2022sqa3d} & - & - & - & - & - & - & - & - & 13.5 & 34.5 & - & - & \textbf{30.5} & \textbf{11.2} & 46.6 & - \\
        D3Net~\cite{d3net} & - & 37.9 & - & \textbf{32.2} & \textbf{35.7} & 62.6 & 25.7 & 53.9 & - & - & - & - & - & - & - & - \\
        3D-ViSTA~\cite{zhu20233d} & \textbf{50.6} & \textbf{45.8} & - & - & 34.0 & \textbf{66.9} & - & - & \textbf{13.9} & \textbf{35.7} & \textbf{69.6} & \textbf{22.4} & - & - & 48.5 & - \\
        3DVG-Transformer~\cite{zhao20213dvg} & 45.9 & 34.5 & - & 25.5 & - & - & - & - & - & - & - & - & - & - & - & - \\
        SIG3D~\cite{man2024SIG3D} & - & - & - & - & - & - & - & - & - & - & - & - & - & - & \textbf{52.6} & - \\
        \midrule
        \multicolumn{17}{l}{\textit{Generalists (Point-Based Methods)}} \\
        \midrule
        Chat-3D~\cite{chat3d} & - & - & - & - & - & - & - & - & - & - & 53.2 & - & - & - & - & - \\
        Chat-3D v2~\cite{huang2023chatV2} & 42.5 & 38.4 & 45.1 & 41.6 & 31.8 & 63.9 & - & - & 16.1 & 40.1 & 87.6 & - & 38.4 & 7.3 & 54.7 & - \\
        LL3DA~\cite{chen2023ll3da} & - & - & - & - & 36.0 & 62.9 & 26.0 & 55.0 & 15.9 & 37.3 & 76.8 & - & - & 13.5 & - & - \\
        Scene-LLM~\cite{fu2024scenellm} & - & - & - & - & - & - & - & - & 16.6 & 40.0 & 80.0 & \textbf{27.2} & - & - & 53.6 & - \\
        LEO~\cite{LEO} & - & - & - & - & 38.2 & 72.4 & \textbf{27.9} & \textbf{58.1} & 16.2 & 39.3 & \textbf{101.4} & 21.5 & - & 11.5 & 50.0 & 52.4 \\
        Chat-Scene~\cite{huang2024chatscene} & 55.5 & 50.2 & 57.1 & 52.4 & 36.3 & 77.1 & - & - & 18.0 & 41.6 & 87.7 & 21.6 & \textbf{43.2} & 14.3 & 54.6 & 57.5 \\
        3D-LLaVA~\cite{deng20253dllave} & 51.2 & 40.6 & - & - & 36.9 & 78.8 & 27.1 & 57.7 & 18.4 & 43.1 & 92.6 & 27.0 & - & \textbf{17.1} & 54.5 & 56.6 \\
        Robin3D~\cite{kang2025robin3d} & \textbf{60.8} & \textbf{55.1} & \textbf{64.9} & \textbf{59.7} & \textbf{38.4} & \textbf{87.2} & - & - & \textbf{19.2} & \textbf{44.0} & - & - & - & - & \textbf{56.9} & \textbf{59.8} \\
        PerLA~\cite{mei2025PerLA} & - & - & - & - & - & - & - & - & 17.4 & 39.6 & 78.1 & - & - & 14.5 & - & -   \\
        Grounded-3DLLM~\cite{chen2024grounded} & 47.9 & 44.1 & 45.2 & 40.6 & 35.5 & 70.6 & - & - & 15.2 & 37.1 & 72.7 & - & - & - & - \\
        Inst3D-LLM~\cite{yu2025inst3d} & 57.8 & 51.6 & 58.3 & 53.5 & 38.3 & 79.7 & - & - & - & - & 88.6 & 24.6 & - & - & - & -  \\
        Spartun3D-LLM~\cite{zhang2024spartun3d} & - & - & - & - & - & - & - & - & - & - & - & - & - & - & 55.0 & - \\
        LSceneLLM~\cite{zhi2025lscenellm} & - & - & - & - & - & - & - & - & 18.0 & 40.8 & 88.2 & - & - & - & - & - \\

        \midrule
        \multicolumn{17}{l}{\textit{Generalists (Multi-View Image-Based Methods)}} \\
        \midrule
        3D-LLM~\cite{3dllm} & 30.3 & - & - & - & - & - & - & - & 14.5 & 35.7 & 69.4 & 20.5 & 39.3 & 12.0 & - & - \\
        
        LLaVA-3D~\cite{zhu2024llava3d} & 54.1 & 42.4 & - & - & 41.1 & 79.2 & 30.2 & 63.4 & - & - & 91.7 & 27.0 & - & - & 55.6 & - \\
        Video-3D LLM~\cite{zheng2025video3dllm} & 58.1 & 51.7 & 58.0 & 52.7 & 41.3 & 83.8 & 28.5 & 61.7 & 19.8 & 49.0 & 102.1 & 30.1 & 47.1 & 16.2 & 58.6 & - \\
        3DRS~\cite{huang20253DRS} & 62.9 & 56.1 & 60.4 & 54.9 & 41.6 & \textbf{86.1} & - & - & - & - & 104.8 & 30.3 & - & - & 60.6 & - \\
        GPT4Scene-HDM~\cite{qi2025gpt4scene} & 62.6 & 57.0 & 64.5 & 59.8 & - & - & - & - & 18.9 & 46.5 & 96.3 & - & 44.4 & 15.5 & 60.6 & 63.3 \\
        VeBrain~\cite{luo2025vebrain} & \textbf{66.4} & \textbf{60.2} & \textbf{67.8} & \textbf{62.7} & - & - & - & - & 20.1 & 48.2 & 101.5 & - & \textbf{47.7} & 17.3 & - & - \\
        VG LLM~\cite{zheng2025VGLLM} & 51.0 & 45.1 & - & - & 38.7 & 74.1 & 28.0 & 61.0 & - & - & - & - & - & - & - & - \\
        Ross3D~\cite{wang2025ross3d} & 61.1 & 54.4 & 59.6 & 54.3 & \textbf{43.4} & 81.3 & \textbf{30.3} & \textbf{66.9} & \textbf{20.9} & \textbf{50.7} & \textbf{107.0} & \textbf{30.8} & - & \textbf{17.9} & \textbf{63.0} & \textbf{65.7} \\
        LEO-VL~\cite{huang2025leovl} & - & - & - & - & - & - & - & - & 17.4 & 39.6 & 78.1 & - & - & 14.5 & - & - \\
        Spatial-MLLM~\cite{ma2025spatialllm} & - & - & - & - & - & - & - & - & 18.4 & 45.0 & 91.8 & - & 44.8 & 14.8 & 55.9 & 58.7 \\
        
        \midrule
        Human (from ~\cite{li2025embodied}) & - & - & - & - & - & - & - & - & - & - & - & \textbf{51.6} & - & - & \textbf{90.1} & - \\
        \rowcolor[gray]{0.95} Best Model & 66.4 & 60.2 & 67.8 & 62.7 & 43.4 & 87.2 & 30.3 & 66.9 & 20.9 & 50.7 & 107.0 & 30.8 & 47.7 & 17.9 & 63.0 & 65.7 \\
        Model vs. Human Gap & - & - & - & - & - & - & - & - & - & - & - & 40.6\% & - & - & 30.1\% & - \\
        \bottomrule
    \end{tabular}%
}
\vspace{-1mm}
\caption{\textbf{Quantitative comparison} of different models on common 3D vision-language benchmarks. Results are from original papers where applicable. 
Best results in each category (Specialists and Generalists) are highlighted in \textbf{bold}. The bottom row shows the performance gap between the best AI models and human performance.}
\vspace{-3mm}
\label{tab:benchmarks}
\end{table*}

\section{Challenges and Opportunities}
\label{sec:7-discussion}

Despite the progress made in integrating LLMs with 3D data, challenges remain in data representation, model training, evaluation and safety, necessitating further research. 


\subsection{Better 3D Representations for Language Alignment}
A central challenge for 3D-LLMs is the choice of 3D data representation, as it directly affects both geometric fidelity and language alignment. Point clouds dominate due to their simplicity but lack fine-grained detail, while voxel grids and meshes provide richer geometry at the cost of high memory use or poor neural compatibility. Recent works have thus explored representations~\cite{lerf2023, kobayashi2022distilledfeaturefields,gaussian_grouping} that combine semantics with geometry, including differentiable formats such as NeRFs~\cite{Mildenhall20} and 3D Gaussian Splats~\cite{kerbl3Dgaussians}, as well as VGGT-style geometry-aware encoders~\cite{wang2025vggt,xu2025uniugg}. Another promising direction is structured, part-level representations~\cite{wu2024ptv3}, which support hierarchical alignment between language and 3D across multiple granularities—from scenes, to objects, down to fine components.

\subsection{Need for 3D-centric Pre-training and Architectures}

Existing 3D-LLM methods typically rely on a frozen 3D encoder combined with an LLM that has been pretrained purely on text sequences. During fine-tuning, the majority of existing methods use alignment modules that transform native 3D features to sequences of 1D tokens that are directly aligned with the text embedding space of the pretrained LLMs. This reduction from inherently three-dimensional structures to one-dimensional text space leads to a substantial loss of spatial information, and these methods have little direct knowledge of 3D spatial structures beyond what can be inferred from linguistic priors.
This highlights an important opportunity: to move toward 3D-centric pre-training and architectures that treat 3D structure as a first-class representation. Promising directions include introducing pretraining objectives that explicitly teach LLMs 3D relationships, such as masked autoencoding in a 3D token space, geometric consistency, and auxiliary reconstruction losses~\cite{wang2025ross3d}, and jointly pretraining the 3D encoder and the LLM so that both components co-adapt. 
In parallel, alignment modules should be redesigned from simple projection into language space toward bidirectional, structure-aware interfaces that better integrate 3D and text.

\subsection{From Static to Dynamic Geometric Understanding}

Most existing 3D-LLMs focus on static and coarse-grained comprehension of object categories, attributes (\eg size, color), positions, and spatial relationships. However, many real-world tasks also require reasoning about dynamic properties from static 3D scans, such as predicting object affordances~\cite{AffordanceLLM2024_CVPR}, functionality, and physical characteristics~\cite{pgvlm2024,zhan20253DPhysical} (\eg material, fragility, deformability), as well as motion dynamics (\eg free fall) and inter-object interactions (\eg collision, friction~\cite{zhan2025inferringdynamicphysicalproperties}). Beyond this, fine-grained geometric reasoning, including symmetry, perspective, occlusion~\cite{zhan2023amodal}, and containment remain essential for precise spatial understanding. A next step is to move from inferring dynamics from static geometry to directly modeling dynamic 3D scenes (4D, i.e. 3D + time)~\cite{hu20253dllm-mem,zhu20254dbench,zhou2025vlm4d}, enabling LLMs to capture both spatial and temporal aspects of the physical world.

\subsection{Acquisition of Higher Spatial Reasoning Ability}

Recent progress such as OpenAI-o3 and DeepSeek-R1~\cite{guo2025deepseek} have shown that reinforcement learning (RL) can significantly enhance the reasoning ability of LLMs by providing reward signals beyond next-token prediction. However, current 3D-LLMs typically generate a single textual answer without engaging in step-by-step multi-hop reasoning or grounding their outputs to corresponding 3D structures. This lack of explicit reasoning chains and spatial grounding limits interpretability, which is essential for higher-order spatial reasoning. 
An important opportunity is to leverage RL for training 3D-LLMs with explicit reasoning traces and verifiable grounding. In particular, constructing 3D-specific chain-of-thought (3D-CoT) datasets could teach models to decompose problems into multi-step spatial reasoning processes. Together with well-designed reward functions and multi-stage training schemes (\eg pre-training followed by RL-based post-training), 3D-LLMs may gradually acquire more robust, interpretable, and generalizable spatial reasoning skills, which is a largely unexplored direction.

\subsection{Lack of 3D-text Paired Data as a Bottleneck}

The success of LLMs is largely driven by massive text corpora, and 2D-VLMs rely on billions of image-text pairs such as LAION-5B~\cite{schuhmann2022laion}. In contrast, existing 3D vision-language datasets remain orders of magnitude smaller, with the largest containing only millions of 3D-text pairs (Tab.~\ref{tab:datasets}). While VLMs can generate textual descriptions for 3D datasets, the scale of available 3D data is still far behind that of 2D images. Moreover, most existing annotations provide only coarse object-level descriptions, whereas fine-grained alignment across spatial, geometric, and semantic levels is critical for 3D-LLMs. 
This scarcity in both quantity and quality of 3D-text data is a major bottleneck for scaling and capability development in 3D-LLMs. Promising directions to alleviate this include: (1) using simulators~\cite{ge2024behavior} to synthesize diverse 3D data of objects, scenes, and interactions, combined with automated annotation tools for multi-granularity language alignment; and (2) leveraging 3D generative models~\cite{ocal2024sceneteller,yang2024scenecraft} to produce 3D content directly from textual descriptions, thereby expanding data scale while ensuring consistent 3D-text alignment.

\subsection{Need for Accurate and Robust Evaluation Metrics}

Current evaluations of 3D-LLMs largely follow a VQA-style protocol~\cite{scanqa}, where performance is measured only at the text level via similarity scores. However, complex 3D geometric relationships are often difficult to capture through language alone, and text-based answers do not necessarily reflect genuine spatial understanding. In particular, hallucinations pose a major issue: models may output the correct answer by exploiting linguistic priors rather than grounding their reasoning in the underlying 3D scene~\cite{bai2024hallucinationmultimodallargelanguage}. Existing metrics fail to detect this behavior, raising concerns about reliability. Going forward, the community needs more robust and interpretable evaluation methods that explicitly test 3D-grounded understanding and reasoning.

\subsection{Safety Issues of 3D Embodied Agents}
The ultimate goal of understanding 3D scenes is to enable actions, interactions, and transformations in the 3D world, where LLMs can play a crucial role.
In \secref{subsec:3D Embodied Agent}, we discuss how current research leverages LLMs to empower 3D embodied agents in tasks such as planning, navigation, and manipulation, illustrating the feasibility of LLMs directly generating action outputs.
However, a critical yet often overlooked challenge is the potential safety risks when LLMs generate physical actions. For instance, when controlling a robotic arm to assist an elderly person during feeding, the agent must carefully regulate its strength and speed to prevent harm. LLMs, however, may lack precise physical awareness or produce hallucinations, leading to excessive force or other hazardous actions.
Thus, safety and reliability of 3D embodied agents must be ensured through more detailed mechanisms, including the introduction of more physical constraints~\cite{zhang2024physdreamer}, real-time feedback systems, and human-value alignment~\cite{wang2023aligninglargelanguagemodels}. These measures are essential to achieving truly safe, reliable and trustworthy actions in complex 3D environments.

\section{Conclusion} \label{sec:conclusion}



This survey paper provides a thorough exploration of the integration of LLMs with 3D data. Systematically reviewing methodologies, applications, and emergent abilities of LLMs in processing, understanding, and generating 3D data, the survey underlines the transformative potential of LLMs across a spectrum of 3D tasks. Key findings include the identification of LLMs' unique advantages such as zero-shot learning, advanced reasoning, and extensive world knowledge, which are instrumental in bridging the gap between textual information and spatial interpretation. The paper showcases the wide array of tasks where LLMs' integration with 3D data has been successfully demonstrated. The exploration of other 3D vision-language methods alongside LLMs reveals a rich landscape of research aiming to deepen our understanding of the 3D world.

Furthermore, the survey highlights significant challenges such as data representation, model scalability, acquisition of spatial reasoning, data bottleneck and evaluation metrics, suggesting that overcoming these hurdles is crucial for the full realization of LLMs' potential in 3D applications. In conclusion, this survey not only offers a comprehensive overview of the current state of 3D tasks using LLMs but also sets the stage for future research directions. It calls for a collaborative effort to explore and expand the capabilities of LLMs in understanding and interacting with the complex 3D world, paving the way for further advancements in the area of spatial intelligence.


\renewcommand\refname{References}

{\scriptsize
\bibliographystyle{unsrt2authabbrvpp}
\bibliography{ref}

\begin{thebibliography}{100}

\bibitem{chen2023driving}
L.~Chen et~al.
\newblock Driving with llms: Fusing object-level vector modality for explainable autonomous driving.
\newblock {\em arXiv preprint arXiv:2310.01957}, 2023.

\bibitem{sha2023languagempc}
H.~Sha et~al.
\newblock Languagempc: Large language models as decision makers for autonomous driving.
\newblock {\em arXiv preprint arXiv:2310.03026}, 2023.

\bibitem{fu2024drive}
D.~Fu et~al.
\newblock Drive like a human: Rethinking autonomous driving with large language models.
\newblock In {\em WACV}, 2024.

\bibitem{xu2023drivegpt4}
Z.~Xu et~al.
\newblock Drivegpt4: Interpretable end-to-end autonomous driving via large language model.
\newblock {\em arXiv preprint arXiv:2310.01412}, 2023.

\bibitem{ma2022both}
X.~Ma et~al.
\newblock Both style and fog matter: Cumulative domain adaptation for semantic foggy scene understanding.
\newblock In {\em CVPR}, 2022.

\bibitem{azuma1997survey}
R.~T. Azuma.
\newblock A survey of augmented reality.
\newblock {\em Presence: teleoperators \& virtual environments}, 1997.

\bibitem{carmigniani2011augmented}
J.~Carmigniani and B.~Furht.
\newblock Augmented reality: an overview.
\newblock {\em Handbook of augmented reality}, 2011.

\bibitem{craig2013understanding}
A.~B. Craig.
\newblock {\em Understanding augmented reality: Concepts and applications}.
\newblock Newnes, 2013.

\bibitem{feiner1997touring}
S.~Feiner et~al.
\newblock A touring machine: Prototyping 3d mobile augmented reality systems for exploring the urban environment.
\newblock {\em Personal Technologies}, 1997.

\bibitem{rt22023arxiv}
A.~Brohan et~al.
\newblock Rt-2: Vision-language-action models transfer web knowledge to robotic control.
\newblock In {\em arXiv preprint arXiv:2307.15818}, 2023.

\bibitem{zheng2023towards}
D.~Zheng et~al.
\newblock Towards learning a generalist model for embodied navigation.
\newblock {\em arXiv preprint arXiv:2312.02010}, 2023.

\bibitem{song2023llmplanner}
C.~H. Song et~al.
\newblock Llm-planner: Few-shot grounded planning for embodied agents with large language models.
\newblock In {\em ICCV}, 2023.

\bibitem{huang2023voxposer}
W.~Huang et~al.
\newblock Voxposer: Composable 3d value maps for robotic manipulation with language models.
\newblock {\em arXiv preprint arXiv:2307.05973}, 2023.

\bibitem{mirjalili2023lan}
R.~Mirjalili et~al.
\newblock Lan-grasp: Using large language models for semantic object grasping.
\newblock {\em arXiv preprint arXiv:2310.05239}, 2023.

\bibitem{li2023manipllm}
X.~Li et~al.
\newblock Manipllm: Embodied multimodal large language model for object-centric robotic manipulation.
\newblock {\em arXiv preprint arXiv:2312.16217}, 2023.

\bibitem{yuan2023visual}
Z.~Yuan et~al.
\newblock Visual programming for zero-shot open-vocabulary 3d visual grounding.
\newblock {\em arXiv preprint arXiv:2311.15383}, 2023.

\bibitem{zhang2024agent3dzero}
S.~Zhang et~al.
\newblock Agent3d-zero: An agent for zero-shot 3d understanding, 2024.

\bibitem{jatavallabhula2023conceptfusion}
K.~M. Jatavallabhula et~al.
\newblock Conceptfusion: Open-set multimodal 3d mapping.
\newblock {\em arXiv preprint arXiv:2302.07241}, 2023.

\bibitem{chen2023ll3da}
S.~Chen et~al.
\newblock Ll3da: Visual interactive instruction tuning for omni-3d understanding, reasoning, and planning.
\newblock {\em arXiv preprint arXiv:2311.18651}, 2023.

\bibitem{guo2023viewrefer}
Z.~Guo et~al.
\newblock Viewrefer: Grasp the multi-view knowledge for 3d visual grounding with gpt and prototype guidance.
\newblock {\em arXiv preprint arXiv:2303.16894}, 2023.

\bibitem{llmgrounder}
J.~Yang et~al.
\newblock Llm-grounder: Open-vocabulary 3d visual grounding with large language model as an agent.
\newblock {\em arXiv preprint arXiv:2309.12311}, 2023.

\bibitem{abdelreheem2023zero}
A.~Abdelreheem et~al.
\newblock Zero-shot 3d shape correspondence.
\newblock In {\em SIGGRAPH Asia}, 2023.

\bibitem{fang2023transcribe3d}
J.~Fang et~al.
\newblock Transcribe3d: Grounding llms using transcribed information for 3d referential reasoning with self-corrected finetuning.
\newblock In {\em 2nd Workshop on Language and Robot Learning: Language as Grounding}, 2023.

\bibitem{hong2024multiply}
Y.~Hong et~al.
\newblock Multiply: A multisensory object-centric embodied large language model in 3d world.
\newblock {\em arXiv preprint arXiv:2401.08577}, 2024.

\bibitem{gala3d}
X.~Zhou et~al.
\newblock Gala3d: Towards text-to-3d complex scene generation via layout-guided generative gaussian splatting.
\newblock {\em arXiv preprint arXiv:2402.07207}, 2024.

\bibitem{3dgpt}
C.~Sun et~al.
\newblock 3d-gpt: Procedural 3d modeling with large language models.
\newblock {\em arXiv preprint arXiv:2310.12945}, 2023.

\bibitem{tang2024videounderstand}
Y.~Tang et~al.
\newblock Video understanding with large language models: A survey, 2024.

\bibitem{zeng2023robot1}
F.~Zeng et~al.
\newblock Large language models for robotics: A survey, 2023.

\bibitem{firoozi2023robot2}
R.~Firoozi et~al.
\newblock Foundation models in robotics: Applications, challenges, and the future, 2023.

\bibitem{hu2023robot3}
Y.~Hu et~al.
\newblock Toward general-purpose robots via foundation models: A survey and meta-analysis, 2023.

\bibitem{cui2023surveyAutonomousDriving}
C.~Cui et~al.
\newblock A survey on multimodal large language models for autonomous driving, 2023.

\bibitem{qi2017pointnet}
C.~R. Qi et~al.
\newblock Pointnet: Deep learning on point sets for 3d classification and segmentation.
\newblock In {\em CVPR}, 2017.

\bibitem{liu2019flownet3d}
X.~Liu et~al.
\newblock Flownet3d: Learning scene flow in 3d point clouds.
\newblock In {\em CVPR}, 2019.

\bibitem{dai2017shape}
A.~Dai et~al.
\newblock Shape completion using 3d-encoder-predictor cnns and shape synthesis.
\newblock In {\em CVPR}, 2017.

\bibitem{dai20183dmv}
A.~Dai and M.~Nie{\ss}ner.
\newblock 3dmv: Joint 3d-multi-view prediction for 3d semantic scene segmentation.
\newblock In {\em ECCV}, 2018.

\bibitem{Peng99}
D.~Peng et~al.
\newblock A pde-based fast local level set method.
\newblock {\em Journal of Computational Physics}, 1999.

\bibitem{Osher04}
S.~Osher et~al.
\newblock Level set methods and dynamic implicit surfaces.
\newblock {\em Applied Mechanics Reviews}, 2004.

\bibitem{Curless96}
B.~Curless and M.~Levoy.
\newblock A volumetric method for building complex models from range images.
\newblock In {\em the 23rd annual conference on Computer graphics and interactive techniques}, 1996.

\bibitem{newcombe2011kinectfusion}
R.~A. Newcombe et~al.
\newblock Kinectfusion: Real-time dense surface mapping and tracking.
\newblock In {\em ISMAR}, 2011.

\bibitem{riegler2017octnet}
G.~Riegler et~al.
\newblock Octnet: Learning deep 3d representations at high resolutions.
\newblock In {\em CVPR}, 2017.

\bibitem{Genova18}
K.~Genova et~al.
\newblock {Unsupervised Training for 3D Morphable Model Regression}.
\newblock In {\em CVPR}, 2018.

\bibitem{Kato19}
H.~Kato and T.~Harada.
\newblock Learning view priors for single-view 3d reconstruction.
\newblock In {\em CVPR}, 2019.

\bibitem{Rhodin15}
H.~Rhodin et~al.
\newblock {A Versatile Scene Model with Differentiable Visibility Applied to Generative Pose Estimation}.
\newblock In {\em ICCV}, 2015.

\bibitem{xie2022neural}
Y.~Xie et~al.
\newblock Neural fields in visual computing and beyond.
\newblock In {\em CGF}, 2022.

\bibitem{Mescheder19}
L.~Mescheder et~al.
\newblock Occupancy networks: Learning 3d reconstruction in function space.
\newblock In {\em CVPR}, 2019.

\bibitem{Peng20}
S.~Peng et~al.
\newblock Convolutional occupancy networks.
\newblock In {\em ECCV}, 2020.

\bibitem{Park19}
J.~J. Park et~al.
\newblock Deepsdf: Learning continuous signed distance functions for shape representation.
\newblock In {\em CVPR}, 2019.

\bibitem{Sitzmann19SRN}
V.~Sitzmann et~al.
\newblock Scene representation networks: Continuous 3{D}-structure aware neural scene representations.
\newblock In {\em NeurIPS}, 2019.

\bibitem{Mildenhall20}
B.~Mildenhall et~al.
\newblock {NeRF Representing scenes as neural radiance fields for view synthesis}.
\newblock In {\em ECCV}, 2020.

\bibitem{barron2022mip}
J.~T. Barron et~al.
\newblock Mip-nerf 360: Unbounded anti-aliased neural radiance fields.
\newblock In {\em CVPR}, 2022.

\bibitem{yu2022plenoxels}
A.~Yu et~al.
\newblock Plenoxels: Radiance fields without neural networks.
\newblock {\em CVPR}, 2022.

\bibitem{muller2022instant}
T.~M{\"u}ller et~al.
\newblock Instant neural graphics primitives with a multiresolution hash encoding.
\newblock {\em ACM TOG}, 2022.

\bibitem{kerbl3Dgaussians}
B.~Kerbl et~al.
\newblock 3d gaussian splatting for real-time radiance field rendering.
\newblock {\em ACM ToG}, 2023.

\bibitem{schoenberger2016sfm}
J.~L. Sch\"{o}nberger and J.-M. Frahm.
\newblock Structure-from-motion revisited.
\newblock In {\em CVPR}, 2016.

\bibitem{Fu_2024_CVPR}
Y.~Fu et~al.
\newblock Colmap-free 3d gaussian splatting.
\newblock In {\em CVPR}, 2024.

\bibitem{fan2024instantsplat}
Z.~Fan et~al.
\newblock Instantsplat: Unbounded sparse-view pose-free gaussian splatting in 40 seconds.
\newblock {\em arXiv preprint arXiv:2403.20309}, 2024.

\bibitem{dai2017scannet}
A.~Dai et~al.
\newblock Scannet: Richly-annotated 3d reconstructions of indoor scenes.
\newblock In {\em CVPR}, 2017.

\bibitem{scanqa}
D.~Azuma et~al.
\newblock Scanqa: 3d question answering for spatial scene understanding.
\newblock In {\em CVPR}, 2022.

\bibitem{ma2022sqa3d}
X.~Ma et~al.
\newblock Sqa3d: Situated question answering in 3d scenes.
\newblock {\em arXiv preprint arXiv:2210.07474}, 2022.

\bibitem{elman1990finding}
J.~L. Elman.
\newblock Finding structure in time.
\newblock {\em Cognitive science}, 1990.

\bibitem{Vaswani17}
A.~Vaswani et~al.
\newblock Attention is all you need.
\newblock In {\em NeurIPS}, 2017.

\bibitem{kaplan2020scaling}
J.~Kaplan et~al.
\newblock Scaling laws for neural language models.
\newblock {\em arXiv preprint arXiv:2001.08361}, 2020.

\bibitem{zhao2023survey}
W.~X. Zhao et~al.
\newblock A survey of large language models.
\newblock {\em arXiv preprint arXiv:2303.18223}, 2023.

\bibitem{minaee2024large}
S.~Minaee et~al.
\newblock Large language models: A survey.
\newblock {\em arXiv preprint arXiv:2402.06196}, 2024.

\bibitem{wei2022emergent}
J.~Wei et~al.
\newblock Emergent abilities of large language models.
\newblock {\em arXiv preprint arXiv:2206.07682}, 2022.

\bibitem{liu2018generating}
P.~J. Liu et~al.
\newblock Generating wikipedia by summarizing long sequences.
\newblock {\em arXiv preprint arXiv:1801.10198}, 2018.

\bibitem{bytepairencoding}
R.~Sennrich et~al.
\newblock Neural machine translation of rare words with subword units.
\newblock {\em arXiv preprint arXiv:1508.07909}, 2015.

\bibitem{wordpiece}
M.~Schuster and K.~Nakajima.
\newblock Japanese and korean voice search.
\newblock In {\em ICASSP}, 2012.

\bibitem{kudo2018sentencepiece}
T.~Kudo and J.~Richardson.
\newblock Sentencepiece: A simple and language independent subword tokenizer and detokenizer for neural text processing.
\newblock {\em arXiv preprint arXiv:1808.06226}, 2018.

\bibitem{brown2020language}
T.~Brown et~al.
\newblock Language models are few-shot learners.
\newblock {\em NeurIPS}, 2020.

\bibitem{wei2022chain}
J.~Wei et~al.
\newblock Chain-of-thought prompting elicits reasoning in large language models.
\newblock {\em NeurIPS}, 2022.

\bibitem{touvron2023llama}
H.~Touvron et~al.
\newblock Llama: Open and efficient foundation language models.
\newblock {\em arXiv preprint arXiv:2302.13971}, 2023.

\bibitem{hu2021lora}
E.~J. Hu et~al.
\newblock Lora: Low-rank adaptation of large language models.
\newblock {\em arXiv preprint arXiv:2106.09685}, 2021.

\bibitem{dettmers2024qlora}
T.~Dettmers et~al.
\newblock Qlora: Efficient finetuning of quantized llms.
\newblock {\em NeurIPS}, 36, 2024.

\bibitem{zhang2023lora}
L.~Zhang et~al.
\newblock Lora-fa: Memory-efficient low-rank adaptation for large language models fine-tuning.
\newblock {\em arXiv preprint arXiv:2308.03303}, 2023.

\bibitem{devlin2018bert}
J.~Devlin et~al.
\newblock Bert: Pre-training of deep bidirectional transformers for language understanding.
\newblock {\em arXiv preprint arXiv:1810.04805}, 2018.

\bibitem{howard2018universal}
J.~Howard and S.~Ruder.
\newblock Universal language model fine-tuning for text classification.
\newblock In {\em ACL}, 2018.

\bibitem{shin2020autoprompt}
T.~Shin et~al.
\newblock Autoprompt: Eliciting knowledge from language models with automatically generated prompts.
\newblock {\em arXiv preprint arXiv:2010.15980}, 2020.

\bibitem{lester2021power}
B.~Lester et~al.
\newblock The power of scale for parameter-efficient prompt tuning.
\newblock In {\em EMNLP}, 2021.

\bibitem{chat3d}
Z.~Wang et~al.
\newblock Chat-3d: Data-efficiently tuning large language model for universal dialogue of 3d scenes, 2023.

\bibitem{huang2023chatV2}
H.~Huang et~al.
\newblock Chat-3d v2: Bridging 3d scene and large language models with object identifiers.
\newblock {\em arXiv preprint arXiv:2312.08168}, 2023.

\bibitem{radford2021learning}
A.~Radford et~al.
\newblock Learning transferable visual models from natural language supervision.
\newblock In {\em ICML}, 2021.

\bibitem{jia2021scaling}
C.~Jia et~al.
\newblock Scaling up visual and vision-language representation learning with noisy text supervision.
\newblock In {\em ICML}, 2021.

\bibitem{zhong2022regionclip}
Y.~Zhong et~al.
\newblock Regionclip: Region-based language-image pretraining.
\newblock In {\em CVPR}, 2022.

\bibitem{luddecke2022clipseg}
T.~L{\"u}ddecke and A.~Ecker.
\newblock Image segmentation using text and image prompts.
\newblock In {\em CVPR}, 2022.

\bibitem{ni2022xclip}
B.~Ni et~al.
\newblock Expanding language-image pretrained models for general video recognition.
\newblock In {\em ECCV}, 2022.

\bibitem{li2023blip2}
J.~Li et~al.
\newblock Blip-2: Bootstrapping language-image pre-training with frozen image encoders and large language models.
\newblock {\em arXiv preprint arXiv:2301.12597}, 2023.

\bibitem{alayrac2022flamingo}
J.-B. Alayrac et~al.
\newblock Flamingo: a visual language model for few-shot learning.
\newblock {\em NeurIPS}, 2022.

\bibitem{liu2024llava}
H.~Liu et~al.
\newblock Visual instruction tuning.
\newblock In {\em NeurIPS}, 2024.

\bibitem{rombach2022sd}
R.~Rombach et~al.
\newblock High-resolution image synthesis with latent diffusion models.
\newblock In {\em CVPR}, 2022.

\bibitem{ho2022imagenvideo}
J.~Ho et~al.
\newblock Imagen video: High definition video generation with diffusion models.
\newblock {\em arXiv preprint arXiv:2210.02303}, 2022.

\bibitem{poole2022dreamfusion}
B.~Poole et~al.
\newblock Dreamfusion: Text-to-3d using 2d diffusion.
\newblock {\em arXiv preprint arXiv:2209.14988}, 2022.

\bibitem{singer2023text24d}
U.~Singer et~al.
\newblock Text-to-4d dynamic scene generation.
\newblock {\em arXiv preprint arXiv:2301.11280}, 2023.

\bibitem{caron2021emerging}
M.~Caron et~al.
\newblock Emerging properties in self-supervised vision transformers.
\newblock In {\em ICCV}, 2021.

\bibitem{zhou2021ibot}
J.~Zhou et~al.
\newblock ibot: Image bert pre-training with online tokenizer.
\newblock In {\em ICLR}, 2022.

\bibitem{oquab2023dinov2}
M.~Oquab et~al.
\newblock Dinov2: Learning robust visual features without supervision.
\newblock {\em arXiv preprint arXiv:2304.07193}, 2023.

\bibitem{kirillov2023segment}
A.~Kirillov et~al.
\newblock Segment anything.
\newblock {\em arXiv preprint arXiv:2304.02643}, 2023.

\bibitem{tang2024dift}
L.~Tang et~al.
\newblock Emergent correspondence from image diffusion.
\newblock {\em NeurIPS}, 2024.

\bibitem{tian2023diffuse}
J.~Tian et~al.
\newblock Diffuse, attend, and segment: Unsupervised zero-shot segmentation using stable diffusion.
\newblock {\em arXiv preprint arXiv:2308.12469}, 2023.

\bibitem{chen2021scan2cap}
Z.~Chen et~al.
\newblock Scan2cap: Context-aware dense captioning in rgb-d scans.
\newblock In {\em CVPR}, 2021.

\bibitem{scanrefer}
D.~Z. Chen et~al.
\newblock Scanrefer: 3d object localization in rgb-d scans using natural language.
\newblock {\em ECCV}, 2020.

\bibitem{celikyilmaz2020evaluation}
A.~Celikyilmaz et~al.
\newblock Evaluation of text generation: A survey.
\newblock {\em arXiv preprint arXiv:2006.14799}, 2020.

\bibitem{papineni2002bleu}
K.~Papineni et~al.
\newblock Bleu: a method for automatic evaluation of machine translation.
\newblock In {\em ACL}, 2002.

\bibitem{lin2004rouge}
C.-Y. Lin.
\newblock Rouge: A package for automatic evaluation of summaries.
\newblock In {\em Text summarization branches out}, 2004.

\bibitem{banerjee2005meteor}
S.~Banerjee and A.~Lavie.
\newblock Meteor: An automatic metric for mt evaluation with improved correlation with human judgments.
\newblock In {\em ACLW}, 2005.

\bibitem{vedantam2015cider}
R.~Vedantam et~al.
\newblock {CIDE}r: Consensus-based image description evaluation.
\newblock In {\em CVPR}, 2015.

\bibitem{reimers2019sentence}
N.~Reimers and I.~Gurevych.
\newblock Sentence-bert: Sentence embeddings using siamese bert-networks.
\newblock {\em arXiv preprint arXiv:1908.10084}, 2019.

\bibitem{zhang2019bertscore}
T.~Zhang et~al.
\newblock {BERTS}core: Evaluating text generation with bert.
\newblock {\em arXiv preprint arXiv:1904.09675}, 2019.

\bibitem{chen2023end}
S.~Chen et~al.
\newblock End-to-end 3d dense captioning with vote2cap-detr.
\newblock In {\em CVPR}, 2023.

\bibitem{chen2023unit3d}
Z.~Chen et~al.
\newblock Unit3d: A unified transformer for 3d dense captioning and visual grounding.
\newblock In {\em ICCV}, 2023.

\bibitem{achlioptas2020referit3d}
P.~Achlioptas et~al.
\newblock Referit3d: Neural listeners for fine-grained 3d object identification in real-world scenes.
\newblock In {\em ECCV}, 2020.

\bibitem{ScanERU}
Z.~Lu et~al.
\newblock Scaneru: Interactive 3d visual grounding based on embodied reference understanding.
\newblock {\em arXiv preprint arXiv:2303.13186}, 2023.

\bibitem{zhang2023multi3drefer}
Y.~Zhang et~al.
\newblock Multi3drefer: Grounding text description to multiple 3d objects.
\newblock In {\em ICCV}, 2023.

\bibitem{densegrounding}
W.~Huang et~al.
\newblock Dense object grounding in 3d scenes.
\newblock In {\em ACM MM}, 2023.

\bibitem{3dllm}
Y.~Hong et~al.
\newblock 3d-llm: Injecting the 3d world into large language models.
\newblock {\em NeurIPS}, 2023.

\bibitem{TGNN}
P.-H. Huang et~al.
\newblock Text-guided graph neural networks for referring 3d instance segmentation.
\newblock In {\em AAAI}, 2021.

\bibitem{qian2023nuscenes}
T.~Qian et~al.
\newblock Nuscenes-qa: A multi-modal visual question answering benchmark for autonomous driving scenario.
\newblock {\em arXiv preprint arXiv:2305.14836}, 2023.

\bibitem{hong20233d}
Y.~Hong et~al.
\newblock 3d concept learning and reasoning from multi-view images.
\newblock In {\em CVPR}, 2023.

\bibitem{anderson2016spice}
P.~Anderson et~al.
\newblock {SPICE}: Semantic propositional image caption evaluation.
\newblock In {\em ECCV}, 2016.

\bibitem{anderson2018evaluation}
P.~Anderson et~al.
\newblock On evaluation of embodied navigation agents.
\newblock {\em arXiv preprint arXiv:1807.06757}, 2018.

\bibitem{anderson2018vision}
P.~Anderson et~al.
\newblock Vision-and-language navigation: Interpreting visually-grounded navigation instructions in real environments.
\newblock In {\em CVPR}, 2018.

\bibitem{gu2022vln}
J.~Gu et~al.
\newblock Vision-and-language navigation: A survey of tasks, methods, and future directions.
\newblock {\em arXiv preprint arXiv:2203.12667}, 2022.

\bibitem{shridhar2022cliport}
M.~Shridhar et~al.
\newblock Cliport: What and where pathways for robotic manipulation.
\newblock In {\em CoRL}. PMLR, 2022.

\bibitem{lee2024text}
H.-H. Lee et~al.
\newblock Text-to-3d shape generation.
\newblock {\em arXiv preprint arXiv:2403.13289}, 2024.

\bibitem{xue2023ulip}
L.~Xue et~al.
\newblock Ulip: Learning a unified representation of language, images, and point clouds for 3d understanding.
\newblock In {\em CVPR}, 2023.

\bibitem{wu2024gpt}
T.~Wu et~al.
\newblock Gpt-4v (ision) is a human-aligned evaluator for text-to-3d generation.
\newblock {\em arXiv preprint arXiv:2401.04092}, 2024.

\bibitem{armeni20163d}
I.~Armeni et~al.
\newblock 3d semantic parsing of large-scale indoor spaces.
\newblock In {\em CVPR}, 2016.

\bibitem{sengupta2013urban}
S.~Sengupta et~al.
\newblock Urban 3d semantic modelling using stereo vision.
\newblock In {\em ICRA}. IEEE, 2013.

\bibitem{mccormac2017semanticfusion}
J.~McCormac et~al.
\newblock Semanticfusion: Dense 3d semantic mapping with convolutional neural networks.
\newblock In {\em ICRA}. IEEE, 2017.

\bibitem{HuangZhan2020PartAssembly}
J.~Huang et~al.
\newblock Generative 3d part assembly via dynamic graph learning.
\newblock In {\em NeurIPS}, 2020.

\bibitem{cheng2023score}
J.~Cheng et~al.
\newblock Score-pa: Score-based 3d part assembly.
\newblock {\em BMVC}, 2023.

\bibitem{jiang2020pointgroup}
L.~Jiang et~al.
\newblock Pointgroup: Dual-set point grouping for 3d instance segmentation.
\newblock In {\em CVPR}, 2020.

\bibitem{hou20193d}
J.~Hou et~al.
\newblock 3d-sis: 3d semantic instance segmentation of rgb-d scans.
\newblock In {\em CVPR}, 2019.

\bibitem{wang2018sgpn}
W.~Wang et~al.
\newblock Sgpn: Similarity group proposal network for 3d point cloud instance segmentation.
\newblock In {\em CVPR}, 2018.

\bibitem{han2020occuseg}
L.~Han et~al.
\newblock Occuseg: Occupancy-aware 3d instance segmentation.
\newblock In {\em CVPR}, 2020.

\bibitem{song2019apollocar3d}
X.~Song et~al.
\newblock Apollocar3d: A large 3d car instance understanding benchmark for autonomous driving.
\newblock In {\em CVPR}, 2019.

\bibitem{zhan2023amodal}
G.~Zhan et~al.
\newblock Amodal ground truth and completion in the wild.
\newblock In {\em CVPR}, 2024.

\bibitem{Zhan2023physd}
G.~Zhan et~al.
\newblock What does stable diffusion know about the 3d scene?
\newblock In {\em arXiv:2310.06836}, 2023.

\bibitem{feng2023exploring}
M.~Feng et~al.
\newblock Exploring hierarchical spatial layout cues for 3d point cloud based scene graph prediction.
\newblock {\em IEEE TMM}, 2023.

\bibitem{zhang2021holistic}
C.~Zhang et~al.
\newblock Holistic 3d scene understanding from a single image with implicit representation.
\newblock In {\em CVPR}, 2021.

\bibitem{zhang2021deeppanocontext}
C.~Zhang et~al.
\newblock Deeppanocontext: Panoramic 3d scene understanding with holistic scene context graph and relation-based optimization.
\newblock In {\em ICCV}, 2021.

\bibitem{zhan2022triocc}
G.~Zhan et~al.
\newblock A tri-layer plugin to improve occluded detection.
\newblock {\em BMVC}, 2022.

\bibitem{SceneFun3D}
A.~Delitzas et~al.
\newblock Scenefun3d: Fine-grained functionality and affordance understanding in 3d scenes.
\newblock In {\em CVPR}, 2024.

\bibitem{cheng2023occlusion}
K.~Cheng et~al.
\newblock Learning environment-aware affordance for 3d articulated object manipulation under occlusions.
\newblock In {\em NeurIPS}, 2023.

\bibitem{qiu20203d}
Y.~Qiu et~al.
\newblock 3d-aware scene change captioning from multiview images.
\newblock {\em IEEE Robotics and Automation Letters}, 2020.

\bibitem{looper20233d}
S.~Looper et~al.
\newblock 3d vsg: Long-term semantic scene change prediction through 3d variable scene graphs.
\newblock In {\em ICRA}. IEEE, 2023.

\bibitem{fu2024scenellm}
R.~Fu et~al.
\newblock Scene-llm: Extending language model for 3d visual understanding and reasoning, 2024.

\bibitem{chen2022leveraging}
W.~Chen et~al.
\newblock Leveraging large language models for robot 3d scene understanding.
\newblock {\em arXiv preprint arXiv:2209.05629}, 2022.

\bibitem{liu20233daxiesprompts}
D.~Liu et~al.
\newblock 3daxiesprompts: Unleashing the 3d spatial task capabilities of gpt-4v.
\newblock {\em arXiv preprint arXiv:2312.09738}, 2023.

\bibitem{ma2024spatialpin}
C.~Ma et~al.
\newblock See, imagine, plan: Discovering and hallucinating tasks from a single image, 2024.

\bibitem{man2024SIG3D}
Y.~Man et~al.
\newblock Situational awareness matters in 3d vision language reasoning.
\newblock In {\em CVPR}, 2024.

\bibitem{LEO}
J.~Huang et~al.
\newblock An embodied generalist agent in 3d world.
\newblock In {\em ICML}, 2024.

\bibitem{xu2023pointllm}
R.~Xu et~al.
\newblock Pointllm: Empowering large language models to understand point clouds.
\newblock {\em ECCV}, 2024.

\bibitem{qi2023gpt4point}
Z.~Qi et~al.
\newblock Gpt4point: A unified framework for point-language understanding and generation.
\newblock {\em arXiv preprint arXiv:2312.02980}, 2023.

\bibitem{yang2023lidar}
S.~Yang et~al.
\newblock Lidar-llm: Exploring the potential of large language models for 3d lidar understanding.
\newblock {\em arXiv preprint arXiv:2312.14074}, 2023.

\bibitem{huang2024chatscene}
H.~Huang et~al.
\newblock Chat-scene: Bridging 3d scene and large language models with object identifiers.
\newblock {\em NeurIPS}, 2024.

\bibitem{li20243dmit}
Z.~Li et~al.
\newblock 3dmit: 3d multi-modal instruction tuning for scene understanding.
\newblock {\em arXiv preprint arXiv:2401.03201}, 2024.

\bibitem{qi2024shapellm}
Z.~Qi et~al.
\newblock Shapellm: Universal 3d object understanding for embodied interaction.
\newblock {\em arXiv preprint arXiv:2402.17766}, 2024.

\bibitem{tang2024minigpt}
Y.~Tang et~al.
\newblock Minigpt-3d: Efficiently aligning 3d point clouds with large language models using 2d priors.
\newblock {\em arXiv preprint arXiv:2405.01413}, 2024.

\bibitem{amaduzzi2024llana}
A.~Amaduzzi et~al.
\newblock Llana: Large language and nerf assistant.
\newblock {\em NeurIPS}, 2024.

\bibitem{tang2024greenPLM}
Y.~Tang et~al.
\newblock More text, less point: Towards 3d data-efficient point-language understanding, 2024.

\bibitem{zhu2024llava3d}
C.~Zhu et~al.
\newblock Llava-3d: A simple yet effective pathway to empowering lmms with 3d-awareness.
\newblock {\em arXiv preprint arXiv:2409.18125}, 2024.

\bibitem{kang2025robin3d}
W.~Kang et~al.
\newblock Robin3d: Improving 3d large language model via robust instruction tuning, 2025.

\bibitem{mei2025PerLA}
G.~Mei et~al.
\newblock Perla: Perceptive 3d language assistant.
\newblock In {\em CVPR}, 2025.

\bibitem{zheng2025video3dllm}
D.~Zheng et~al.
\newblock Video-3d llm: Learning position-aware video representation for 3d scene understanding.
\newblock In {\em CVPR}, 2025.

\bibitem{qi2025gpt4scene}
Z.~Qi et~al.
\newblock Gpt4scene: Understand 3d scenes from videos with vision-language models.
\newblock {\em arXiv preprint arXiv:2501.01428}, 2025.

\bibitem{zhi2025lscenellm}
H.~Zhi et~al.
\newblock Lscenellm: Enhancing large 3d scene understanding using adaptive visual preferences.
\newblock In {\em CVPR}, 2025.

\bibitem{yu2025inst3d}
H.~Yu et~al.
\newblock Inst3d-lmm: Instance-aware 3d scene understanding with multi-modal instruction tuning.
\newblock In {\em CVPR}, 2025.

\bibitem{thai2025splattalk}
A.~Thai et~al.
\newblock Splattalk: 3d vqa with gaussian splatting.
\newblock {\em arXiv preprint arXiv:2503.06271}, 2025.

\bibitem{wang2025ross3d}
H.~Wang et~al.
\newblock Ross3d: Reconstructive visual instruction tuning with 3d-awareness.
\newblock {\em arXiv preprint arXiv:2504.01901}, 2025.

\bibitem{deng20253dllave}
J.~Deng et~al.
\newblock 3d-llava: Towards generalist 3d lmms with omni superpoint transformer.
\newblock {\em arXiv preprint arXiv:2501.01163}, 2025.

\bibitem{ma2025spatialllm}
W.~Ma et~al.
\newblock Spatialllm: A compound 3d-informed design towards spatially-intelligent large multimodal models.
\newblock In {\em CVPR}, 2025.

\bibitem{zheng2025VGLLM}
D.~Zheng et~al.
\newblock Learning from videos for 3d world: Enhancing mllms with 3d vision geometry priors, 2025.

\bibitem{fan2025vlm3r}
Z.~Fan et~al.
\newblock Vlm-3r: Vision-language models augmented with instruction-aligned 3d reconstruction.
\newblock {\em arXiv preprint arXiv:2505.20279}, 2025.

\bibitem{huang2025leovl}
J.~Huang et~al.
\newblock Leo-vl: Towards 3d vision-language generalists via data scaling with efficient representation.
\newblock {\em arXiv preprint arXiv:2506.09935}, 2025.

\bibitem{huang20253DRS}
X.~Huang et~al.
\newblock Mllms need 3d-aware representation supervision for scene understanding.
\newblock {\em arXiv preprint arXiv:2506.01946}, 2025.

\bibitem{guo2023point}
Z.~Guo et~al.
\newblock Point-bind \& point-llm: Aligning point cloud with multi-modality for 3d understanding, generation, and instruction following.
\newblock {\em arXiv preprint arXiv:2309.00615}, 2023.

\bibitem{wang2023beyond}
H.~Wang et~al.
\newblock Beyond first impressions: Integrating joint multi-modal cues for comprehensive 3d representation.
\newblock In {\em ACM MM}, 2023.

\bibitem{hu20253dllm-mem}
W.~Hu et~al.
\newblock 3dllm-mem: Long-term spatial-temporal memory for embodied 3d large language model.
\newblock {\em arXiv}, 2025.

\bibitem{rana2023sayplan}
K.~Rana et~al.
\newblock Sayplan: Grounding large language models using 3d scene graphs for scalable task planning.
\newblock {\em arXiv preprint arXiv:2307.06135}, 2023.

\bibitem{xiao2023unified}
Z.~Xiao et~al.
\newblock Unified human-scene interaction via prompted chain-of-contacts.
\newblock {\em arXiv preprint arXiv:2309.07918}, 2023.

\bibitem{zhen20243d}
H.~Zhen et~al.
\newblock 3d-vla: A 3d vision-language-action generative world model.
\newblock {\em arXiv preprint arXiv:2403.09631}, 2024.

\bibitem{nash2020polygen}
C.~Nash et~al.
\newblock Polygen: An autoregressive generative model of 3d meshes.
\newblock In {\em ICML}, 2020.

\bibitem{de2023llmr}
F.~De~La~Torre et~al.
\newblock Llmr: Real-time prompting of interactive worlds using large language models.
\newblock {\em arXiv preprint arXiv:2309.12276}, 2023.

\bibitem{siddiqui2023meshgpt}
Y.~Siddiqui et~al.
\newblock Meshgpt: Generating triangle meshes with decoder-only transformers.
\newblock {\em arXiv preprint arXiv:2311.15475}, 2023.

\bibitem{shapegpt}
F.~Yin et~al.
\newblock Shapegpt: 3d shape generation with a unified multi-modal language model.
\newblock {\em arXiv preprint arXiv:2311.17618}, 2023.

\bibitem{yang2024holodeck}
Y.~Yang et~al.
\newblock Holodeck: Language guided generation of 3d embodied ai environments.
\newblock In {\em CVPR}, 2024.

\bibitem{wang2024llamamesh}
Z.~Wang et~al.
\newblock Llama-mesh: Unifying 3d mesh generation with language models.
\newblock {\em arXiv preprint arXiv:2411.09595}, 2024.

\bibitem{li-etal-2022-systematic}
X.~L. Li et~al.
\newblock A systematic investigation of commonsense knowledge in large language models.
\newblock In Y.~Goldberg et~al., editors, {\em EMNLP}, 2022.

\bibitem{Matterport3D}
A.~Chang et~al.
\newblock Matterport3d: Learning from rgb-d data in indoor environments.
\newblock {\em 3DV}, 2017.

\bibitem{chatgpt}
Chatgpt.
\newblock \url{https://openai.com/blog/chatgpt}.
\newblock Accessed: 2023-07-22.

\bibitem{he2021transrefer3d}
D.~He et~al.
\newblock Transrefer3d: Entity-and-relation aware transformer for fine-grained 3d visual grounding.
\newblock In {\em ACM MM}, 2021.

\bibitem{zhang2024spartun3d}
Y.~Zhang et~al.
\newblock Spartun3d: Situated spatial understanding of 3d world in large language models.
\newblock {\em arXiv preprint arXiv:2410.03878}, 2024.

\bibitem{yu2022point}
X.~Yu et~al.
\newblock Point-bert: Pre-training 3d point cloud transformers with masked point modeling.
\newblock In {\em CVPR}, 2022.

\bibitem{qi2017pointnet++}
C.~R. Qi et~al.
\newblock Pointnet++: Deep hierarchical feature learning on point sets in a metric space.
\newblock {\em NeurIPS}, 30, 2017.

\bibitem{fei2024kestrel}
J.~Fei et~al.
\newblock Kestrel: Point grounding multimodal llm for part-aware 3d vision-language understanding, 2024.

\bibitem{takmaz2023openmask3d}
A.~Takmaz et~al.
\newblock {OpenMask3D: Open-Vocabulary 3D Instance Segmentation}.
\newblock In {\em NeurIPS}, 2023.

\bibitem{zhou2018voxelnet}
Y.~Zhou and O.~Tuzel.
\newblock Voxelnet: End-to-end learning for point cloud based 3d object detection.
\newblock In {\em CVPR}, 2018.

\bibitem{jin2025revisiting}
J.~Jin et~al.
\newblock Revisiting 3d llm benchmarks: Are we really testing 3d capabilities?
\newblock {\em arXiv preprint arXiv:2502.08503}, 2025.

\bibitem{wang2025vggt}
J.~Wang et~al.
\newblock Vggt: Visual geometry grounded transformer.
\newblock In {\em CVPR}, 2025.

\bibitem{objaverse}
M.~Deitke et~al.
\newblock Objaverse: A universe of annotated 3d objects.
\newblock In {\em CVPR}, 2023.

\bibitem{zhu20233d}
Z.~Zhu et~al.
\newblock 3d-vista: Pre-trained transformer for 3d vision and text alignment.
\newblock In {\em ICCV}, 2023.

\bibitem{Wald2019RIO}
J.~Wald et~al.
\newblock Rio: 3d object instance re-localization in changing indoor environments.
\newblock In {\em ICCV}, 2019.

\bibitem{chang2015shapenet}
A.~X. Chang et~al.
\newblock Shapenet: An information-rich 3d model repository.
\newblock {\em arXiv preprint arXiv:1512.03012}, 2015.

\bibitem{uy-scanobjectnn-iccv19}
M.~A. Uy et~al.
\newblock Revisiting point cloud classification: A new benchmark dataset and classification model on real-world data.
\newblock In {\em ICCV}, 2019.

\bibitem{zhang2023instruction}
S.~Zhang et~al.
\newblock Instruction tuning for large language models: A survey.
\newblock {\em arXiv preprint arXiv:2308.10792}, 2023.

\bibitem{beaumont2022clip}
R.~Beaumont.
\newblock Clip retrieval: Easily compute clip embeddings and build a clip retrieval system with them, 2022.

\bibitem{guzhov2022audioclip}
A.~Guzhov et~al.
\newblock Audioclip: Extending clip to image, text and audio.
\newblock In {\em ICASSP}, 2022.

\bibitem{mu2022slip}
N.~Mu et~al.
\newblock Slip: Self-supervision meets language-image pre-training.
\newblock In {\em ECCV}, 2022.

\bibitem{girdhar2023imagebind}
R.~Girdhar et~al.
\newblock Imagebind: One embedding space to bind them all.
\newblock In {\em CVPR}, 2023.

\bibitem{li2022lseg}
B.~Li et~al.
\newblock Language-driven semantic segmentation.
\newblock In {\em ICLR}, 2022.

\bibitem{ghiasi2021openseg}
G.~Ghiasi et~al.
\newblock Scaling open-vocabulary image segmentation with image-level labels.
\newblock In {\em ECCV}, 2022.

\bibitem{gu2023conceptgraphs}
Q.~Gu et~al.
\newblock Conceptgraphs: Open-vocabulary 3d scene graphs for perception and planning.
\newblock {\em arXiv preprint arXiv:2309.16650}, 2023.

\bibitem{chen2022nlmapsaycan}
B.~Chen et~al.
\newblock Open-vocabulary queryable scene representations for real world planning.
\newblock In {\em arXiv preprint arXiv:2209.09874}, 2022.

\bibitem{2020t5}
C.~Raffel et~al.
\newblock Exploring the limits of transfer learning with a unified text-to-text transformer.
\newblock {\em Journal of Machine Learning Research (JMLR)}, 2020.

\bibitem{hu2024scenecraft}
Z.~Hu et~al.
\newblock {SceneCraft}: An {LLM} agent for synthesizing {3D} scene as {Blender} code.
\newblock In {\em ICLRW}, 2024.

\bibitem{Peng2023OpenScene}
S.~Peng et~al.
\newblock Openscene: 3d scene understanding with open vocabularies.
\newblock In {\em CVPR}, 2023.

\bibitem{zhang2023clipfo3d}
J.~Zhang et~al.
\newblock Clip-fo3d: Learning free open-world 3d scene representations from 2d dense clip.
\newblock In {\em ICCV}, 2023.

\bibitem{semanticabstraction}
H.~Ha and S.~Song.
\newblock {S}emantic {A}bstraction: Open-world 3{D} scene understanding from 2{D} vision-language models.
\newblock In {\em CoRL}, 2022.

\bibitem{kashu2023openfusion}
K.~Yamazaki et~al.
\newblock Open-fusion: Real-time open-vocabulary 3d mapping and queryable scene representation.
\newblock {\em arXiv preprint arXiv:2310.03923}, 2023.

\bibitem{zou2024segment}
X.~Zou et~al.
\newblock Segment everything everywhere all at once.
\newblock {\em NeurIPS}, 36, 2024.

\bibitem{ding2022language}
R.~Ding et~al.
\newblock Pla: Language-driven open-vocabulary 3d scene understanding.
\newblock In {\em CVPR}, 2023.

\bibitem{yang2023regionplc}
J.~Yang et~al.
\newblock Regionplc: Regional point-language contrastive learning for open-world 3d scene understanding.
\newblock {\em arXiv preprint arXiv:2304.00962}, 2023.

\bibitem{lu2023ovir3d}
S.~Lu et~al.
\newblock Ovir-3d: Open-vocabulary 3d instance retrieval without training on 3d data.
\newblock In {\em CoRL}, 2023.

\bibitem{cao2023coda}
Y.~Cao et~al.
\newblock Coda: Collaborative novel box discovery and cross-modal alignment for open-vocabulary 3d object detection.
\newblock {\em NeurIPS}, 36, 2023.

\bibitem{nguyen2023open3dis}
P.~D. Nguyen et~al.
\newblock Open3dis: Open-vocabulary 3d instance segmentation with 2d mask guidance.
\newblock {\em arXiv preprint arXiv:2312.10671}, 2023.

\bibitem{huang2023openins3d}
Z.~Huang et~al.
\newblock Openins3d: Snap and lookup for 3d open-vocabulary instance segmentation.
\newblock {\em arXiv preprint}, 2023.

\bibitem{rozenberszki2022language}
D.~Rozenberszki et~al.
\newblock Language-grounded indoor 3d semantic segmentation in the wild.
\newblock In {\em ECCV}, 2022.

\bibitem{zhu2024open}
X.~Zhu et~al.
\newblock Open-vocabulary 3d semantic segmentation with text-to-image diffusion models.
\newblock In {\em ECCV}, pp. 357--375. Springer, 2024.

\bibitem{kobayashi2022distilledfeaturefields}
S.~Kobayashi et~al.
\newblock Decomposing nerf for editing via feature field distillation.
\newblock {\em NeurIPS}, 2022.

\bibitem{lerf2023}
J.~Kerr et~al.
\newblock Lerf: Language embedded radiance fields.
\newblock In {\em ICCV}, 2023.

\bibitem{tsagkas2023vlfields}
N.~Tsagkas et~al.
\newblock Vl-fields: Towards language-grounded neural implicit spatial representations.
\newblock {\em arXiv preprint arXiv:2305.12427}, 2023.

\bibitem{liu20243dovs}
K.~Liu et~al.
\newblock Weakly supervised 3d open-vocabulary segmentation.
\newblock {\em NeurIPS}, 36, 2024.

\bibitem{qin2023langsplat}
M.~Qin et~al.
\newblock Langsplat: 3d language gaussian splatting.
\newblock {\em arXiv preprint arXiv:2312.16084}, 2023.

\bibitem{bhalgat2024n2f2}
Y.~Bhalgat et~al.
\newblock N2f2: Hierarchical scene understanding with nested neural feature fields.
\newblock In {\em ECCV}, 2024.

\bibitem{rombach2022high}
R.~Rombach et~al.
\newblock High-resolution image synthesis with latent diffusion models.
\newblock In {\em CVPR}, 2022.

\bibitem{saharia2022imagen}
C.~Saharia et~al.
\newblock Photorealistic text-to-image diffusion models with deep language understanding.
\newblock {\em NeurIPS}, 2022.

\bibitem{jain2022dreamfields}
A.~Jain et~al.
\newblock Zero-shot text-guided object generation with dream fields.
\newblock In {\em CVPR}, 2022.

\bibitem{khalid2022clipmesh}
N.~M. Khalid et~al.
\newblock Clip-mesh: Generating textured meshes from text using pretrained image-text models.
\newblock {\em arXiv preprint arXiv:2203.13333}, 2022.

\bibitem{sanghi2022clipforge}
A.~Sanghi et~al.
\newblock Clip-forge: Towards zero-shot text-to-shape generation.
\newblock In {\em CVPR}, 2022.

\bibitem{michel2022text2mesh}
O.~Michel et~al.
\newblock Text2mesh: Text-driven neural stylization for meshes.
\newblock In {\em CVPR}, 2022.

\bibitem{lin2022magic3d}
C.-H. Lin et~al.
\newblock Magic3d: High-resolution text-to-3d content creation.
\newblock {\em arXiv preprint arXiv:2211.10440}, 2022.

\bibitem{chen2023fantasia3d}
R.~Chen et~al.
\newblock Fantasia3d: Disentangling geometry and appearance for high-quality text-to-3d content creation.
\newblock In {\em ICCV}, 2023.

\bibitem{shen2021dmtet}
T.~Shen et~al.
\newblock Deep marching tetrahedra: a hybrid representation for high-resolution 3d shape synthesis.
\newblock In {\em NeurIPS}, 2021.

\bibitem{wang2024prolificdreamer}
Z.~Wang et~al.
\newblock Prolificdreamer: High-fidelity and diverse text-to-3d generation with variational score distillation.
\newblock {\em NeurIPS}, 2024.

\bibitem{xu2023dream3d}
J.~Xu et~al.
\newblock Dream3d: Zero-shot text-to-3d synthesis using 3d shape prior and text-to-image diffusion models.
\newblock In {\em CVPR}, 2023.

\bibitem{shi2023mvdream}
Y.~Shi et~al.
\newblock Mvdream: Multi-view diffusion for 3d generation.
\newblock {\em arXiv preprint arXiv:2308.16512}, 2023.

\bibitem{zhang2024text2nerf}
J.~Zhang et~al.
\newblock Text2nerf: Text-driven 3d scene generation with neural radiance fields.
\newblock {\em TVCG}, 2024.

\bibitem{sarkar2025crossover}
S.~D. Sarkar et~al.
\newblock Crossover: 3d scene cross-modal alignment.
\newblock In {\em CVPR}, pp. 8985--8994, 2025.

\bibitem{richardson2023texture}
E.~Richardson et~al.
\newblock Texture: Text-guided texturing of 3d shapes.
\newblock In {\em ACM SIGGRAPH}, 2023.

\bibitem{chen2023text2tex}
D.~Z. Chen et~al.
\newblock Text2tex: Text-driven texture synthesis via diffusion models.
\newblock In {\em ICCV}, 2023.

\bibitem{chen2023scenetex}
D.~Z. Chen et~al.
\newblock Scenetex: High-quality texture synthesis for indoor scenes via diffusion priors.
\newblock {\em arXiv preprint arXiv:2311.17261}, 2023.

\bibitem{jiang2023avatarcraft}
R.~Jiang et~al.
\newblock Avatarcraft: Transforming text into neural human avatars with parameterized shape and pose control.
\newblock In {\em ICCV}, 2023.

\bibitem{hong2022avatarclip}
F.~Hong et~al.
\newblock Avatarclip: Zero-shot text-driven generation and animation of 3d avatars.
\newblock {\em ACM TOG}, 2022.

\bibitem{li2024genzi}
L.~Li and A.~Dai.
\newblock Genzi: Zero-shot 3d human-scene interaction generation.
\newblock In {\em CVPR}, 2024.

\bibitem{diller2024cghoi}
C.~Diller and A.~Dai.
\newblock Cg-hoi: Contact-guided 3d human-object interaction generation.
\newblock In {\em CVPR}, 2024.

\bibitem{vilesov2023cg3d}
A.~Vilesov et~al.
\newblock Cg3d: Compositional generation for text-to-3d via gaussian splatting.
\newblock {\em arXiv preprint arXiv:2311.17907}, 2023.

\bibitem{po2023compositional}
R.~Po and G.~Wetzstein.
\newblock Compositional 3d scene generation using locally conditioned diffusion.
\newblock {\em arXiv preprint arXiv:2303.12218}, 2023.

\bibitem{gao2024graphdreamer}
G.~Gao et~al.
\newblock Graphdreamer: Compositional 3d scene synthesis from scene graphs.
\newblock In {\em CVPR}, 2024.

\bibitem{3dvista}
Z.~Ziyu et~al.
\newblock 3d-vista: Pre-trained transformer for 3d vision and text alignment.
\newblock In {\em ICCV}, 2023.

\bibitem{chen2024spatialvlm}
B.~Chen et~al.
\newblock Spatialvlm: Endowing vision-language models with spatial reasoning capabilities.
\newblock {\em arXiv preprint arXiv:2401.12168}, 2024.

\bibitem{delitzas2023multiclip}
A.~Delitzas et~al.
\newblock Multi-clip: Contrastive vision-language pre-training for question answering tasks in 3d scenes.
\newblock {\em arXiv preprint arXiv:2306.02329}, 2023.

\bibitem{man2024lexicon3d}
Y.~Man et~al.
\newblock Lexicon3d: Probing visual foundation models for complex 3d scene understanding.
\newblock {\em NeurIPS}, 37:76819--76847, 2024.

\bibitem{d3net}
D.~Z. Chen et~al.
\newblock D3net: A unified speaker-listener architecture for 3d dense captioning and visual grounding.
\newblock In {\em ECCV}, 2022.

\bibitem{li2023uni3dl}
X.~Li et~al.
\newblock Uni3dl: Unified model for 3d and language understanding.
\newblock {\em arXiv:2310.09478}, 2023.

\bibitem{yuan2021instancerefer}
Z.~Yuan et~al.
\newblock Instancerefer: Cooperative holistic understanding for visual grounding on point clouds through instance multi-level contextual referring.
\newblock In {\em ICCV}, 2021.

\bibitem{roh2022languagerefer}
J.~Roh et~al.
\newblock Languagerefer: Spatial-language model for 3d visual grounding.
\newblock In {\em CoRL}, 2022.

\bibitem{zhao20213dvg}
L.~Zhao et~al.
\newblock 3dvg-transformer: Relation modeling for visual grounding on point clouds.
\newblock In {\em ICCV}, 2021.

\bibitem{cen2025segment}
J.~Cen et~al.
\newblock Segment any 3d gaussians.
\newblock In {\em Proceedings of the AAAI Conference on Artificial Intelligence}, volume~39, pp. 1971--1979, 2025.

\bibitem{luo2024scalable}
T.~Luo et~al.
\newblock Scalable 3d captioning with pretrained models.
\newblock {\em NeurIPS}, 36, 2024.

\bibitem{chen2019text2shape}
K.~Chen et~al.
\newblock Text2shape: Generating shapes from natural language by learning joint embeddings.
\newblock In {\em ACCV}, 2019.

\bibitem{jia2024sceneverse}
B.~Jia et~al.
\newblock Sceneverse: Scaling 3d vision-language learning for grounded scene understanding.
\newblock {\em arXiv preprint arXiv:2401.09340}, 2024.

\bibitem{wang2023embodiedscan}
T.~Wang et~al.
\newblock Embodiedscan: A holistic multi-modal 3d perception suite towards embodied ai.
\newblock {\em arXiv preprint arXiv:2312.16170}, 2023.

\bibitem{abdelreheem2024scanents3d}
A.~Abdelreheem et~al.
\newblock Scanents3d: Exploiting phrase-to-3d-object correspondences for improved visio-linguistic models in 3d scenes.
\newblock In {\em WACV}, 2024.

\bibitem{lin2023wildrefer}
Z.~Lin et~al.
\newblock Wildrefer: 3d object localization in large-scale dynamic scenes with multi-modal visual data and natural language.
\newblock {\em arXiv preprint arXiv:2304.05645}, 2023.

\bibitem{RIORefer}
T.~Miyanishi et~al.
\newblock Cross3dvg: Baseline and dataset for cross-dataset 3d visual grounding on different rgb-d scans.
\newblock {\em arXiv preprint arXiv:2305.13876}, 2023.

\bibitem{ARKitSceneRefer}
S.~Kato et~al.
\newblock Arkitscenerefer: Text-based localization of small objects in diverse real-world 3d indoor scenes.
\newblock In {\em EMNLP}, 2023.

\bibitem{yuan2022toward}
Z.~Yuan et~al.
\newblock Toward explainable and fine-grained 3d grounding through referring textual phrases.
\newblock {\em arXiv preprint arXiv:2207.01821}, 2022.

\bibitem{etesam20223dvqa}
Y.~Etesam et~al.
\newblock 3dvqa: Visual question answering for 3d environments.
\newblock In {\em Conference on Robots and Vision (CRV)}, 2022.

\bibitem{ye2021tvcg3dqa}
S.~Ye et~al.
\newblock 3d question answering, 2021.

\bibitem{yan2023comprehensive}
X.~Yan et~al.
\newblock Comprehensive visual question answering on point clouds through compositional scene manipulation.
\newblock {\em TVCG}, 2023.

\bibitem{linghu2024multi}
X.~Linghu et~al.
\newblock Multi-modal situated reasoning in 3d scenes.
\newblock {\em arXiv preprint arXiv:2409.02389}, 2024.

\bibitem{li2023m3dbench}
M.~Li et~al.
\newblock M3dbench: Let's instruct large models with multi-modal 3d prompts.
\newblock {\em arXiv preprint arXiv:2312.10763}, 2023.

\bibitem{yin2023lamm}
Z.~Yin et~al.
\newblock Lamm: Language-assisted multi-modal instruction-tuning dataset, framework, and benchmark.
\newblock {\em arXiv preprint arXiv:2306.06687}, 2023.

\bibitem{yang2024_3D_GRAND}
J.~Yang et~al.
\newblock 3d-grand: A million-scale dataset for 3d-llms with better grounding and less hallucination, 2024.

\bibitem{chen2025integrating}
Y.~Chen et~al.
\newblock Integrating chain-of-thought for multimodal alignment: A study on 3d vision-language learning.
\newblock {\em arXiv preprint arXiv:2503.06232}, 2025.

\bibitem{huang2025unveiling}
J.~Huang et~al.
\newblock Unveiling the mist over 3d vision-language understanding: Object-centric evaluation with chain-of-analysis.
\newblock In {\em CVPR}, pp. 24570--24581, 2025.

\bibitem{zhang2025flatland}
J.~Zhang et~al.
\newblock From flatland to space: Teaching vision-language models to perceive and reason in 3d.
\newblock {\em arXiv preprint arXiv:2503.22976}, 2025.

\bibitem{szymanska2025space3d}
E.~Szyma{\'n}ska et~al.
\newblock Space3d-bench: Spatial 3d question answering benchmark.
\newblock In {\em European Conference on Computer Vision}, pp. 68--85. Springer, 2025.

\bibitem{sun20243d}
P.~Sun et~al.
\newblock 3d question answering for city scene understanding.
\newblock In {\em Proceedings of the 32nd ACM International Conference on Multimedia}, pp. 2156--2165, 2024.

\bibitem{lyu2024mmscan}
R.~Lyu et~al.
\newblock Mmscan: A multi-modal 3d scene dataset with hierarchical grounded language annotations.
\newblock {\em Advances in Neural Information Processing Systems}, 37:50898--50924, 2024.

\bibitem{nuscenes2019}
H.~Caesar et~al.
\newblock nuscenes: A multimodal dataset for autonomous driving.
\newblock {\em arXiv preprint arXiv:1903.11027}, 2019.

\bibitem{zeng2023large}
F.~Zeng et~al.
\newblock Large language models for robotics: A survey.
\newblock {\em arXiv preprint arXiv:2311.07226}, 2023.

\bibitem{zhou2023language}
H.~Zhou et~al.
\newblock Language-conditioned learning for robotic manipulation: A survey.
\newblock {\em arXiv preprint arXiv:2312.10807}, 2023.

\bibitem{chen2024grounded}
Y.~Chen et~al.
\newblock Grounded 3d-llm with referent tokens.
\newblock {\em arXiv preprint arXiv:2405.10370}, 2024.

\bibitem{luo2025vebrain}
G.~Luo et~al.
\newblock Visual embodied brain: Let multimodal large language models see, think, and control in spaces.
\newblock {\em arXiv preprint arXiv:2506.00123}, 2025.

\bibitem{li2025embodied}
Z.~Li et~al.
\newblock Embodied intelligence for 3d understanding: A survey on 3d scene question answering.
\newblock {\em arXiv preprint arXiv:2502.00342}, 2025.

\bibitem{gaussian_grouping}
M.~Ye et~al.
\newblock Gaussian grouping: Segment and edit anything in 3d scenes.
\newblock In {\em ECCV}, 2024.

\bibitem{xu2025uniugg}
Y.~Xu et~al.
\newblock Uniugg: Unified 3d understanding and generation via geometric-semantic encoding.
\newblock {\em arXiv preprint arXiv:2508.11952}, 2025.

\bibitem{wu2024ptv3}
X.~Wu et~al.
\newblock Point transformer v3: Simpler, faster, stronger.
\newblock In {\em CVPR}, 2024.

\bibitem{AffordanceLLM2024_CVPR}
S.~Qian et~al.
\newblock Affordancellm: Grounding affordance from vision language models.
\newblock In {\em CVPRW}, 2024.

\bibitem{pgvlm2024}
J.~Gao et~al.
\newblock Physically grounded vision-language models for robotic manipulation.
\newblock In {\em ICRA}. IEEE, 2024.

\bibitem{zhan20253DPhysical}
G.~Zhan et~al.
\newblock A general protocol to probe large vision models for 3d physical understanding.
\newblock In {\em NeurIPS}, 2024.

\bibitem{zhan2025inferringdynamicphysicalproperties}
G.~Zhan et~al.
\newblock Inferring dynamic physical properties from video foundation models, 2025.

\bibitem{zhu20254dbench}
W.~Zhu et~al.
\newblock 4d-bench: Benchmarking multi-modal large language models for 4d object understanding.
\newblock {\em arXiv preprint arXiv:2503.17827}, 2025.

\bibitem{zhou2025vlm4d}
S.~Zhou et~al.
\newblock Vlm4d: Towards spatiotemporal awareness in vision language models.
\newblock {\em arXiv preprint arXiv:2508.02095}, 2025.

\bibitem{guo2025deepseek}
D.~Guo et~al.
\newblock Deepseek-r1: Incentivizing reasoning capability in llms via reinforcement learning.
\newblock {\em arXiv preprint arXiv:2501.12948}, 2025.

\bibitem{schuhmann2022laion}
C.~Schuhmann et~al.
\newblock Laion-5b: An open large-scale dataset for training next generation image-text models.
\newblock {\em NeurIPS}, 2022.

\bibitem{ge2024behavior}
Y.~Ge et~al.
\newblock Behavior vision suite: Customizable dataset generation via simulation.
\newblock In {\em CVPR}, 2024.

\bibitem{ocal2024sceneteller}
B.~M. {\"O}cal et~al.
\newblock Sceneteller: Language-to-3d scene generation.
\newblock {\em arXiv preprint arXiv:2407.20727}, 2024.

\bibitem{yang2024scenecraft}
X.~Yang et~al.
\newblock Scenecraft: Layout-guided 3d scene generation.
\newblock In {\em NeurIPS}, 2024.

\bibitem{bai2024hallucinationmultimodallargelanguage}
Z.~Bai et~al.
\newblock Hallucination of multimodal large language models: A survey, 2024.

\bibitem{zhang2024physdreamer}
T.~Zhang et~al.
\newblock {PhysDreamer}: Physics-based interaction with 3d objects via video generation.
\newblock {\em arxiv}, 2024.

\bibitem{wang2023aligninglargelanguagemodels}
Y.~Wang et~al.
\newblock Aligning large language models with human: A survey, 2023.

\end{thebibliography}
}

\end{document}